\documentclass{article} 
\usepackage{arxiv_style,times}

\usepackage{amsmath,amsfonts,bm}

\def\figref#1{Figure~\ref{#1}}

\def\Secref#1{Section~\ref{#1}}

\def\eqref#1{equation~\ref{#1}}
\def\Eqref#1{Equation~\ref{#1}}

\def\1{\bm{1}}

\DeclareMathAlphabet{\mathsfit}{\encodingdefault}{\sfdefault}{m}{sl}
\SetMathAlphabet{\mathsfit}{bold}{\encodingdefault}{\sfdefault}{bx}{n}

\usepackage{hyperref}
\usepackage{url}
\usepackage{multirow}

\title{Noise-free Score Distillation}

\author{Oren Katzir$^{1}$\thanks{\,\,Denotes Equal Contribution} \And Or Patashnik$^{1}$\footnotemark[1] \And Daniel Cohen-Or$^{1}$ \And Dani Lischinski$^{2}$ \vspace{-15pt} \And \hspace{30pt} $^{1}$\normalfont{\textit{Tel-Aviv University}} \And $^{2}$\normalfont{\textit{The Hebrew University of Jerusalem}}}

\usepackage{graphicx}

\usepackage{tikz}
\usetikzlibrary{spy}
\usepackage{comment}

\usepackage{color}

\usepackage{array}
\usepackage{tabu}
\usepackage{array}
\usepackage{booktabs}
\usepackage{soul}
\usepackage{xspace}
\usepackage{amsmath}
\usepackage{amssymb}
\usepackage{wrapfig}

\usepackage{xcolor}

\usepackage{bbm}

\usepackage[normalem]{ulem}
\usepackage{multirow}
\usepackage{adjustbox}

\usepackage{arydshln}
\usetikzlibrary{arrows.meta}

\newif\ifdraft
\drafttrue
\draftfalse

\ifdraft
\newcommand{\okc}[1]{{\color{orange}[\textbf{Oren:} #1]}}
\newcommand{\opc}[1]{{\color{blue}[\textbf{Or:} #1]}}
\newcommand{\dcc}[1]{{\color{red}[\textbf{DC:} #1]}}
\newcommand{\dlc}[1]{{\color{purple}[\textbf{DL:} #1]}}

\newcommand{\op}[1]{{\color{blue}#1}}

\newcommand{\dl}[1]{{\color{purple}#1}}

\else
\newcommand{\okc}[1]{}
\newcommand{\opc}[1]{}
\newcommand{\dcc}[1]{}
\newcommand{\dlc}[1]{}

\newcommand{\op}[1]{{\color{black}#1}}

\newcommand{\dl}[1]{{\color{black}#1}}
\fi

\definecolor{mygray}{RGB}{140, 140, 140}

\definecolor{dashedpurple}{RGB}{118,0,103}
\definecolor{dashedorange}{RGB}{181,100,13}

\newcommand{\Loss}{\mathcal{L}}
\newcommand{\Rspace}{\mathbb{R}}

\newcommand{\dirn}{\delta_\text{N}}
\newcommand{\dirr}{\delta_\text{D}}
\newcommand{\dirc}{\delta_\text{C}}

\newcommand{\pneg}{p_\text{neg}}
\newcommand{\z}{\mathbf{z}}
\newcommand{\x}{\mathbf{x}}
\newcommand{\grad}[1]{\nabla_{\!#1}}
\newcommand{\unet}{\epsilon_{\phi}}
\newcommand{\epred}{\unet(\z_t;y,t)}
\newcommand{\epuncond}{\unet(\z_t;\varnothing,t)}

\newcommand{\epneg}{\unet(\z_t;y=\pneg,t)}
\newcommand{\epredcfg}{\unet^{s}(\z_t;y,t)}
\newcommand{\epredcfgx}{\unet^{s}(\z_t(\x);y,t)}

\makeatletter
\newcommand{\thickhline}{%
    \noalign {\ifnum 0=`}\fi \hrule height 1pt
    \futurelet \reserved@a \@xhline
}
\makeatletter
\newcommand{\thickvline}{%
    \noalign {\ifnum 0=`}\fi \vrule height 1pt
    \futurelet \reserved@a \@xvline
}
\newcolumntype{"}{@{\hskip\tabcolsep\vrule width 1pt\hskip\tabcolsep}}
\makeatother

\DeclareMathAlphabet\mathbfcal{OMS}{cmsy}{b}{n}

\newcolumntype{C}[1]{>{\centering\arraybackslash}p{#1}}
\iclrfinalcopy 
\begin{document}

\maketitle
\begin{abstract}
Score Distillation Sampling (SDS) has emerged as the de facto approach for text-to-content generation in non-image domains. In this paper, we reexamine the SDS process and introduce a straightforward interpretation that demystifies the necessity for large Classifier-Free Guidance (CFG) scales, rooted in the distillation of an undesired noise term. Building upon our interpretation, we propose a novel Noise-Free Score Distillation (NFSD) process, which requires minimal modifications to the original SDS framework.
Through this streamlined design, we achieve more effective distillation of pre-trained text-to-image diffusion models while using a nominal CFG scale. This strategic choice allows us to prevent the over-smoothing of results, ensuring that the generated data is both realistic and complies with the desired prompt.
To demonstrate the efficacy of NFSD, we provide qualitative examples that compare NFSD and SDS, as well as several other methods.

\vspace{-6pt}
%

\end{abstract}

\section{Introduction}
\label{sec:intro}
\vspace{-6pt}

Image synthesis has recently witnessed significant progress in terms of image quality and diversity~\citep{Yu2022ScalingAM, Ding2021CogViewMT, Gafni2022MakeASceneST, Chang2022MaskGITMG, kang2023gigagan, chang2023muse}. Specifically, text-to-image models are rapidly improving, with diffusion-based methods leading the way~\citep{SohlDickstein2015DeepUL, ho2020denoising, Dhariwal2021DiffusionMB, rombach2022high, balaji2022eDiff-I, Saharia2022PhotorealisticTD}. 
Seeking to project the great power of such diffusion-based text-to-image models to other domains beyond images, Score Distillation Sampling (SDS) was introduced. In their seminal work, DreamFusion, \citet{poole2022dreamfusion} introduce the SDS loss which utilizes the strong prior learned by a text-to-image diffusion model to optimize a NeRF~\citep{mildenhall2020nerf} based on a single text prompt. Other works have shown that this mechanism can also be used to optimize other representations, such as meshes~\citep{Chen_2023_ICCV}, texture maps~\citep{metzer2022latent, tsalicoglou2023textmesh}, fonts~\citep{iluz2023word, tanveer2023ds}, and SVG~\citep{jain2023vectorfusion}.

Despite the widespread adoption of the SDS loss in various domains and representations, there is still a gap in visual quality between images generated by the standard denoising diffusion process (ancestral sampling) and those resulting from an SDS-based optimization process. Specifically, as noted by previous works~\citep{poole2022dreamfusion, wang2023prolificdreamer, zhu2023hifa}, SDS tends to produce over-smoothed and over-saturated results, exhibiting limited ability to generate fine details, a trait where modern text-to-image models typically excel. Furthermore, the SDS loss remains intriguing, as it is still not fully understood.

In this paper, inspired by SDS~\citep{poole2022dreamfusion}, we present a general framework that allows using a pretrained diffusion model to optimize a differentiable image renderer. Treating the diffusion model as a score function~\citep{song2020score}, we propose a formulation that decomposes the score into three intuitively interpretable components: alignment with the condition, domain correction, and denoising. Based on insights gained by viewing the score function in light of our new decomposition, we introduce a new Noise-Free Score Distillation (NFSD) loss, and show that it outperforms SDS without incurring any additional computational costs.

\begin{figure}
        \centering
    \setlength{\tabcolsep}{0pt}
    \begin{tabular}{c c c c c}
        \hspace{-0.17cm} \includegraphics[width=0.318\linewidth]{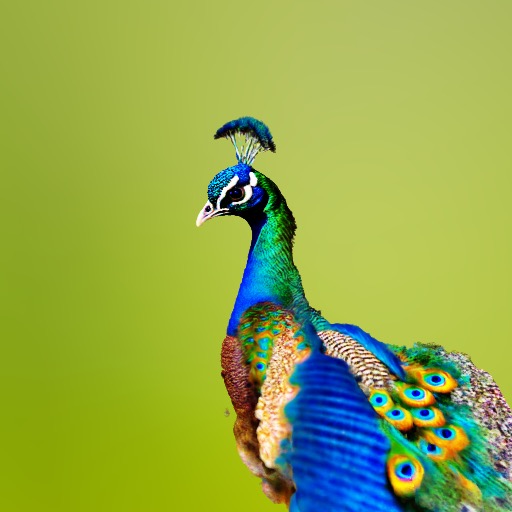} &
        \raisebox{0.783\height}{%
            \begin{tabular}[b]{c}
            \adjustbox{valign=t}{\includegraphics[width=0.159\linewidth]{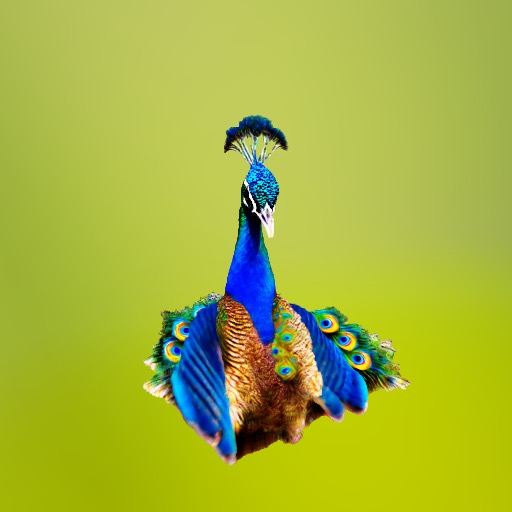}}
            \\
            \adjustbox{valign=t}{\includegraphics[width=0.159\linewidth]{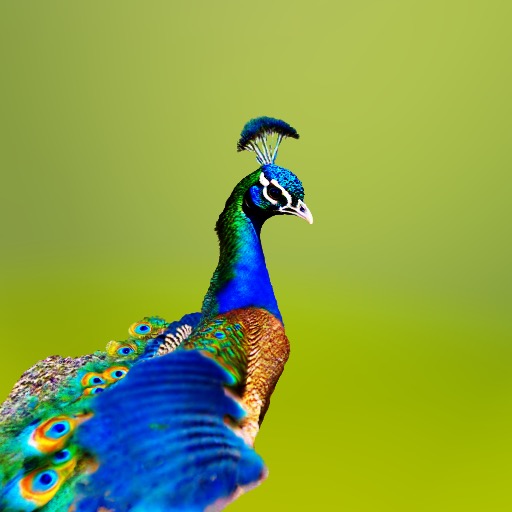}} 
        \end{tabular} 
        } &
         &
        \includegraphics[width=0.318\linewidth]{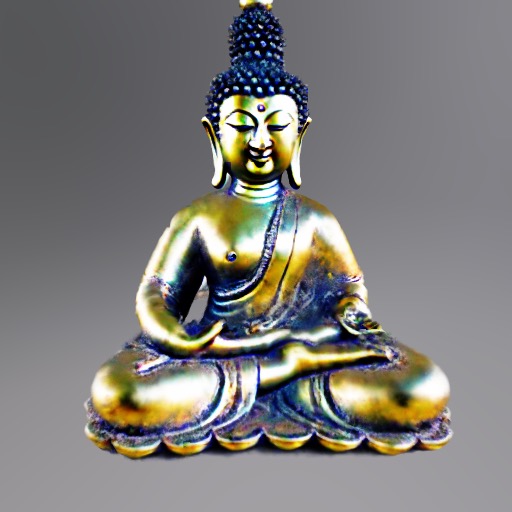} &
        \raisebox{0.783\height}{%
            \begin{tabular}[b]{c}
            \adjustbox{valign=t}{\includegraphics[width=0.159\linewidth]{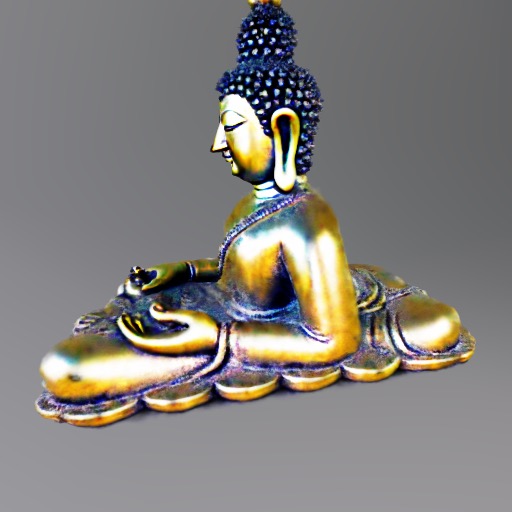}}
            \\
            \adjustbox{valign=t}{\includegraphics[width=0.159\linewidth]{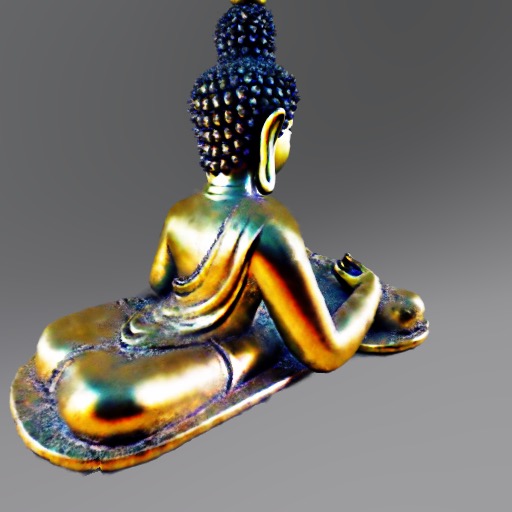}} 
        \end{tabular} 
        } 
    \end{tabular}

    \begin{tabular}{c c c }
        \includegraphics[width=0.32\linewidth]{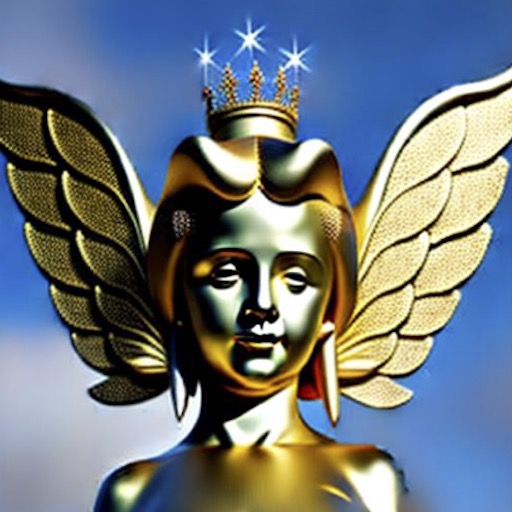} &
        \includegraphics[width=0.32\linewidth]{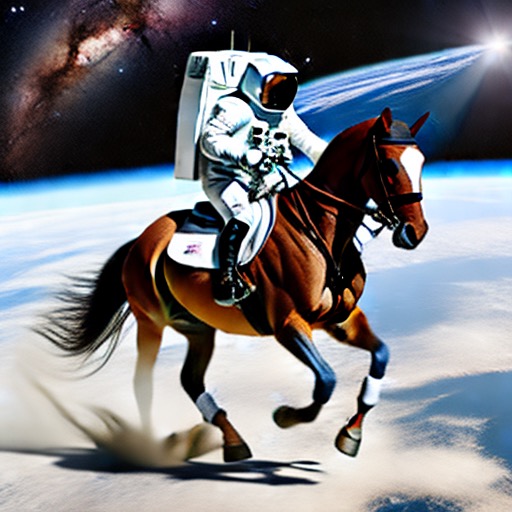} &
        \raisebox{0.78401\height}{%
        \begin{tabular}[b]{c c}
            \adjustbox{valign=t}{\includegraphics[width=0.16\linewidth]{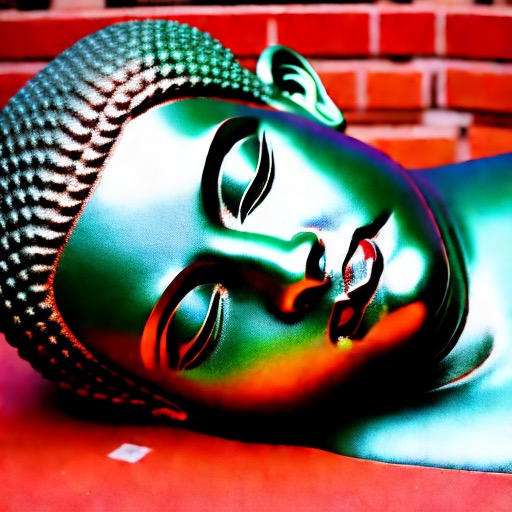}} &
            \adjustbox{valign=t}{\includegraphics[width=0.16\linewidth]{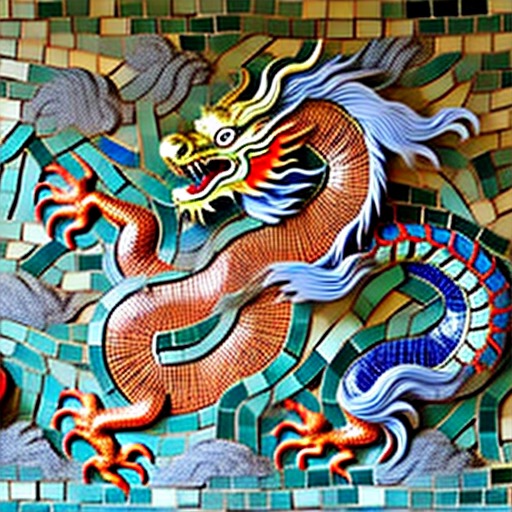}}
            \\
            \adjustbox{valign=t}{\includegraphics[width=0.16\linewidth]{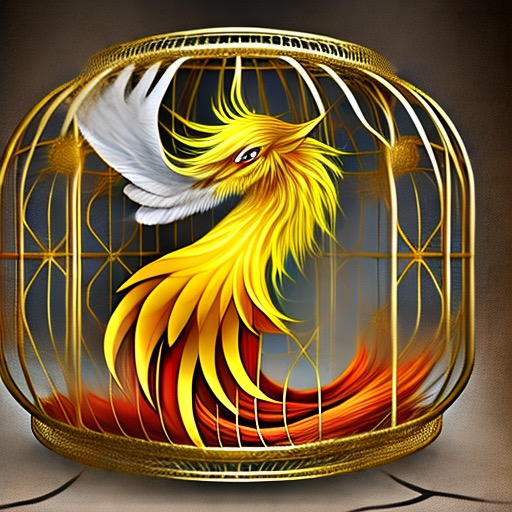}} &
            \adjustbox{valign=t}{\includegraphics[width=0.16\linewidth]{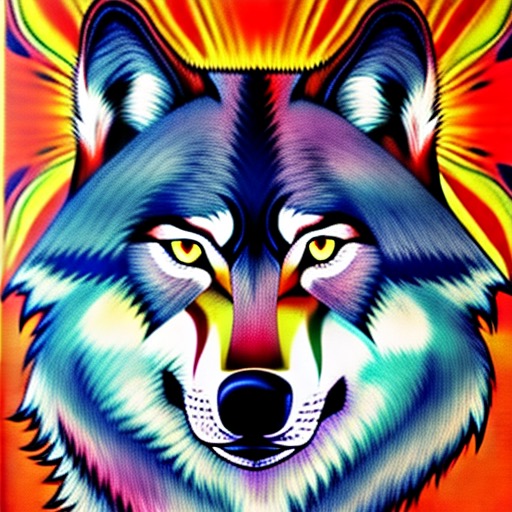}}
        \end{tabular}
        }
    \end{tabular}
    \vspace{-6pt}
    \caption{Results obtained with our Noise Free Score Distillation (NFSD). Top: two learnt NeRFs (movies of these and many other examples are included in the supplementary materials). Bottom: a gallery of images optimized with NFSD.}
    \vspace{-10pt}
    \label{fig:teaser}
\end{figure}

To demonstrate the general nature of our framework, we show that our novel formulation supports and provides a more concise and straightforward explanation for recent methods, such as VSD~\citep{wang2023prolificdreamer} and DDS~\citep{hertz2023delta}, which have shown improvements over SDS.
We validate our formulation and approach by utilizing Stable Diffusion~\citep{rombach2022high} as our score function with a focus on images and NeRFs as our representations. A few example results are showcased in Figure~\ref{fig:teaser}. 
Through careful design, our Noise-Free Score Distillation (NFSD) addresses some of the issues present in SDS and leads to improved visual results.

\section{Background}
\label{sec:sd}
\vspace{-6pt}

In this section, we provide the necessary background regarding diffusion models and the SDS loss \citep{poole2022dreamfusion} that enables text-to-3D generation by optimizing the parameters of a differentiable image generation function.
\vspace{-5pt}

\paragraph{Diffusion models}
Diffusion models~\citep{SohlDickstein2015DeepUL, ho2020denoising} are a family of generative models that are trained to gradually transform Gaussian noise into samples from a target distribution $p_{\text{data}}$. Starting from an initial noise $\z_T \sim \mathcal{N}(0,\mathrm{I})$, at each diffusion timestep $t$, the model takes as input a noisy sample $\z_t$, and predicts a cleaner sample $\z_{t-1}$, until finally obtaining $\z_0 = \x \sim p_{\text{data}}$. Thus, such models effectively learn the transitions $p(\z_{t-1} | \z_t)$.

Commonly, diffusion models are parameterized by a U-net $\unet(\z_t,t)$~\citep{ho2020denoising}, which predicts the noise $\epsilon$ that was used to produce $\z_t$ from $\x = \z_0$, rather than predicting $\x$ or $\z_{t-1}$ directly. This is known as $\epsilon$-prediction. Previous works~\citep{ho2020denoising, song2020score} have also observed that $\unet(\z_t,t)$ is proportional to the predicted score function~\citep{hyvarinen2005estimation} of the smoothed denisty $\grad{\z_t} \log p_{t}(\z_t)$, where $p_t$ is the marginal distribution of the samples noised to time $t$. The score function is a vector ﬁeld that points towards higher density of data at a given noise level. Thus, intuitively, taking steps in the direction of the score function gradually moves the sample towards the data distribution. 

In this work, we focus on diffusion models that strive to generate samples aligned with a given condition $y$ (e.g., class, text prompt). To this end, the diffusion process is conditioned on $y$. This is typically achieved via classifier-free guidance (CFG)~\citep{ho2022classifier}, where the conditioned prediction $\epred$ of the noise is extrapolated away from the unconditioned prediction $\epuncond$ by an amount controlled by a scalar $s \in \Rspace$:
\begin{equation}
    \label{eq:cfg}
	\epredcfg = \unet(\z_t ; y=\varnothing, t) + s \left(\epred - \unet(\z_t ; y=\varnothing, t) \right),
\end{equation}
where $\varnothing$ indicates a null condition (unconditioned). CFG modifies the score function to steer the process towards regions with a higher ratio of conditional density to the unconditional one. However, it has been observed that CFG trades sample ﬁdelity for diversity~\citep{ho2022classifier}. 

\vspace{-5pt}
\paragraph{Score Distillation Sampling (SDS)}
Over the last two years, text-to-image diffusion methods~\citep{rombach2022high, Saharia2022PhotorealisticTD, Ramesh2022HierarchicalTI, podell2023sdxl} have achieved unprecedented image generation results by incorporating textual encoder outputs as a condition to the diffusion model. These powerful models are trained on billions of text-image pairs, and such extensive data is currently not available for other domains. The recent introduction of Score Distillation Sampling (SDS)~\citep{poole2022dreamfusion,wang2023score} enables leveraging the priors of pre-trained text-to-image models to facilitate text-conditioned generation in other domains, particularly 3D content generation.

Specifically, given a pretrained diffusion model $\unet$, SDS optimizes a set of parameters $\theta$ of a differentiable parametric image generator $g$, using the gradient of the loss $\Loss_\text{SDS}$ with respect to $\theta$:
\begin{equation}
    \grad{\theta} \Loss_\text{SDS} = w(t) \left(\epredcfgx - \epsilon \right) \frac{\partial \x}{\partial \theta},
   \label{eq:SDS-loss}
\end{equation}
where $\x = g(\theta)$ is an image rendered by $\theta$, $\z_t(\x)$ is obtained by adding a Gaussian noise $\epsilon$ to $\x$ corresponding to the $t$-th timestep of the diffusion process, and $y$ is a condition to the diffusion model. In practice, at every optimization iteration, different values of $t$ and Gaussian noise $\epsilon$ are randomly drawn. The parameters $\theta$ are then optimized by computing the gradient of $\Loss_\text{SDS}$ with respect to $\x$ and backpropagating this gradient through the differentiable parametric function $g$. 

\citet{poole2022dreamfusion} formally show that $\Loss_\text{SDS}$ minimizes the KL divergence between a family of Gaussian distributions around $\x$ and the distributions $p(\z_t,y,t)$ learned by the pretrained diffusion model.
Intuitively, \Eqref{eq:SDS-loss} can be interpreted as follows:
since $\x = g(\theta)$ is a clean rendered image, Gaussian noise is first added to it in order to approximately project it to the manifold of noisy images corresponding to timestep $t$. Next, the score $\epredcfgx$ provides the direction in which this noised version of $\x$ should be moved towards a denser region in the distribution of real images (noised to timestep $t$ and aligned with the condition $y$).  Finally, before the resulting direction can be used to optimize $\theta$, the initially added noise $\epsilon$ is subtracted. We interpret this last step as an attempt to adapt the direction back to the domain of clean rendered images. 

While SDS provides an elegant mechanism for leveraging pretrained text-to-image models, SDS-generated results often suffer from oversaturation and \op{lack of fine realistic details}. 
These issues were, in part, attributed to the use of a high CFG value \citep{wang2023prolificdreamer}, which \citet{poole2022dreamfusion} empirically found to be necessary to obtain their results.
Several derivative approaches have emerged to address these challenges~\citep{metzer2022latent, lin2023magic3d, Chen_2023_ICCV, wang2023prolificdreamer, huang2023dreamtime}.

One effective approach for improving the generation quality is time annealing, which gradually reduces the diffusion timesteps $t$ drawn by the optimization process, as it progresses~\citep{lin2023magic3d, zhu2023hifa, wang2023prolificdreamer, huang2023dreamtime}. 
Recently, VSD~\citep{wang2023prolificdreamer} and HiFA~\citep{zhu2023hifa} reformulated the distillation loss. HiFA uses a denoised image version instead of the noise prediction, while VSD offers a variational approach, matching the prediction of noisy real images with that of the noisy rendered images via an additional fine-tuned diffusion model.
In image editing, DDS~\citep{hertz2023delta} observed artifacts when applying SDS to edit real images, which was attributed to a bias in SDS. To mitigate this bias, DDS employs a subtraction of two SDS terms.

In the next section we propose our novel interpretation of SDS via decomposition of the predicted score function into three interpretable components. The insights gained from this decomposition lead us to propose a simple yet effective improvement to SDS, which we call Noise-Free Score Distillation (NFSD), in Section~\ref{sec:nfsd}. Furthermore, this decomposition
enables a simple and unified interpretation of the recent progress in SDS, as discussed in Section~\ref{sec:discussion}. 

\section{Score Decomposition}
\label{sec:score-decomp}
\vspace{-4pt}

As discussed earlier, the noise predicted by a trained diffusion model aims to be proportional to the score function $\grad{\z_t} \log p_{t}(\z_t)$, where $p_t$ is the marginal distribution of the samples noised to time $t$.
In order to gain a better understanding of SDS, it is helpful to examine the decomposition of the score direction into several intuitively interpretable components.

First, consider the difference $\dirc = \epred - \epuncond$ in \Eqref{eq:cfg}.
While $\epred$ ideally points towards a local maximum in the probability density of noisy real images conditioned on $y$, $\epuncond$ points towards a denser region in the distribution of unconditioned noisy images. Thus, the difference $\dirc$ between the two predictions may be thought of as the direction that steers the generated image towards alignment with the condition $y$, and we henceforth refer to it as the \emph{condition direction}. 

\begin{figure}
    \centering
    \setlength{\tabcolsep}{1pt}
    {\small
    \begin{tabular}{C{0.135\linewidth} c C{0.135\linewidth} C{0.135\linewidth} C{0.135\linewidth} C{0.135\linewidth} C{0.135\linewidth} C{0.135\linewidth}}
        \includegraphics[width=\linewidth]{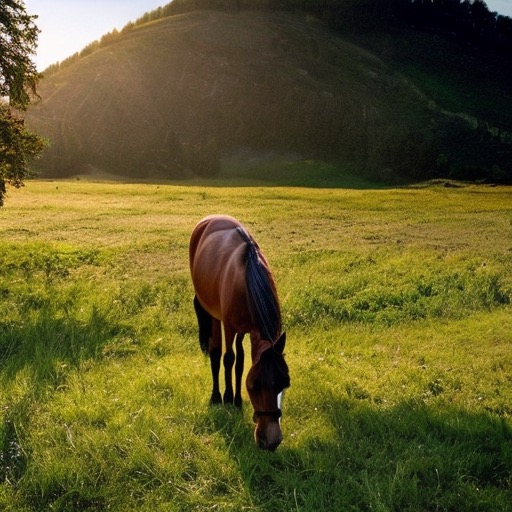} &
        { } &
        \includegraphics[width=\linewidth]{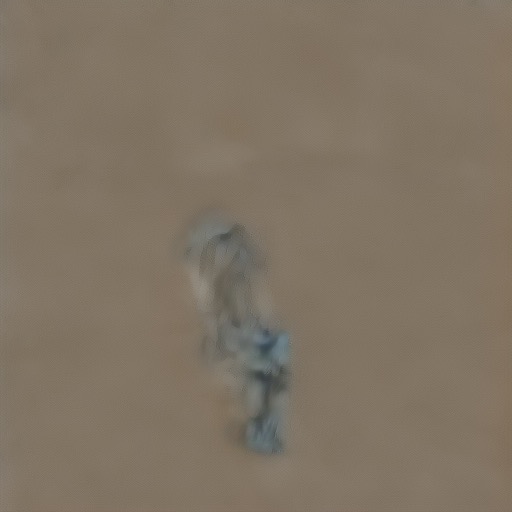} &
        \includegraphics[width=\linewidth]{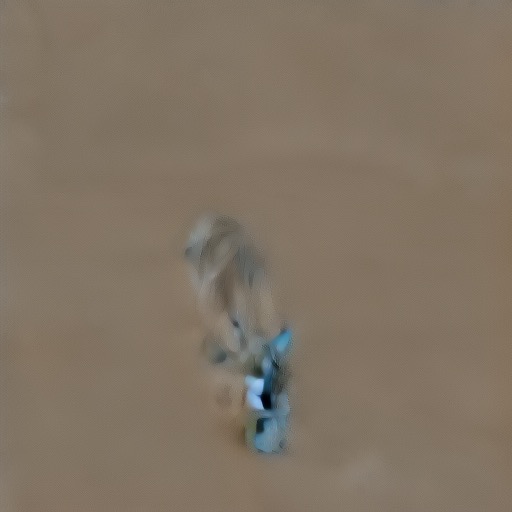} &
        \includegraphics[width=\linewidth]{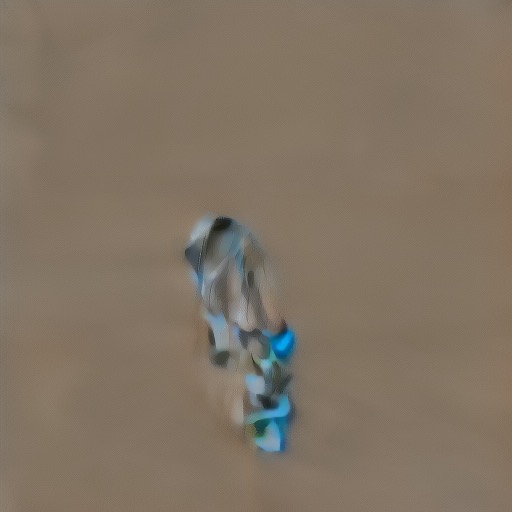} &
        \includegraphics[width=\linewidth]{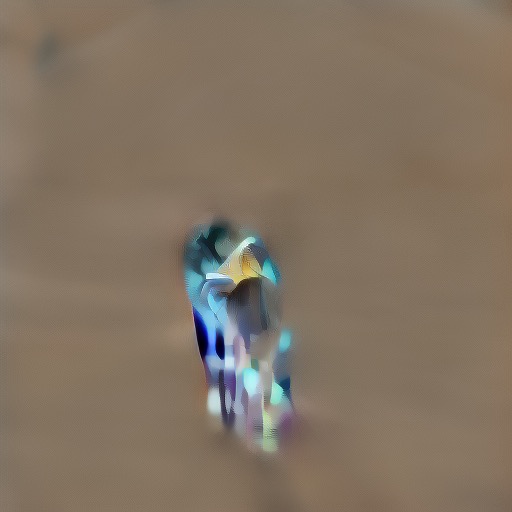} &
        \includegraphics[width=\linewidth]{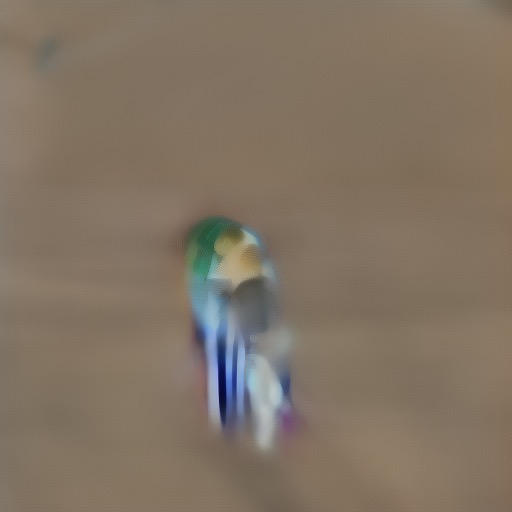} &
        \includegraphics[width=\linewidth]{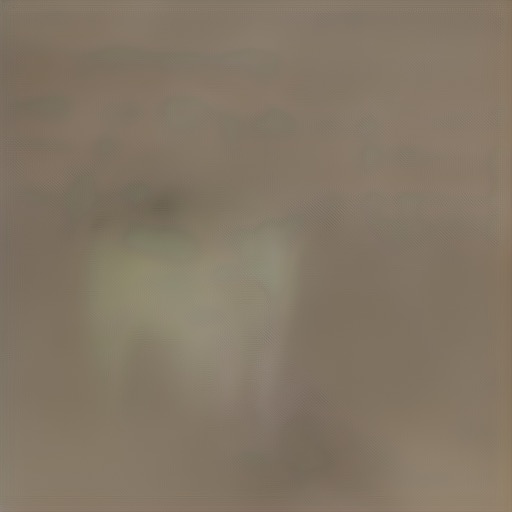} 
        \\
        \includegraphics[width=\linewidth]{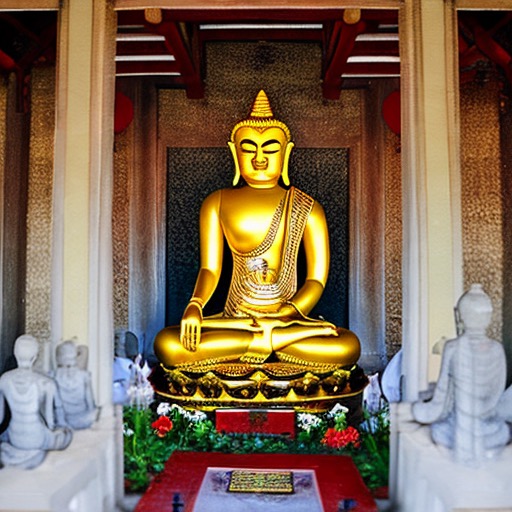} &
        { } &
        \includegraphics[width=\linewidth]{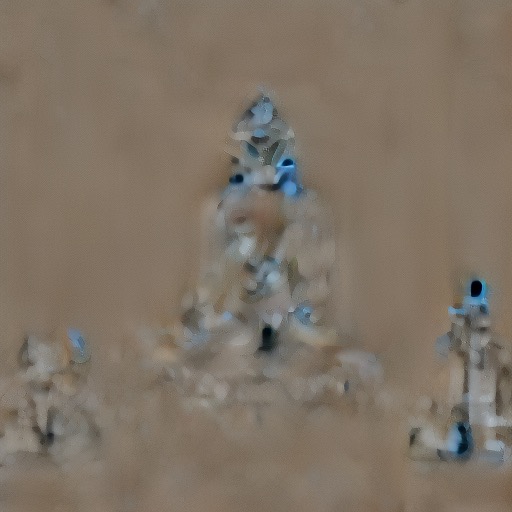} &
        \includegraphics[width=\linewidth]{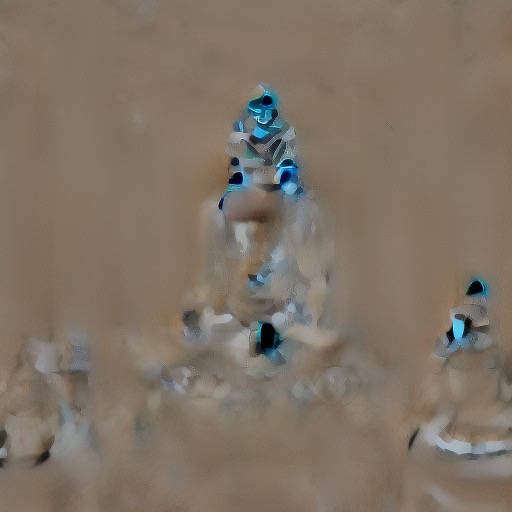} &
        \includegraphics[width=\linewidth]{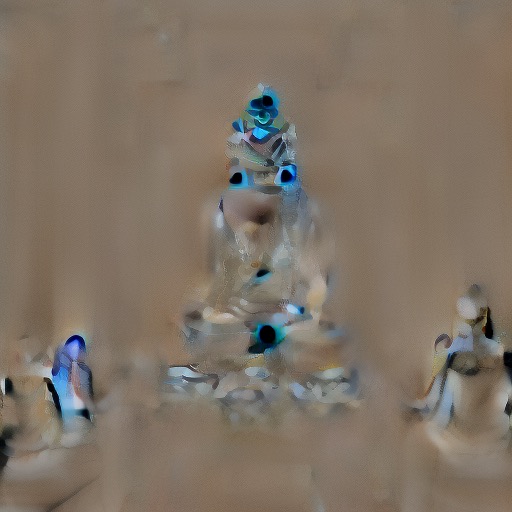} &
        \includegraphics[width=\linewidth]{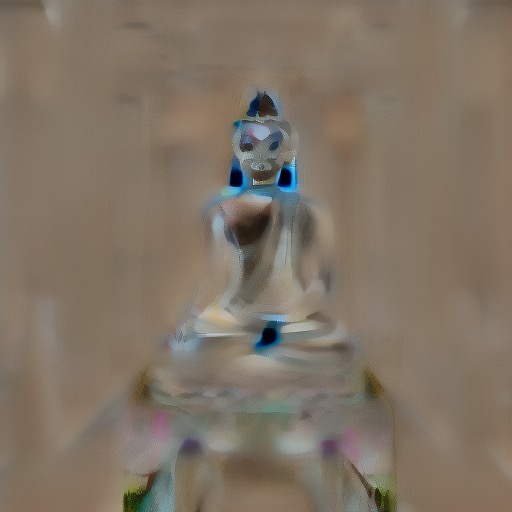} &
        \includegraphics[width=\linewidth]{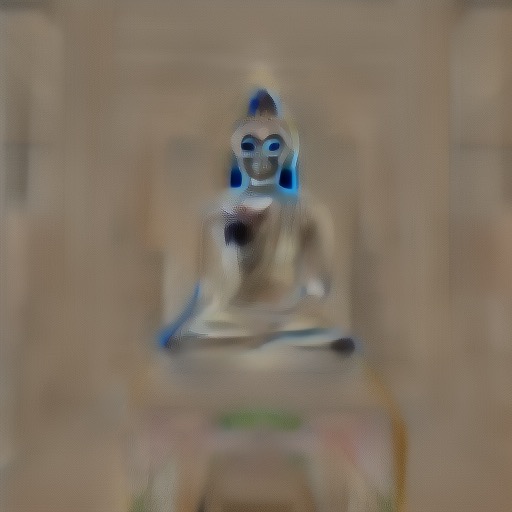} &
        \includegraphics[width=\linewidth]{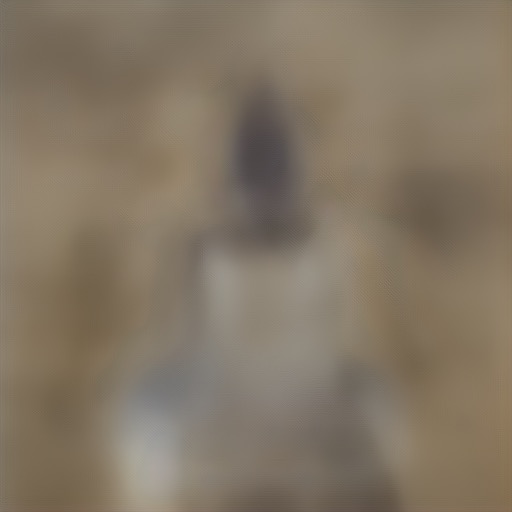} 
        \\
        %
        && $t=100$ & $t=200$ & $t=300$ & $t=500$ & $t=700$ & $t=1000$
    \end{tabular}
    }
    \vspace{-6pt}
    \caption{\label{fig:d_c}
Visualization of $\dirc$. The images in the left column are generated by Stable Diffusion (SD version 2.1-base) with the prompts ``A photo of a horse in a meadow'' and ``A statue of Buddha''. The other columns visualize $\dirc$ for added noise $\epsilon$ with magnitude corresponding to different diffusion timesteps $t$. As can be seen, $\dirc$ is fairly clean and concentrated around the main object in the image. (The visualization is done by decoding each $\dirc$ using the VAE decoder of SD, please refer to \op{Appendix~\ref{sec:app-vis-latent}} for more details).
}
\vspace{-12pt}

\end{figure}

The condition direction $\dirc$ is empirically observed to be uncorrelated with the added noise $\epsilon$ and having its significant magnitudes around the condition-specific image regions. As demonstrated in \figref{fig:d_c}, $\dirc$ is consistently aligned with the condition $y$ for noise corresponding to different timesteps $t$ of the diffusion process. 
This observation is consistent with the inspiration behind CFG~\citep{ho2022classifier}: classifier-guidance of an implicit classifier $\grad{\z_t} \log p^i(y|\z_t)$. Extending this rationale, such a classifier, trained on noisy data $\z_t$, should be invariant to the additive noise $\epsilon$, and its gradients with respect to the input image $\z_t$ should focus on details in $\z_t$ that are most relevant to $y$.

Rewriting \Eqref{eq:cfg} using the condition direction $\dirc$ defined above, we obtain:
\begin{equation}
    \epredcfg = \epuncond + s(\epred - \epuncond) = \epuncond + s\dirc.
\end{equation}

By the nature of its training, the unconditional term $\epuncond$ is expected to predict the noise $\epsilon$ that was added to an image $\x \sim p_{\text{data}}$ to produce $\z_t$. However, in SDS, $\z_t$ is obtained by adding noise to an out-of-distribution (OOD) rendered image $\x = g(\theta)$, which is not sampled from $p_{\text{data}}$. Thus, we can think of $\epuncond$ as a combination of two components, $\epuncond = \dirr + \dirn$,
where $\dirr$ is the \emph{domain correction} induced by the difference between the distributions of rendered and real images, while $\dirn$ is the \emph{denoising direction}, pointing towards a cleaner image. Intuitively, we expect $\dirr$ to be correlated with the content of $\x(\theta)$, while no such correlation is expected for $\dirn$.

\begin{figure}
     \centering
    \setlength{\tabcolsep}{1pt}
    {\small
    \begin{tabular}{C{0.16\linewidth} C{0.16\linewidth} C{0.16\linewidth} C{0.16\linewidth} C{0.16\linewidth} C{0.16\linewidth}}
        \includegraphics[width=\linewidth]{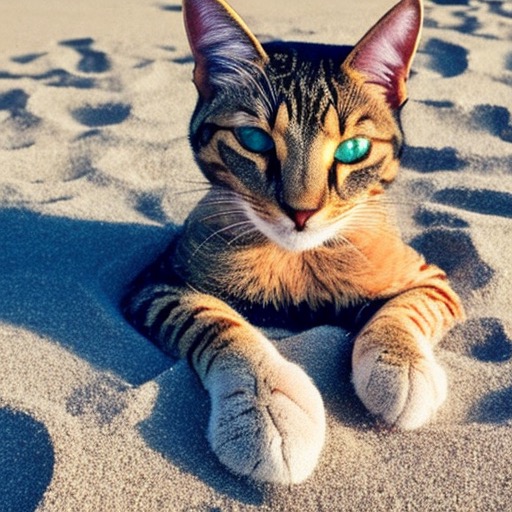} &
        \includegraphics[width=\linewidth]{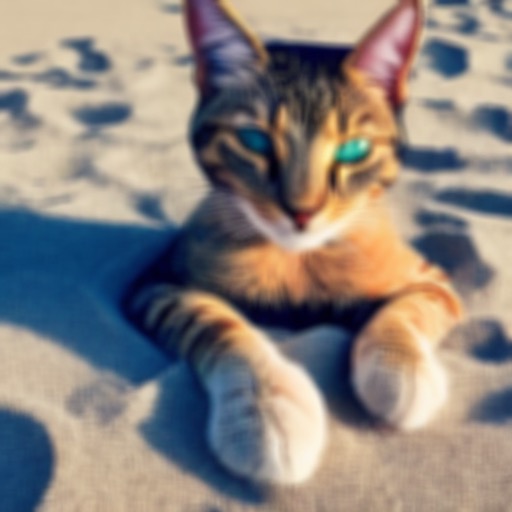} &
        \includegraphics[width=\linewidth]{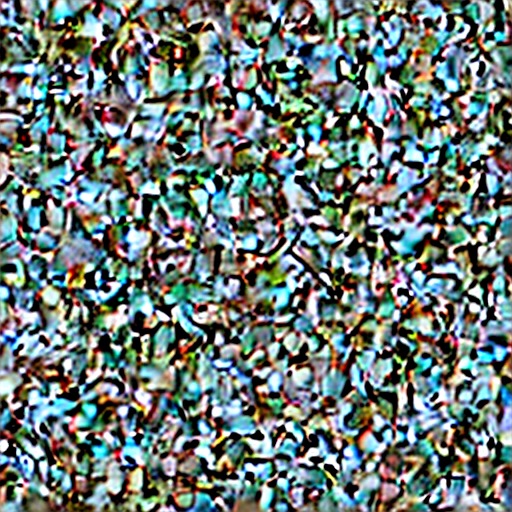} &
        \includegraphics[width=\linewidth]{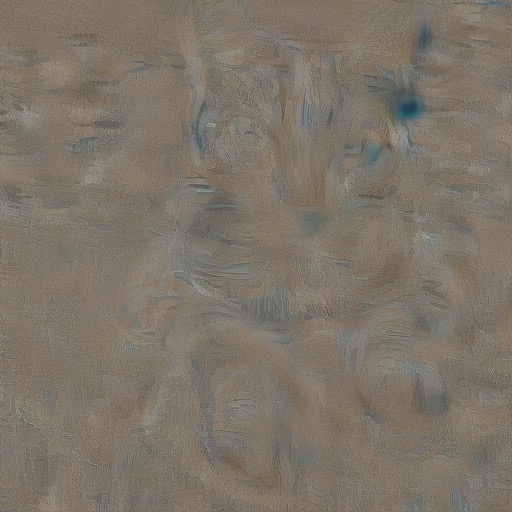} &
        \includegraphics[width=\linewidth]{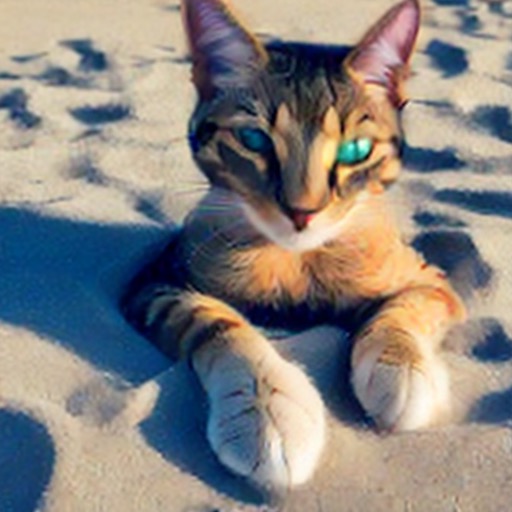}
        \\
        \includegraphics[width=\linewidth]{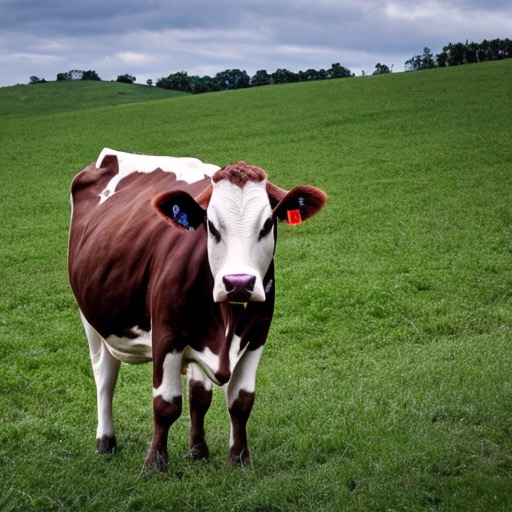} &
        \includegraphics[width=\linewidth]{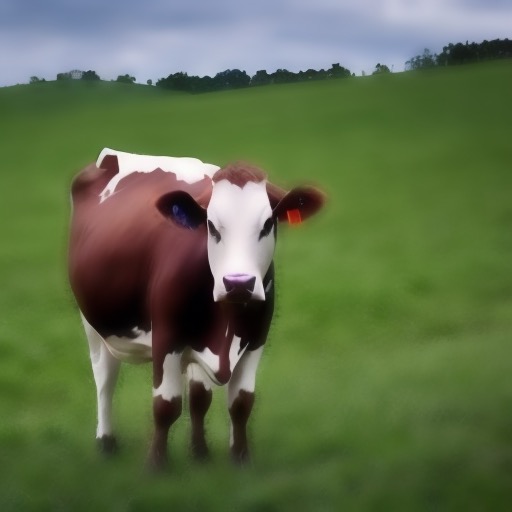} &
        \includegraphics[width=\linewidth]{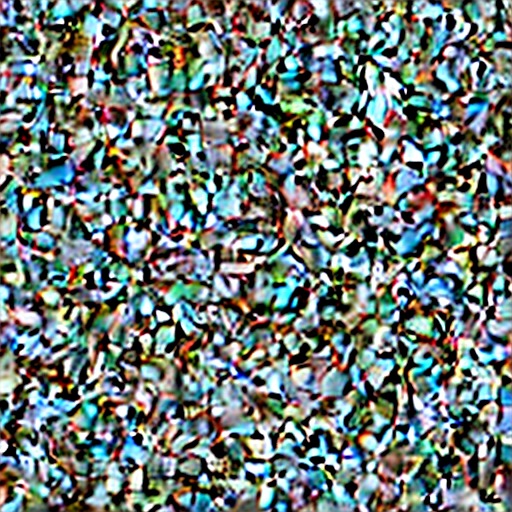} &
        \includegraphics[width=\linewidth]{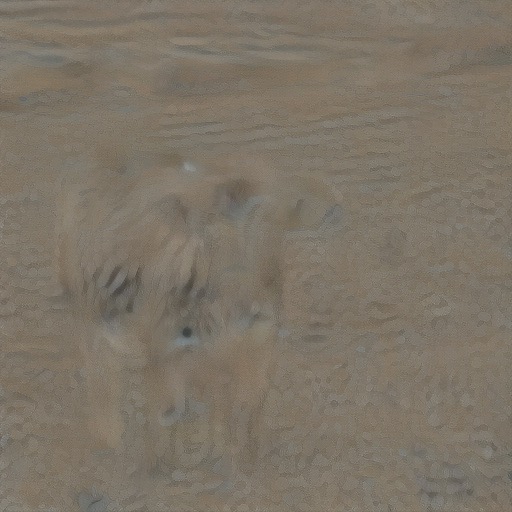} &
        \includegraphics[width=\linewidth]{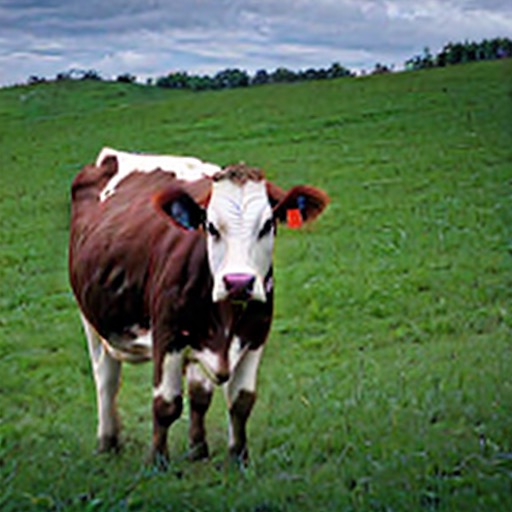}
        \\
        (a) $\x_\textrm{ID}$ & (b) $\x_\textrm{OOD}$  & (c) $\dirn$ & (d) $\dirr$ & (e) $\x_\textrm{OOD}  + \dirr$
    \end{tabular}
    }
    \caption{\label{fig:dr-dn}
        Visualization of $\dirn$ and $\dirr$. Columns (a) and (b) show a pair of  in-domain ($\x_\textrm{ID}$) and out-of-domain ($\x_\textrm{OOD}$) images, both depicting the same underlying content. We add the same noise to both images, and use the pre-trained diffusion model to predict the score. Intuitively, the noised $\x_\textrm{ID}$ image requires no domain correction, and thus the predicted score consists of only $\dirn$, shown in (c). Subtracting $\dirn$ from the prediction for the noised $\x_\textrm{OOD}$ image gives us the domain correction $\dirr$, shown in (d). Indeed, adding $\dirr$ to $\x_\textrm{OOD}$ produces a more realistic image, as shown in (e).}
\end{figure}

We are not aware of any general way to explicitly separate $\epuncond$ into these two components. 
Nevertheless, we attempt to isolate the two components for visualization purposes in \figref{fig:dr-dn}.
The idea is to examine the difference between two unconditional predictions $\epsilon_\phi(\z_{t}(\x_{\textrm{ID}}); \varnothing,t)$ and $\epsilon_\phi(\z_{t}(\x_{\textrm{OOD}}); \varnothing,t)$, where $\z_{t}(\x_{\textrm{ID}})$ and $\z_{t}(\x_{\textrm{OOD}})$ are noised in-domain and out-of-domain images, respectively, that depict the same content and are added the same noise $\epsilon$.
Intuitively, while $\epsilon_\phi(\z_{t}(\x_{\textrm{OOD}}); \varnothing,t)$ both removes noise ($\dirn$) and steers the sample towards the model's domain ($\dirr$), the prediction $\epsilon_\phi(\z_{t}(\x_{\textrm{ID}})$ mostly just removes noise ($\dirn$), since the image is already in-domain. Thus, in \figref{fig:dr-dn} we use the latter to visualize $\dirn$ (column (c)) and the difference between the two predictions to visualize $\dirr$ (column (d)).
As can be seen, $\dirn$ indeed appears to consist of noise uncorrelated with the image content, while $\dirr$ is large in areas where the distortion is most pronounced and adding $\dirr$ to $\x_{\textrm{OOD}}$ effectively enhances the realism of the image (column (e)). More details about this process can be found in the appendix.

\begin{figure}[t]
    \centering
    \setlength{\tabcolsep}{1pt}
    \begin{tabular}{c C{0.135\linewidth} C{0.135\linewidth} C{0.135\linewidth} C{0.135\linewidth} C{0.135\linewidth} C{0.135\linewidth} C{0.135\linewidth}}
        \raisebox{19pt}{\rotatebox{90}{ {\scriptsize$\z_t$}}} &
        \includegraphics[width=\linewidth]{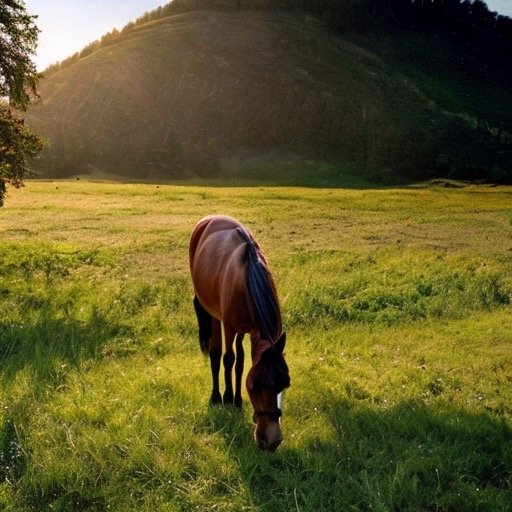} &
        \includegraphics[width=\linewidth]{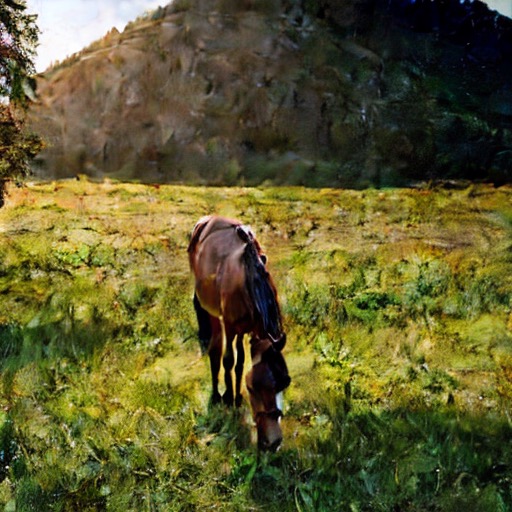} &
        \includegraphics[width=\linewidth]{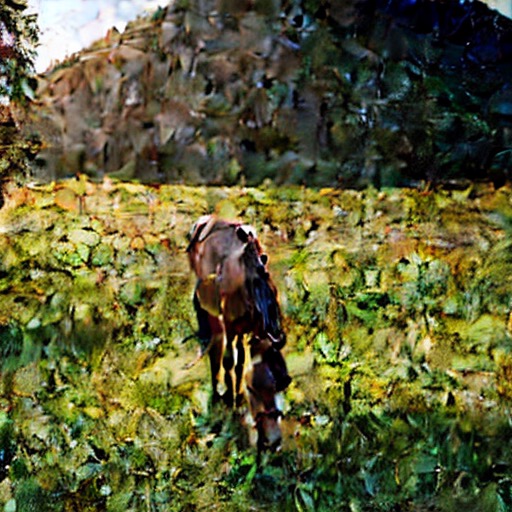} &
        \includegraphics[width=\linewidth]{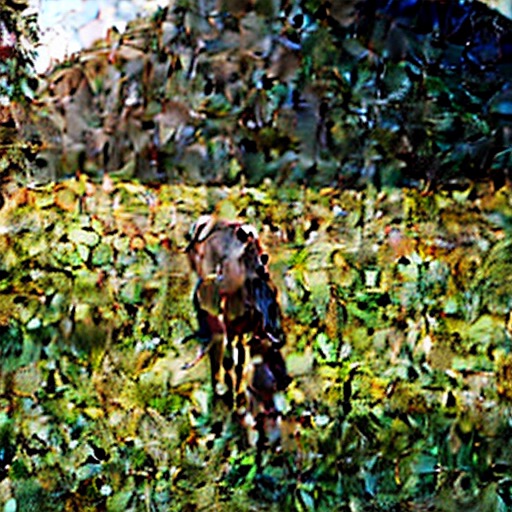} &
        \includegraphics[width=\linewidth]{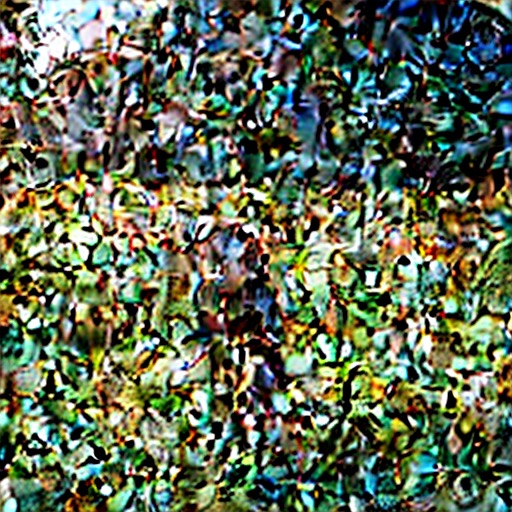} &
        \includegraphics[width=\linewidth]{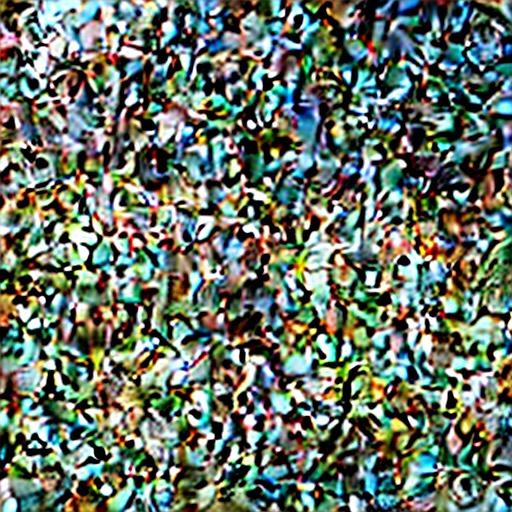} &
        \includegraphics[width=\linewidth]{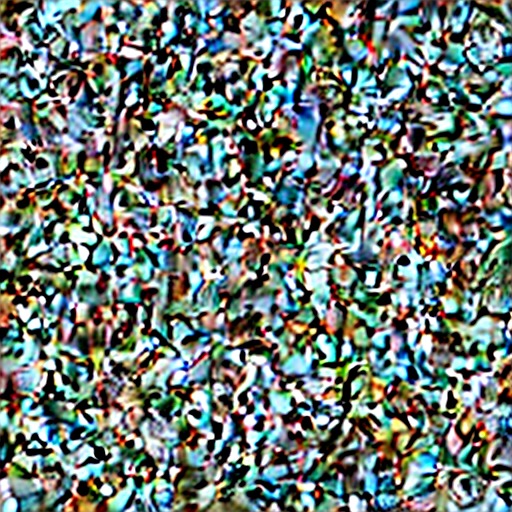} 
        \\
        \raisebox{-1pt}{\rotatebox{90}{ {\scriptsize $\epuncond - \epsilon$}}} &
        \includegraphics[width=\linewidth]{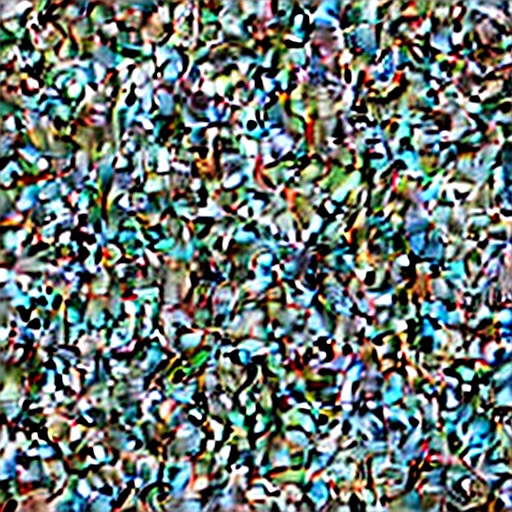} &
        \includegraphics[width=\linewidth]{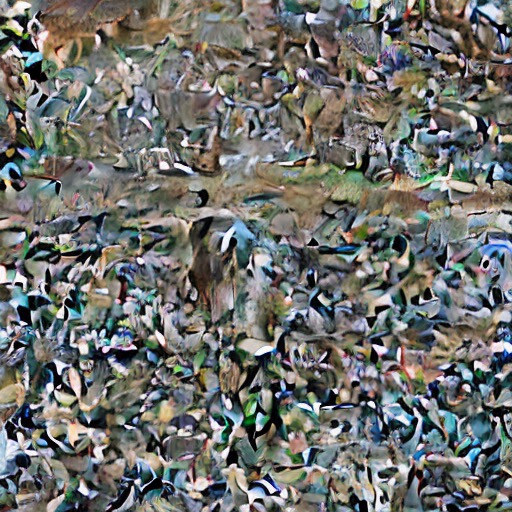} &
        \includegraphics[width=\linewidth]{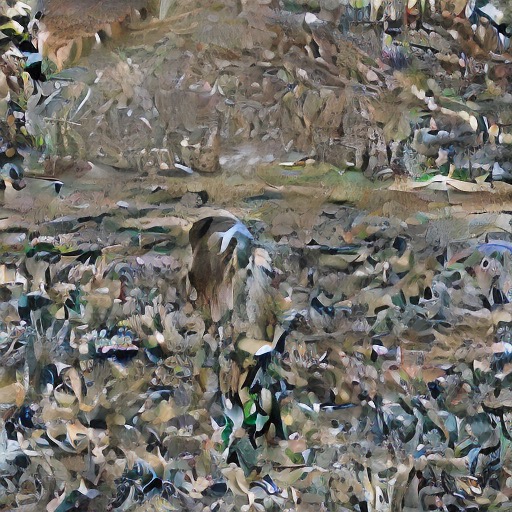} &
        \includegraphics[width=\linewidth]{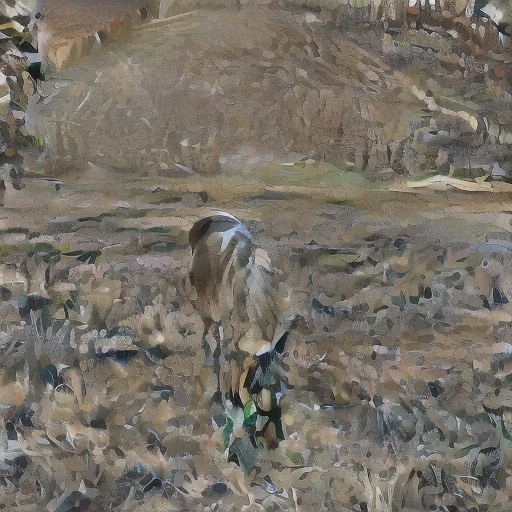} &
        \includegraphics[width=\linewidth]{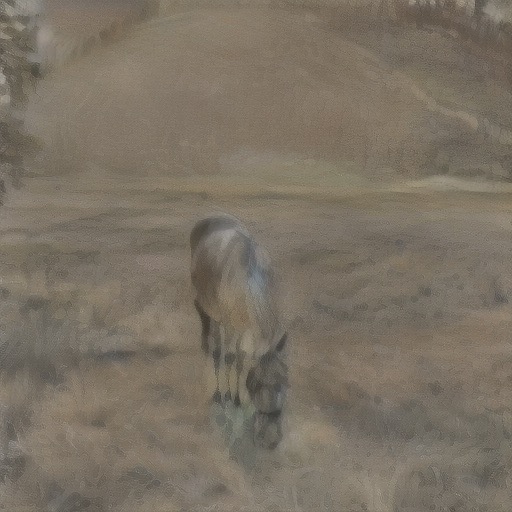} &
        \includegraphics[width=\linewidth]{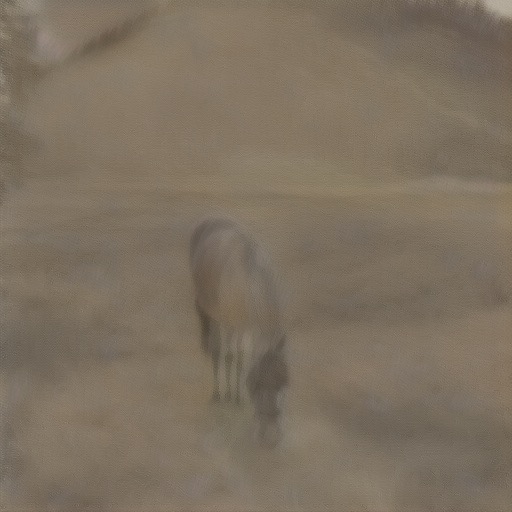} &
        \includegraphics[width=\linewidth]{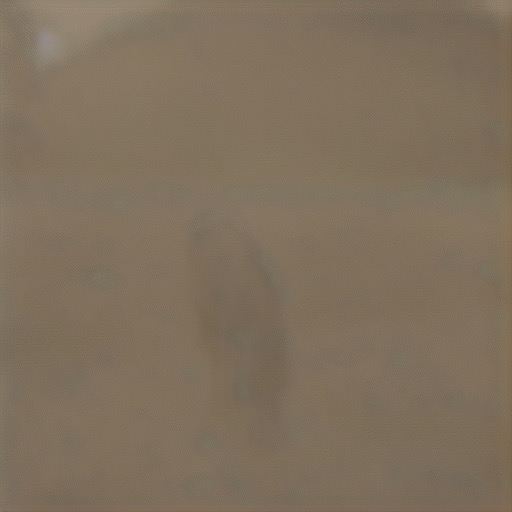} 
        \\
        %
         & {\small $t=1$} & {\small $t=100$} & {\small $t=200$} & {\small $t=300$} & {\small $t=500$} & {\small $t=700$} & {\small $t=1000$}
    \end{tabular}
    \vspace{-8pt}
    \caption{
        Visualization of $\dirn - \epsilon$.
        Top row: noise $\epsilon$ corresponding to different diffusion timesteps $t$ is added to an in-domain image of a horse (as indicated below each column). Bottom row: the residual $\epuncond - \epsilon$ between the network prediction and the actual noise. Since the 
        original image is in-domain (generated by SD), $\dirr \approx 0$, and therefore,
        $\epuncond \approx \dirn$.
        For visualization purposes, the residual is decoded and clamped between -1 and 1. 
        Although we do not expect the residual $\dirn - \epsilon$ to be correlated with the image, it may be seen that some correlation in fact exists, and furthermore, the residual becomes progressively noisier at smaller timesteps $t$.
        }
    
    \vspace{-10pt}
    \label{fig:sds-dr}
\end{figure} 

To summarize so far, using the components discussed above, we can rewrite the CFG score as:
\begin{equation}
    \label{eq:cfg-decomp}
    \epredcfg = \dirr + \dirn + s\dirc.
\end{equation}
\citet{poole2022dreamfusion} define the SDS loss using the difference between the CFG score and the noise $\epsilon$ that was added to the rendered image $\x$ to produce $\z_t$, i.e.,
\begin{equation}
    \label{eq:sds-decomp}
    \grad{\theta} \Loss_\text{SDS} = w(t)(\epredcfg - \epsilon) \frac{\partial \x}{\partial \theta} = w(t)(\dirr + \dirn + s\dirc - \epsilon) \frac{\partial \x}{\partial \theta}.    
\end{equation}
Note that while both $\dirr$ and $\dirc$ are needed to steer the rendered image towards an in-domain image aligned with the condition $y$, the residual $\dirn - \epsilon$ is generally non-zero and noisy, and this issue becomes increasingly pronounced when smaller time steps, responsible for the formation of fine details, are employed, as visualized in \figref{fig:sds-dr}.
This residual may explain, in part, the lower quality images generated using SDS, compared to ancestral sampling: at each optimization step, the optimized parameters $\theta$ are guided at a random direction depending on $\dirn - \epsilon$, resulting in an averaging effect. While higher level semantics are roughly less affected, fine and medium level details, from lower diffusion times $t$ tend to be over-smoothed. Previous works \citep{hertz2023delta, wang2023prolificdreamer} have also observed that the subtraction of $\epsilon$ indeed leads to blurry results.

\begin{figure}
    \centering
    \setlength{\tabcolsep}{1pt}
    {\small
    \begin{tabular}{C{0.155\linewidth} C{0.155\linewidth} C{0.155\linewidth} C{0.155\linewidth} C{0.155\linewidth} C{0.155\linewidth}}
        \includegraphics[width=\linewidth]{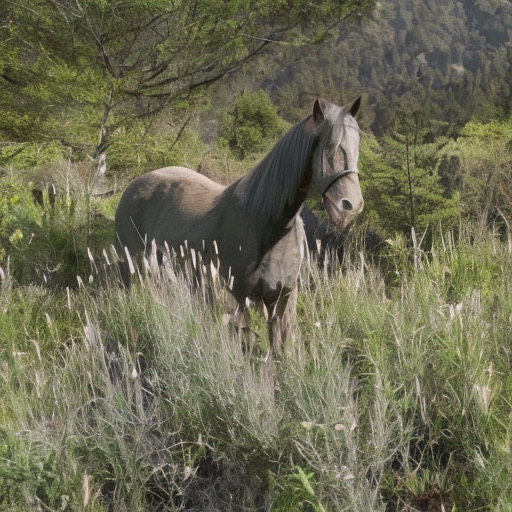} &
        \includegraphics[width=\linewidth]{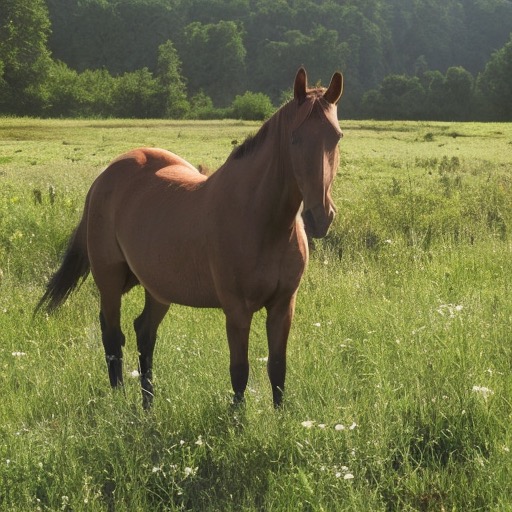} &
        \includegraphics[width=\linewidth]{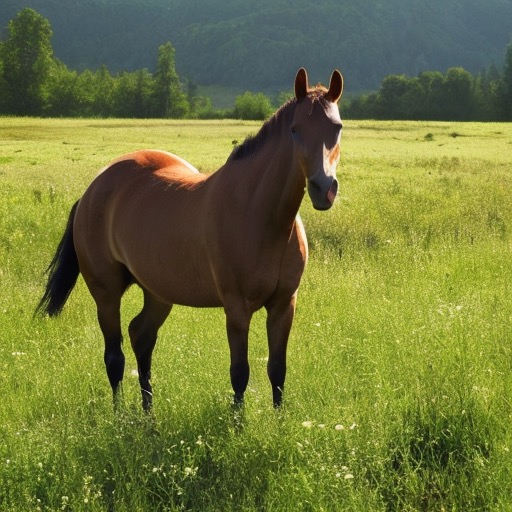} &
        \includegraphics[width=\linewidth]{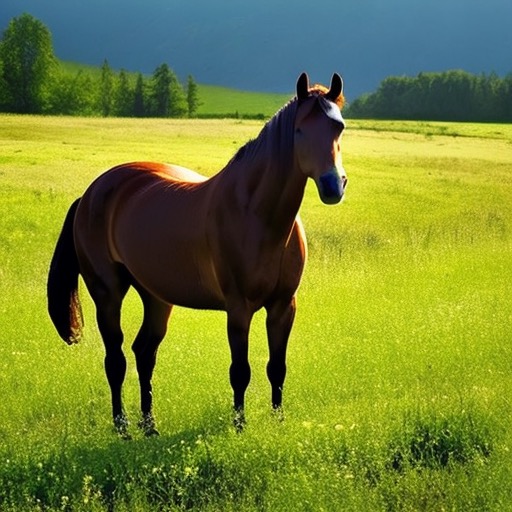} &
        \includegraphics[width=\linewidth]{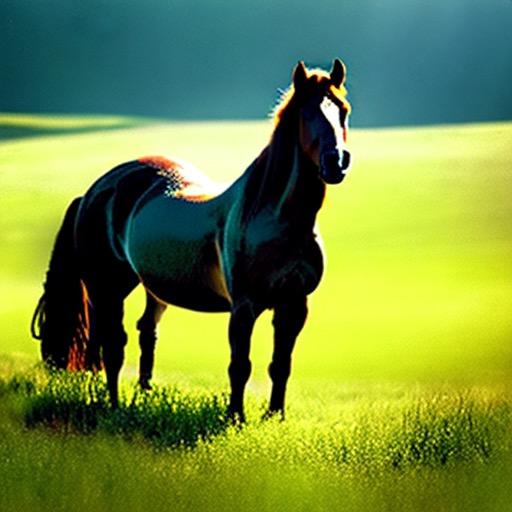} &
        \includegraphics[width=\linewidth]{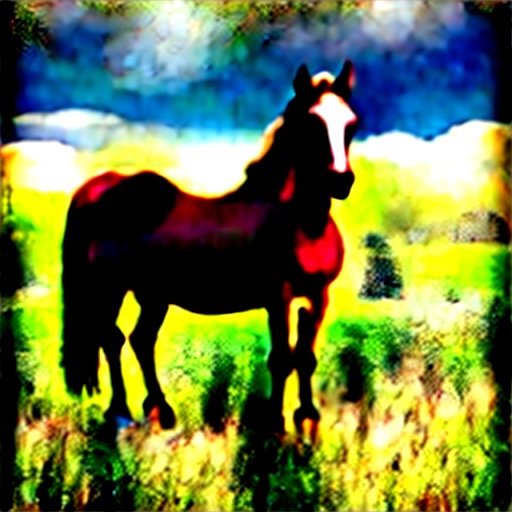} 
        \\
        $s = 1$ &
        $s = 4$ &
        $s = 7.5$ &
        $s = 15$ &
        $s = 30$ &
        $s = 100$ 
        \\
    \end{tabular}
    }
    
    \caption{
The impact of CFG on ancestral sampling. We generate all images using the same prompt ``A photo of a horse in a meadow'' with the same seed and different values of the CFG parameter $s$. As can be seen, large values of $s$ lead to over-saturated and less realistic results.
    }
    \vspace{-6pt}
    \label{fig:diff_cfg}
\end{figure}

Importantly, our decomposition in \Eqref{eq:sds-decomp} explains the need for using a large CFG coefficient in SDS (e.g., $s=100$), as this enables the image-correlated $s\dirc$ term to dominate the loss, making the noisy residual $\dirn - \epsilon$ relatively negligible. However, high CFG coefficients are known to yield less realistic and less diverse results, as demonstrated in \figref{fig:diff_cfg}, typically leading to over-saturated images and NeRFs.

\vspace{-5pt}
\section{Noise Free Score Distillation}
\label{sec:nfsd}
\vspace{-4pt}

As discussed above, ideally only the $s\dirc$ and the $\dirr$ terms should be used to guide the optimization of the parameters $\theta$.
While $\dirc$ is simply the difference between the conditioned and the null-conditioned predictions, $\dirr$ is more challenging to separate from $\dirn$, as they are both part of the predicted noise $\epuncond$.

To extract $\dirr$, we distinguish between different stages in the backward (denoising) diffusion process. First, note that the noise variance, i.e., the magnitude of the noise to be removed, is monotonically decreased in the backward process. Thus, for sufficiently small timestep values $t$, $\dirn$ is rather small, and the score $\epuncond = \dirn + \dirr$ is increasingly dominated by $\dirr$.
Specifically, in our experiments, we find that for $t<200$ the noise is small enough to be negligible, implying that $\epuncond \approx \dirr$.

As for the larger timestep values, $t \geq 200$, we propose to approximate $\dirr$ by the difference $\epuncond - \epneg$, where $\pneg =$ ``unrealistic, blurry, low quality, out of focus, ugly, low contrast, dull, dark, low-resolution, gloomy''. 
Here, we are making the assumption that $\delta_{C=\pneg} \approx -\dirr$, and thus $\epuncond - \epneg = \dirr + \dirn - (\dirr + \dirn + \delta_{C = \pneg}) \approx \dirr$.

To conclude, we approximate $\dirr$ by
\begin{equation}
    \dirr = 
    \begin{cases}
            \epuncond , & \text{if } t < 200 \\
            \epuncond - \unet(\z_t; y=\pneg,t), & \text{otherwise,}
    \end{cases}
\end{equation}
and use the resulting $\dirc$ and $\dirr$ to define an alternative, \emph{noise-free score distillation} loss $\Loss_\text{NFSD}$, whose gradients are used to optimize the parameters $\theta$, instead of $\grad{\theta}\Loss_\text{SDS}$:
\begin{equation}\label{eq:nfsd}
    \grad{\theta} \Loss_\text{NFSD} = w(t) \left(\dirr + s\dirc \right) \frac{\partial \x}{\partial \theta},
\end{equation}

In \figref{fig:2d_comparison} and \figref{fig:3d_comparison} we show that these seemingly small changes in the definition of the loss lead to a noticeable improvement in the quality of generated images, as well as NeRFs. Note, that while in the results of SDS we use $s = 100$, in our results we use the commonly used value of $s = 7.5$.
The reason for this is that by taking the measures described above to approximately eliminate the $\dirn$ component, it is no longer necessary to resort to a large value of $s$ to make the $\dirc$ term dominant.

 \begin{figure}
        \centering
    \setlength{\tabcolsep}{1pt}
    \begin{tabular}{c c c c c c c c c}
        \includegraphics[width=0.12\linewidth]{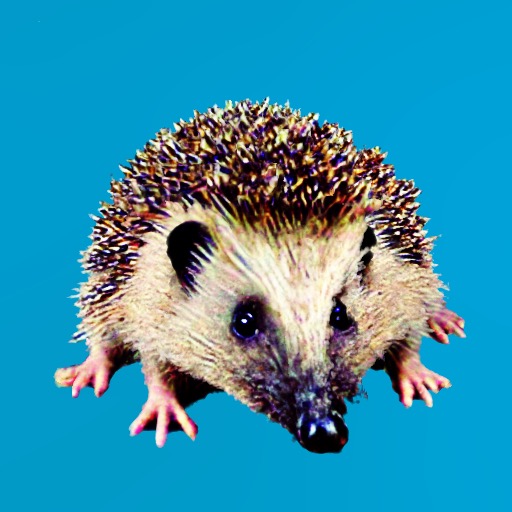} &
        \includegraphics[width=0.12\linewidth]{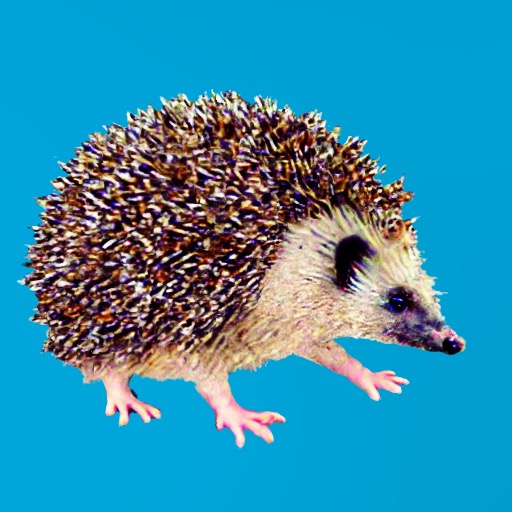} &
        \raisebox{0.51\height}{%
        \begin{tabular}[b]{c}
            \adjustbox{valign=t}{\includegraphics[width=0.06\linewidth]{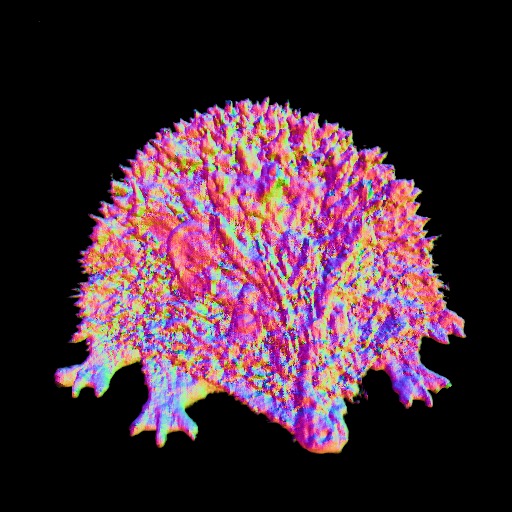}} \\
            \adjustbox{valign=t}{\includegraphics[width=0.06\linewidth]{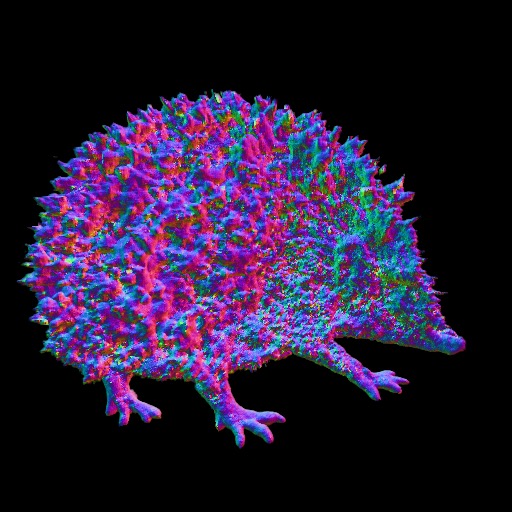}}
        \end{tabular}
        } &
        \includegraphics[width=0.12\linewidth]{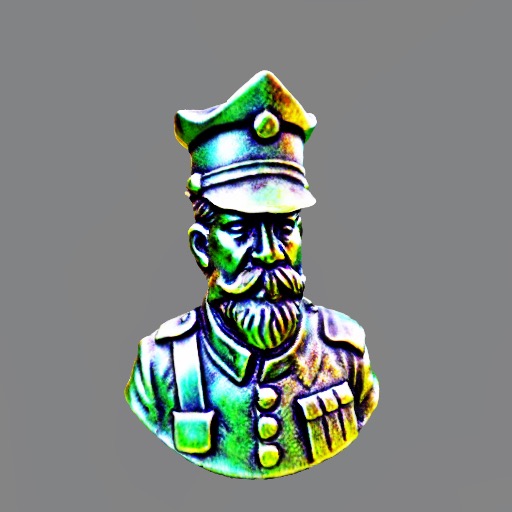} &
        \includegraphics[width=0.12\linewidth]{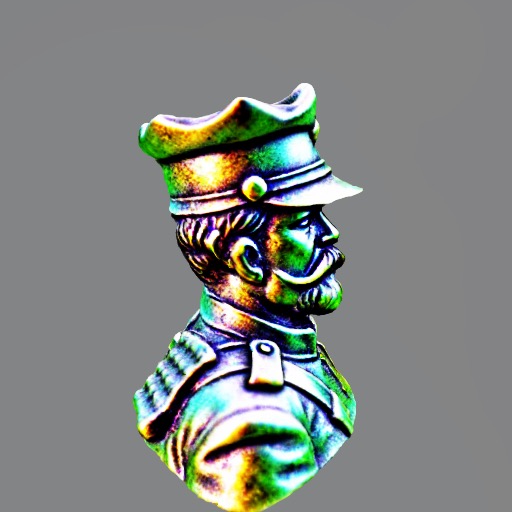} &
        \raisebox{0.51\height}{%
        \begin{tabular}[b]{c}
            \adjustbox{valign=t}{\includegraphics[width=0.06\linewidth]{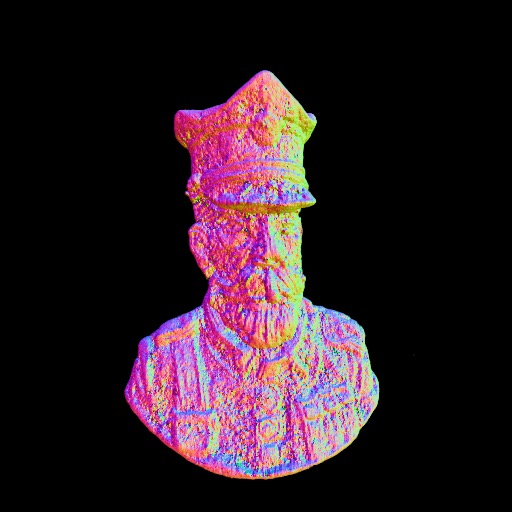}} \\
            \adjustbox{valign=t}{\includegraphics[width=0.06\linewidth]{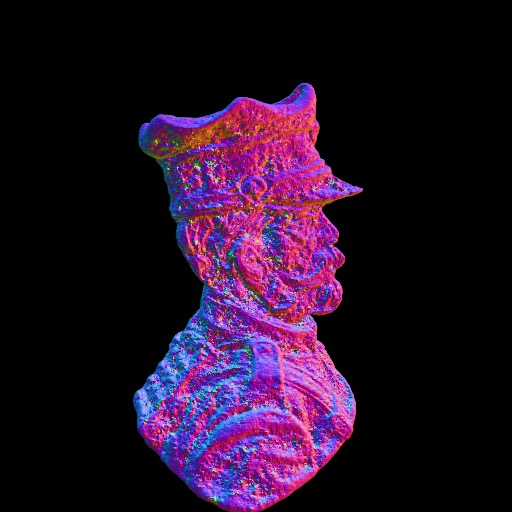}}
        \end{tabular}
        }  &
        \includegraphics[width=0.12\linewidth]{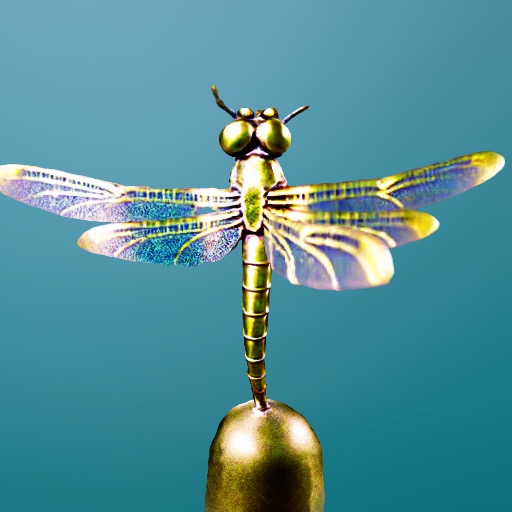} &
        \includegraphics[width=0.12\linewidth]{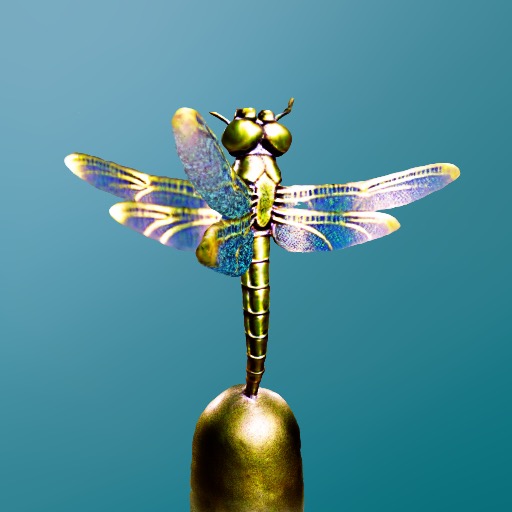} &
        \raisebox{0.51\height}{%
        \begin{tabular}[b]{c}
            \adjustbox{valign=t}{\includegraphics[width=0.06\linewidth]{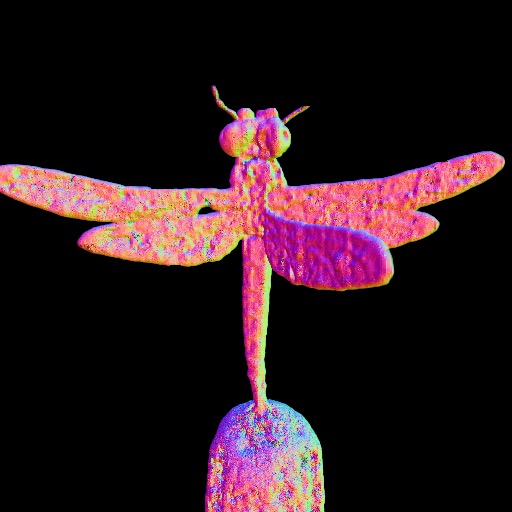}} \\
            \adjustbox{valign=t}{\includegraphics[width=0.06\linewidth]{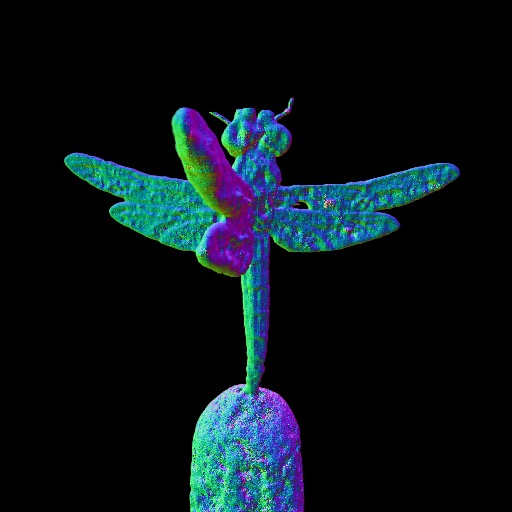}}
        \end{tabular}
        }
        \\
        \includegraphics[width=0.12\linewidth]{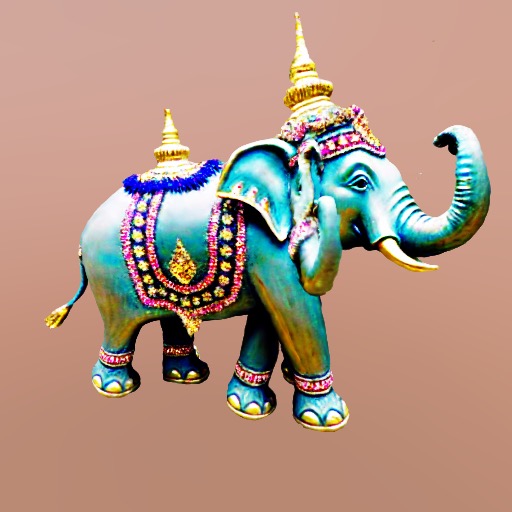} &
        \includegraphics[width=0.12\linewidth]{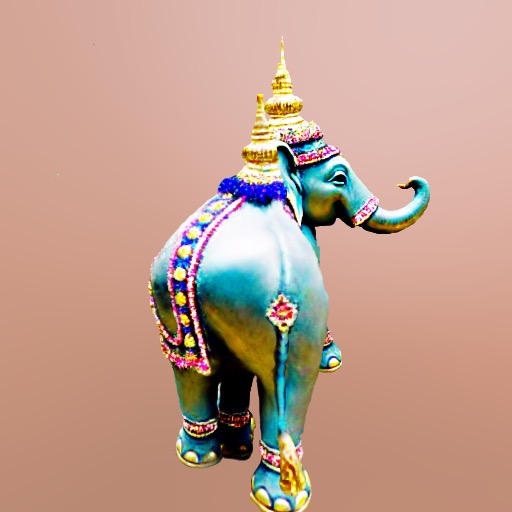} &
        \raisebox{0.51\height}{%
        \begin{tabular}[b]{c}
            \adjustbox{valign=t}{\includegraphics[width=0.06\linewidth]{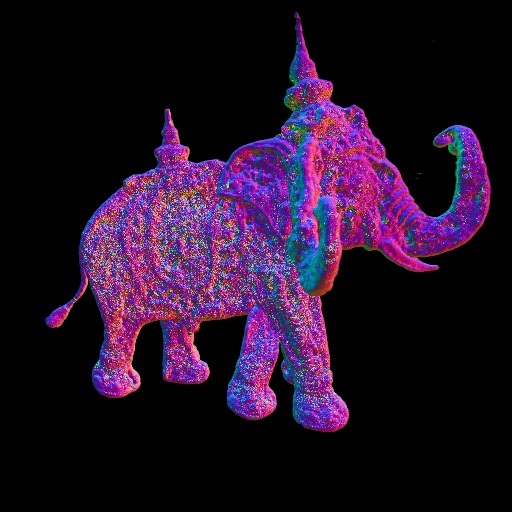}} \\
            \adjustbox{valign=t}{\includegraphics[width=0.06\linewidth]{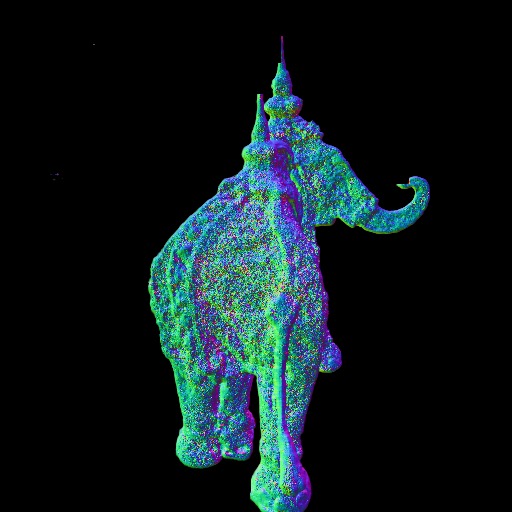}}
        \end{tabular}
        } &
        \includegraphics[width=0.12\linewidth]{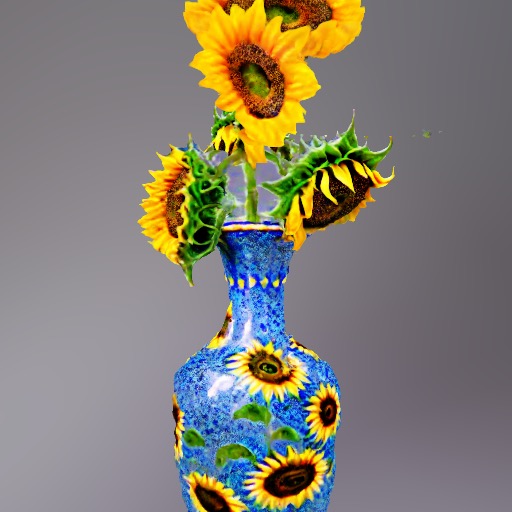} &
        \includegraphics[width=0.12\linewidth]{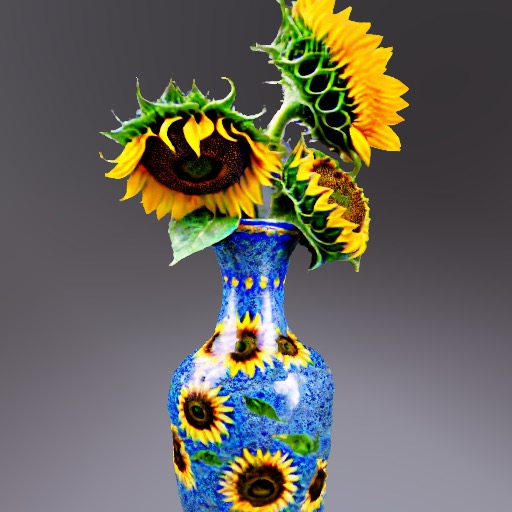} &
        \raisebox{0.51\height}{%
        \begin{tabular}[b]{c}
            \adjustbox{valign=t}{\includegraphics[width=0.06\linewidth]{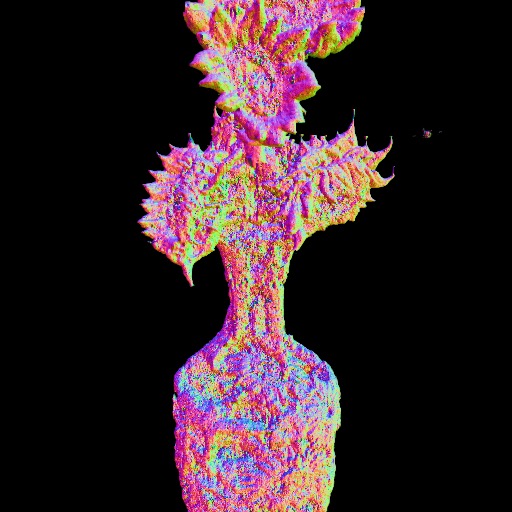}} \\
            \adjustbox{valign=t}{\includegraphics[width=0.06\linewidth]{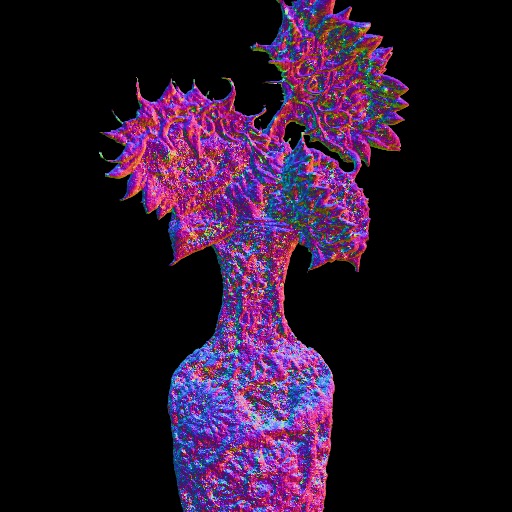}}
        \end{tabular}
        } &
        \includegraphics[width=0.12\linewidth]{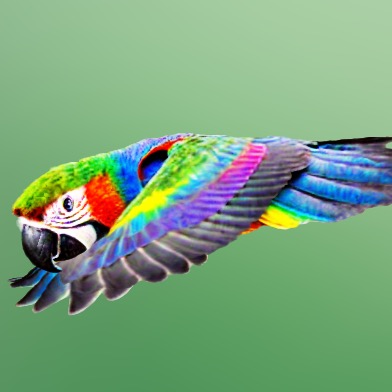} &
        \includegraphics[width=0.12\linewidth]{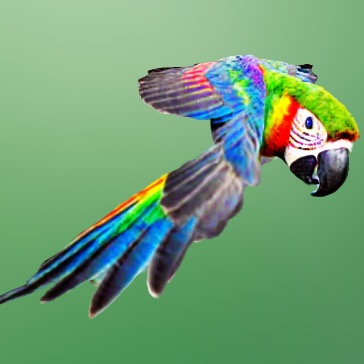} &
        \raisebox{0.51\height}{%
        \begin{tabular}[b]{c}
            \adjustbox{valign=t}{\includegraphics[width=0.06\linewidth]{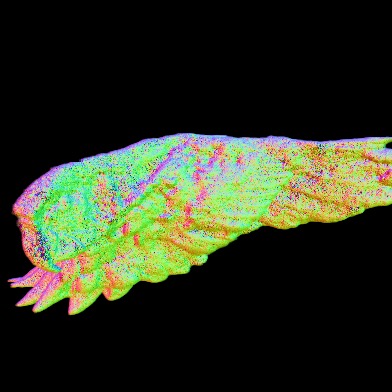}} \\
            \adjustbox{valign=t}{\includegraphics[width=0.06\linewidth]{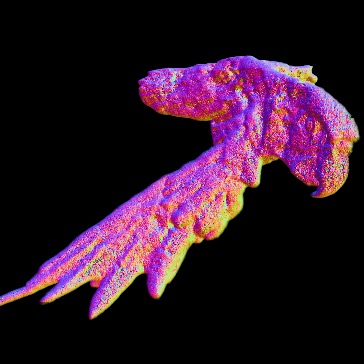}}
        \end{tabular}
        } 

    \end{tabular}

    \caption{NeRFs optimized with NFSD.}
    \vspace{-11pt}
    \label{fig:3d-generation}
\end{figure}

\vspace{-5pt}
\section{Discussion}
\vspace{-5pt}
\label{sec:discussion}

Our score decomposition formulation can be used to explain previous works that were proposed to improve the SDS loss. This demonstrates the wide scope and applicability of our formulation.

\vspace{-6pt}

\paragraph{DDS.} \citet{hertz2023delta} propose an adaptation of the SDS loss for image editing task. Specifically, instead of randomly initializing the optimization process as in SDS, it is initialized with the (in-domain) input image. DDS optimizes the input image according to the text condition, while preserving image attributes that are irrelevant to the edit task guided by the input prompt. The gradients used by DDS are defined by
\begin{equation}
    \label{eq:dds}
    \grad{\theta} \Loss_\text{DDS} = 
    \grad{\theta} \Loss_\text{SDS}(\z_t(\x), y) - 
    \grad{\theta} \Loss_\text{SDS}(\Tilde{\z}_t(\Tilde{\x}), \Tilde{y}),
\end{equation}
where $\Tilde{\z}_t(\Tilde{\x}), \Tilde{y}$ denote the noisy original input image and its corresponding prompt, respectively. Here $y$ denotes the prompt that describes the edit, and $\x, \Tilde{\x}$ are noised with the same noise $\epsilon$. Incorporating our score decomposition into \Eqref{eq:dds} yields
\begin{equation}
    \grad{\theta} \Loss_\text{DDS} = 
    w(t)(\dirr + \dirn + s\delta_{C_\text{edit}} - \epsilon) - w(t)(\dirr + \dirn + s\delta_{C_\text{orig}} - \epsilon) = w(t) s(\delta_{C_\text{edit}} - \delta_{C_\text{orig}}).
\end{equation}
Our formulation helps to understand the high-quality results achieved by DDS: the residual component which makes the results in SDS over-smoothed and over-saturated is cancelled out. Moreover, since the optimization is initialized with an in-domain image, the $\dirr$ component is not effectively needed and cancelled out. The remaining direction is the one relevant to the difference between the original prompt and the new one.

\vspace{-6pt}

\paragraph{ProlificDreamer.} \citet{wang2023prolificdreamer} tackle the generation task, and propose the VSD loss, which successfully alleviates the over-smoothed and over-saturated results obtained by SDS. In VSD, alongside the pretrained diffusion model $\unet$, another diffusion model $\epsilon_\text{LoRA}$ is trained during the optimization process.  The $\epsilon_\text{LoRA}$ model is initialized with the weights of $\unet$, and during the optimization process it is fine-tuned with rendered images $\x = g(\theta)$. Effectively, the rendered images during the optimization are out-of-domain for the original pretrained model distribution, but are in-domain for $\epsilon_\text{LoRA}$.
Hence, the gradients of the VSD loss are defined as
\begin{equation}
    \label{eq:vsd}
    \grad{\theta} \Loss_\text{VSD} = w(t) \left(\epredcfgx - \epsilon_\text{LoRA}(\z_t(\x);y,t,c \right) \frac{\partial \x}{\partial \theta},
\end{equation}
where $c$ is another condition that is added to $\epsilon_\text{LoRA}$ and represents the camera viewpoint of the rendered image $\x$. Viewed in terms of our score decomposition, since $\epsilon_\text{LoRA}$ is fine-tuned on $\x$, both $\dirc$ and $\dirr$ are approximately 0, thus it simply predicts $\dirn$.
Therefore, $\grad{\theta} \Loss_\text{VSD}$ can be written as
\begin{equation}
    \grad{\theta} \Loss_\text{VSD} = 
    w(t) (\dirr + \dirn + s\dirc - \dirn) \frac{\partial \x}{\partial \theta} = 
    w(t) (\dirr + s\dirc) \frac{\partial \x}{\partial \theta},
\end{equation}
i.e., it approximates exactly the same terms as our NFSD. It should be noted that unlike our approach, VSD has a considerable computational overhead of fine-tuning the additional diffusion model during the optimization process.

\vspace{-4pt}
\section{Experiments}
\label{sec:experiments}
\vspace{-4pt}

We implement NFSD using the threestudio \citep{threestudio2023} framework for text-based 3D generation. 
Unless stated otherwise, all 3D models are optimized for $25,000$ iterations using AdamW optimizer~\citep{loshchilov2017adamw} with a learning rate of $0.01$. The initial rendering resolution of $64 \times 64$ is increased to $512 \times 512$ after $5,000$ iterations; at the same time we anneal the maximum diffusion time to $500$ as proposed by \citet{lin2023magic3d, wang2023prolificdreamer}. The implicit volume is initialized according to the object-centric initialization \citep{lin2023magic3d, wang2023prolificdreamer}. We alternate the background between random solid-color and a learned neural environment map. The pre-trained text-to-image diffusion model for all experiments is Stable Diffusion 2.1-base~\citep{rombach2022high}, a latent diffusion model with $\epsilon$-prediction.

\textbf{3D generation.}
\figref{fig:3d-generation} showcases several NeRFs optimized using our NFSD. As can be seen the rendered images are sharp and contain highly intricate details. The prompts used and additional examples can be found in Appendices~\ref{sec:app-prompts} and~\ref{sec:app-results}, respectively.

\textbf{Comparison with SDS.}
We compare our NFSD with SDS under different parametric generators and different configurations. In each comparison, the same seed is used by both methods. Specifically, we set the same seed for the guidance process, i.e., the noise and the diffusion time are the same for both methods, at every step of the optimization process.

\begin{figure}
    \centering
    \setlength{\tabcolsep}{1pt}
    {\small
    \begin{tabular}{c C{0.155\linewidth} C{0.155\linewidth} C{0.155\linewidth} C{0.155\linewidth} C{0.155\linewidth}}
        \raisebox{15pt}{\rotatebox{90}{ SDS $7.5$}} &
        \includegraphics[width=\linewidth]{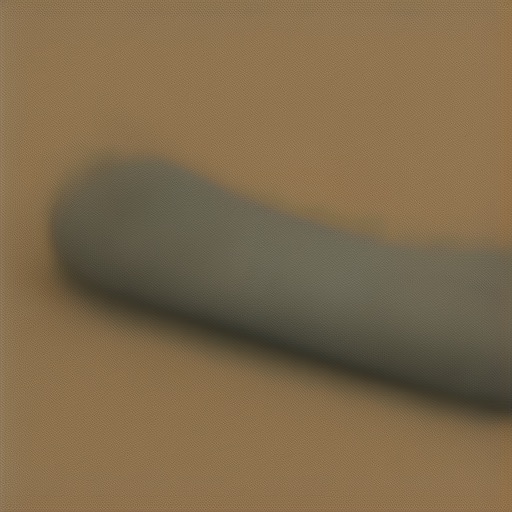} &
        \includegraphics[width=\linewidth]{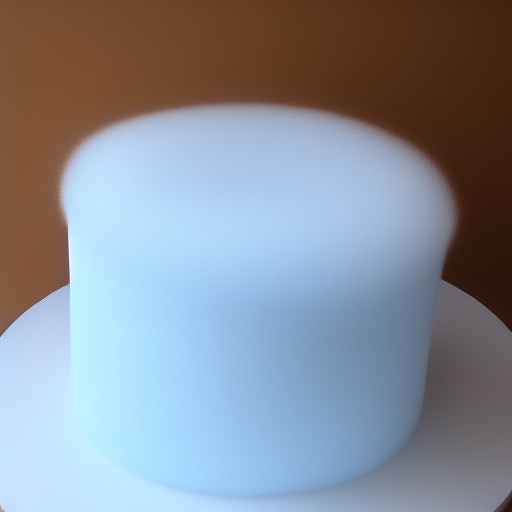} &
        \includegraphics[width=\linewidth]{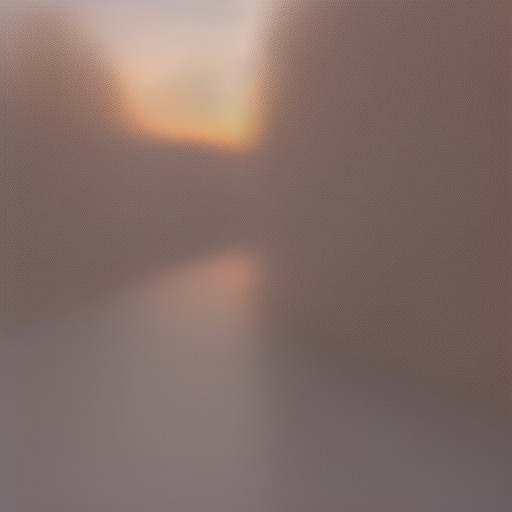} &
        \includegraphics[width=\linewidth]{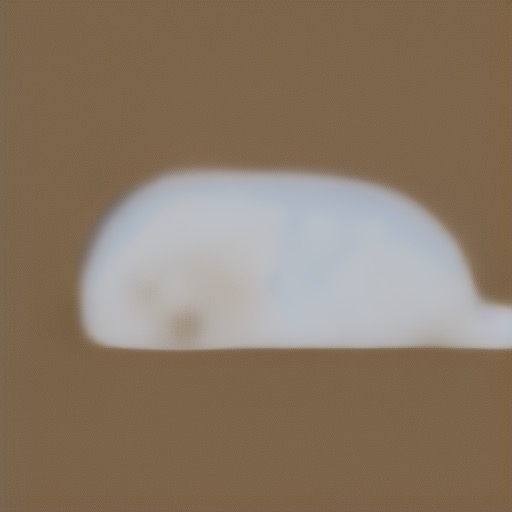} &
        \includegraphics[width=\linewidth]{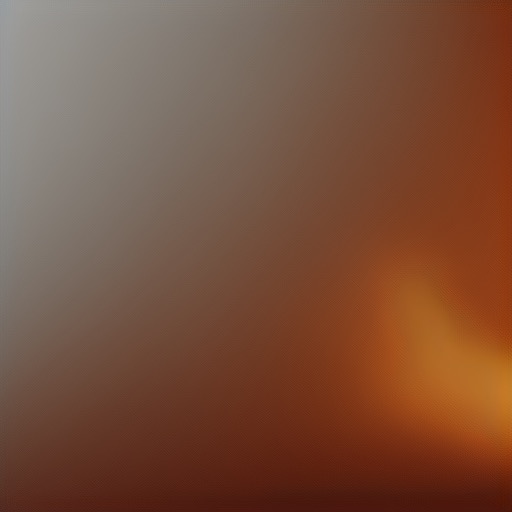}
        \\
        \raisebox{15pt}{\rotatebox{90}{ SDS $100$}} &
        \includegraphics[width=\linewidth]{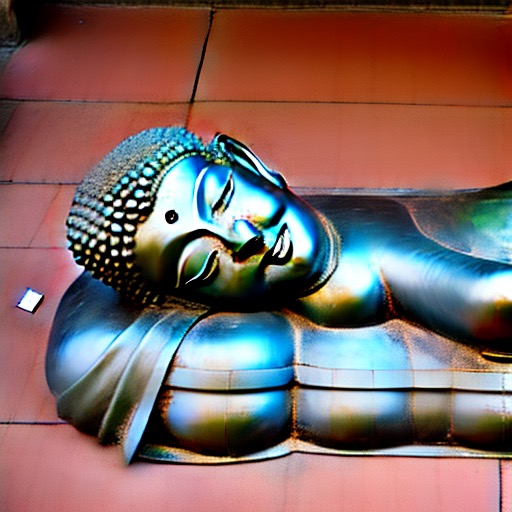} &
        \includegraphics[width=\linewidth]{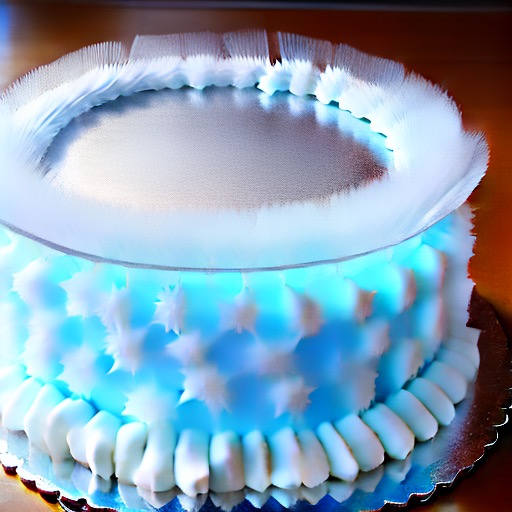} &
        \includegraphics[width=\linewidth]{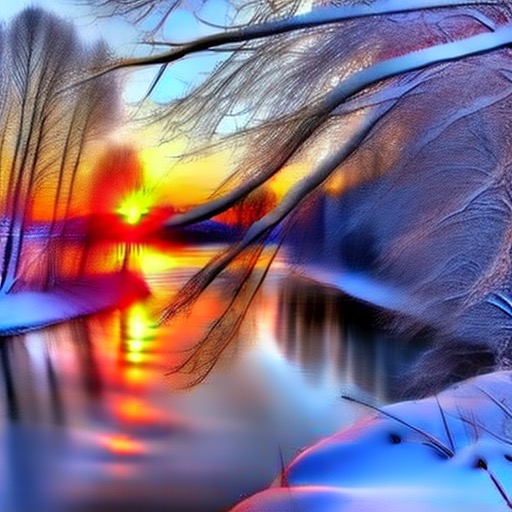} &
        \includegraphics[width=\linewidth]{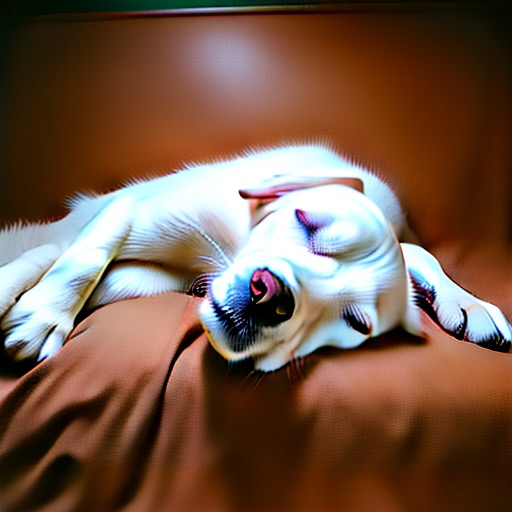} &
        \includegraphics[width=\linewidth]{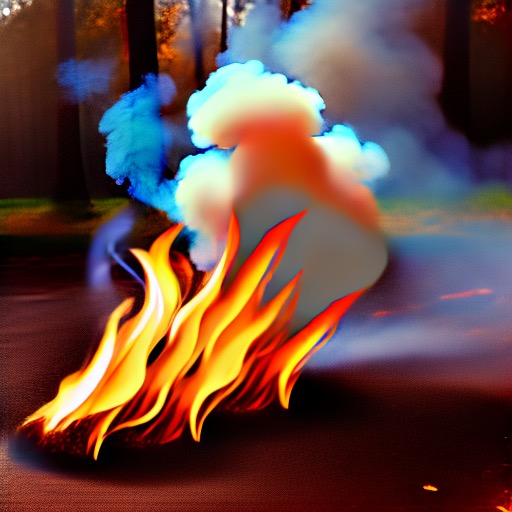}
        \\
        \raisebox{15pt}{\rotatebox{90}{ NFSD $7.5$}} &
        \includegraphics[width=\linewidth]{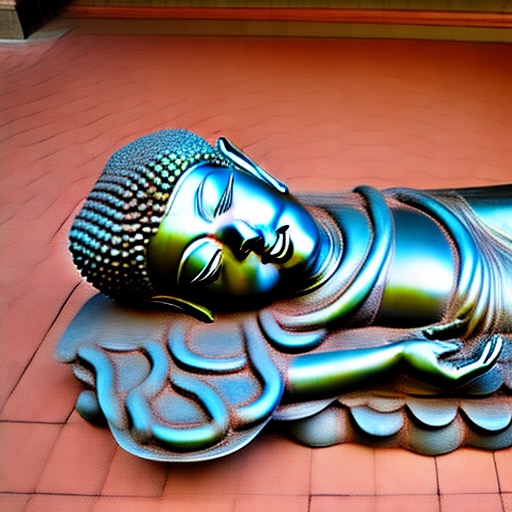} &
        \includegraphics[width=\linewidth]{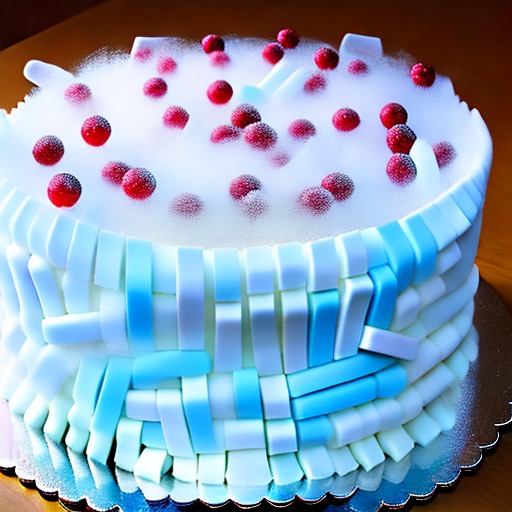} &
        \includegraphics[width=\linewidth]{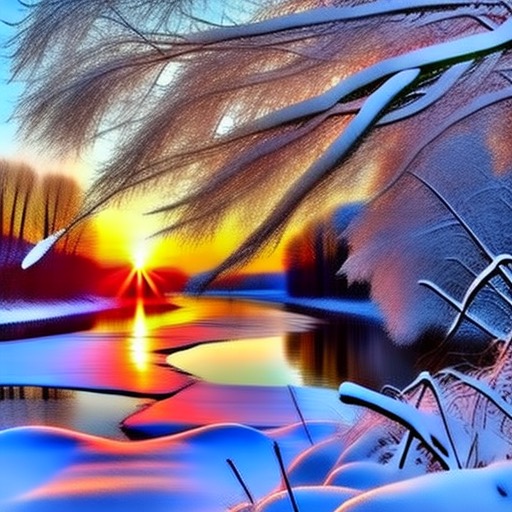} &
        \includegraphics[width=\linewidth]{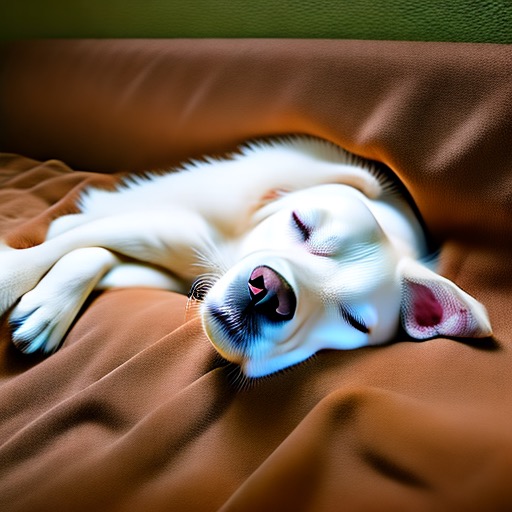} &
        \includegraphics[width=\linewidth]{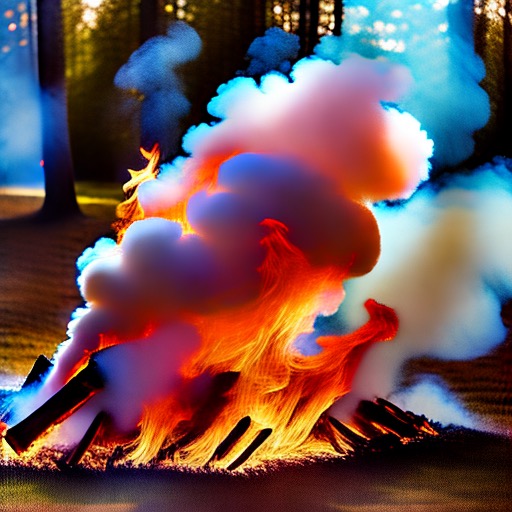}
        \\
        &
        {\scriptsize ``A metal lying Buddha'' }&
        {\scriptsize ``A cake made of ice'' }&
        {\scriptsize ``A photo of sunset, winter, river'' } &
        {\scriptsize ``A white dog sleeping'' }&
        {\scriptsize ``A fire with smoke'' }
        \\
    \end{tabular}
    }
    \vspace{-6pt}
    \caption{
2D image generation with SDS and NFSD. We directly optimize the latent space of SD-2.1-base \citep{rombach2022high}. Top row: SDS with CFG of 7.5 generates overly-smooth images that severely lack detail. The common solution is to increase the CFG scale to 100 (middle row). The high CFG enables a reasonable distillation of the score, but some artifacts remain (e.g., the dog's purple eyes, the frosting in the cake) and realistic details are still lacking (e.g., the fire). Bottom row: using the standard CFG of 7.5, our NFSD succeeds in distilling finer details, such as the frosting on the cake or the fire example.
    }
    \vspace{-20pt}
    \label{fig:2d_comparison}
\end{figure}

\emph{2D image generation:}
Here, we directly optimize the latent code of Stable Diffusion, a $64 \times 64 \times 4$ tensor. In the notations defined in \Secref{sec:sd}, $\theta \in \mathbb{R}^{64 \times 64 \times 4}$ and $g(\theta) = \theta$. To initialize the optimization process, $\theta$ is sampled from a Gaussian Distribution.
We then use either $\Loss_\textrm{NFSD}$ or $\Loss_\textrm{SDS}$ as the only loss for $1,000$ iterations. As illustrated in \figref{fig:2d_comparison}, SDS optimization with a nominal CFG scale ($7.5$) yields over-smoothed images, while using a high CFG scale ($100$) generates the main object but lacks background details and occasionally introduces aritfacts. 
In contrast, NFSD optimization is able to produce more pleasing results in which the object is clear, the background is detailed, and the image looks more realistic, even when using a CFG value of $7.5$.
For example, observe the fur of the dog which contains artifacts in the SDS-$100$ configuration, and looks more realistic with NFSD-$7.5$. In addition, observe the fire flames which feature fewer details and therefore look painted with SDS-$100$, and exhibit more detail and realism when using NFSD-$7.5$.

\emph{Text-to-NeRF synthesis:}
In \figref{fig:3d_comparison} we use SDS and NFSD to optimize NeRFs given text prompts. Here we do not preform time annealing of the diffusion, which hinders a bit the quality of the output model, but allows better emphasis on the differences between SDS and NFSD. As can be seen, although SDS succeeds in generating plausible 3D objects, it typically generates fewer fine details. Note for example the wings of the eagle and the mane of the horse.
Note that, unlike images, the NeRF representation is inherently smooth; furthermore, additional losses, such as shape density, are applied. These regularizations, combined with a high CFG value, enable SDS to produce plausible results, despite the unwanted noise distillation. In contrast, NFSD produces more detailed results with ordinary CFG values by attempting to eliminate the noise and explicitly approximate $\dirr$.

\input{figures/3d_comparison}

\begin{figure}
    \centering
    \setlength{\tabcolsep}{1pt}
    {\small
    \begin{tabular}{C{0.145\linewidth} C{0.145\linewidth} C{0.145\linewidth} C{0.145\linewidth} C{0.145\linewidth} C{0.145\linewidth}}
        \includegraphics[width=\linewidth]{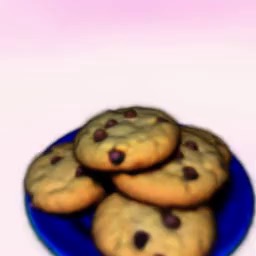} &
        \includegraphics[width=\linewidth]{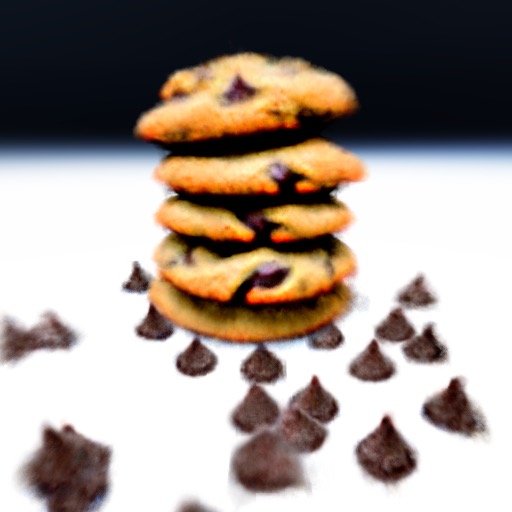} &
        \includegraphics[width=\linewidth]{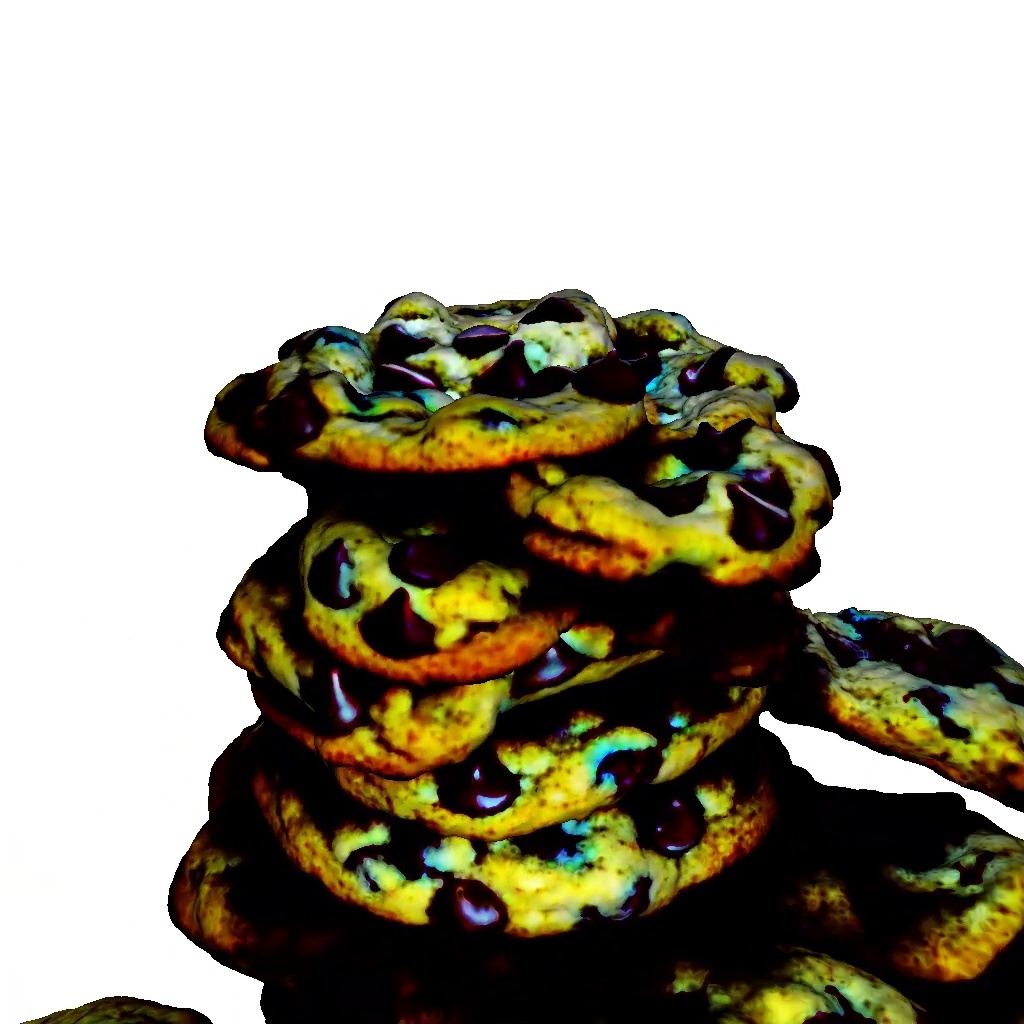} &
        \includegraphics[width=\linewidth]{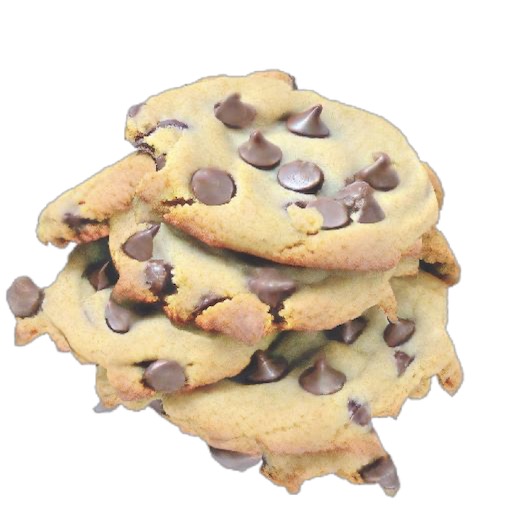} &
        \includegraphics[width=\linewidth]{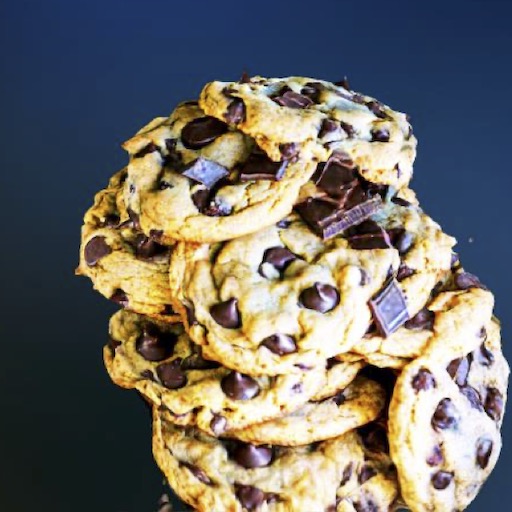} &
        \includegraphics[width=\linewidth]{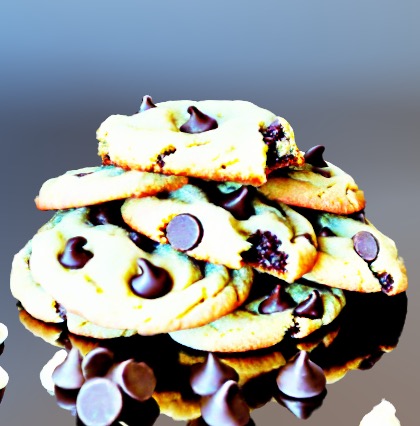} 
        \\
        \includegraphics[width=\linewidth]{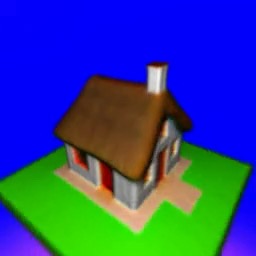} &
        \includegraphics[width=\linewidth]{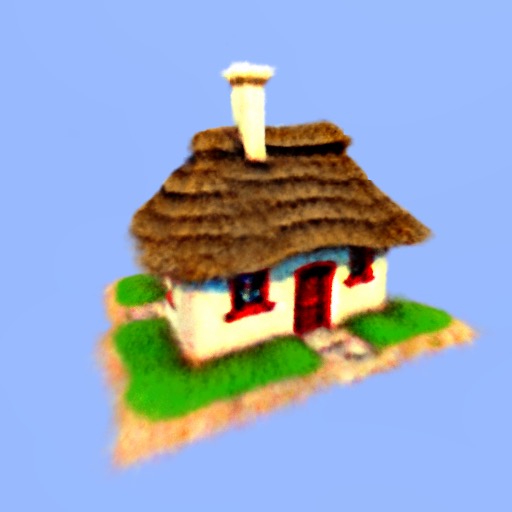} &
        \includegraphics[width=\linewidth]{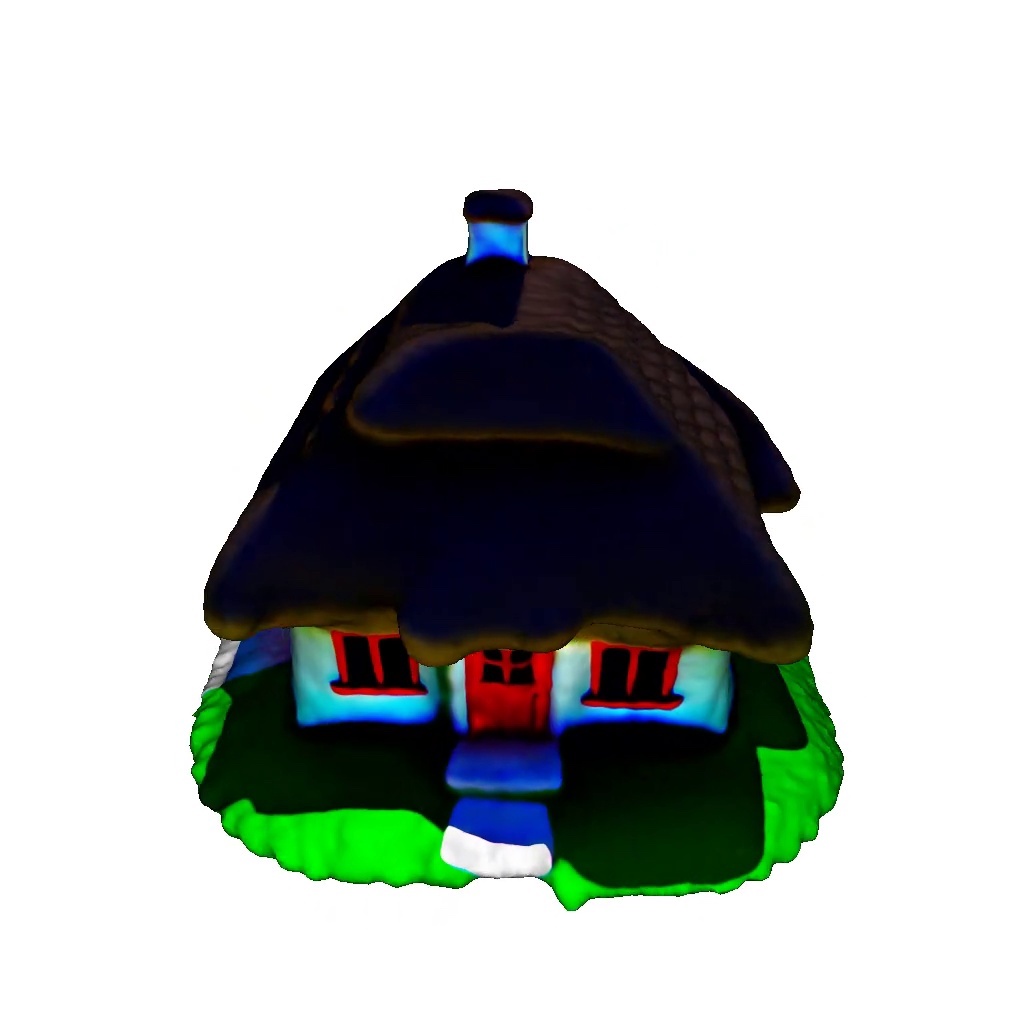} &
        \includegraphics[width=\linewidth]{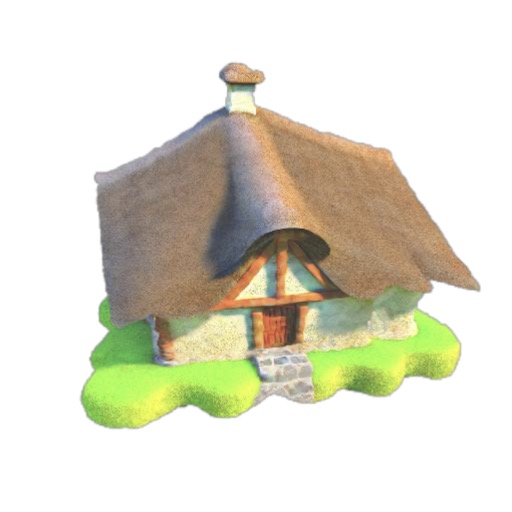} &
        \includegraphics[width=\linewidth]{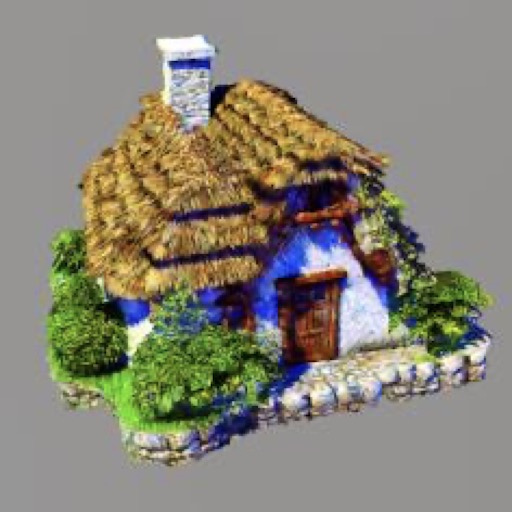} &
        \includegraphics[width=\linewidth]{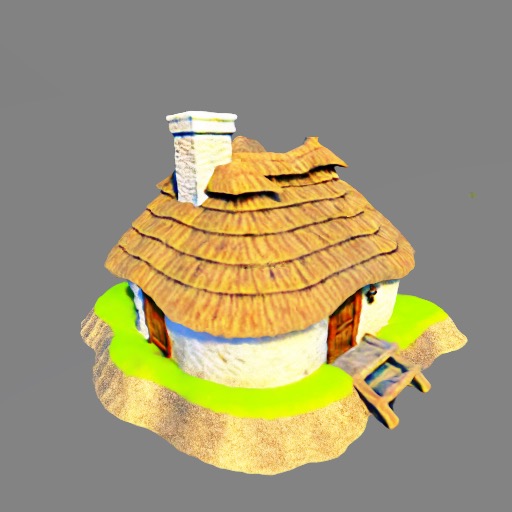} 
        \\
        DreamFusion & Latent-NeRF & Magic3D & Fantasia3D & ProlificDreamer & NFSD (ours)
    \end{tabular}
    }
    \vspace{-5pt}
    \caption{Comparison of NFSD with different methods. Our NFSD is of high resolution and exhibits detailed features. In the top row we used the prompt ``A plate piled high with chocolate chip cookies'', and in the bottom one we used ``A 3D model of an adorable cottage with a thatched roof''.}
    \vspace{-16pt}
    \label{fig:method_comparison}
\end{figure}

\textbf{Comparison with related methods.}
In \figref{fig:method_comparison} we compare our method with recent approaches including DreamFusion~\citep{poole2022dreamfusion}, Magic3D~\citep{lin2023magic3d}, Latent-NeRF~\citep{metzer2022latent}, Fantasia3D~\citep{Chen_2023_ICCV}, and ProlificDreamer~\citep{wang2023prolificdreamer}. 
Following previous works~\citep{metzer2022latent, wang2023prolificdreamer, Chen_2023_ICCV, lin2023magic3d}, for each of these methods we compare to results that were reported by the authors. As can be seen, our method achieves comparable or better results, while being significantly simpler than most of these methods. For example, observe the roof of the cottage, which is highly detailed in our method and in ProlificDreamer, but exhibits less detail in the other methods. Unlike ProlificDreamer which requires optimizing a diffusion model in parallel to the NeRF optimization, our method is simpler to implement and the optimization process is much faster. Please note that the original methods differ in their implementation details, including the diffusion model that guides the optimization. Therefore, in Appendix~\ref{sec:app-comparison}, we provide comparison using threestudio~\citep{threestudio2023} for all method results.

\vspace{-6pt}
\section{Conclusion and Future Work}
\label{sec:conclusion}
\vspace{-8pt}

In this paper, we have revisited the SDS process and introduced a novel interpretation, dissecting the score into three distinct components: the condition, the domain, and the denoising components.
Through this novel perspective, we proposed a simple distillation process, which we refer to as Noise-Free Score Distillation (NFSD). NFSD was developed with the explicit goal of preventing noise distillation during the optimization process. Notably, NFSD requires only minimal adjustments to the SDS framework, all while operating with a nominal Classifier-Free Guidance (CFG) scale.
Despite its simplicity, NFSD has shown promising potential in advancing the generation of 3D objects, demonstrating notable improvements when compared to both SDS and existing approaches.

While NFSD enables better score distillation compared with SDS, two main drawbacks of the SDS process still exist, namely the well known Janus problem (multi-face) and low diversity.
We believe that the latter is a direct result of the distillation process mechanism: 
\dl{the diffusion scores that guide the optimization are uncorrelated across successive iterations, even if noise is successfully eliminated. While annealing the diffusion time as the optimization progresses is helpful, designing a more principled noise scheduling might prove more effective, and lead to improved diversity.}

Additionally, we recognize the challenging nature of evaluating NeRFs generated from text, with a current absence of suitable metrics and benchmarks to assess NeRF quality and enable comprehensive comparisons across various methodologies. We believe that the development of such metrics is essential for advancing research in this domain and and would like to investigate this in the future.

\bibliography{iclr2024_conference}
\bibliographystyle{iclr2024_conference}

\clearpage
\appendix
\section{Appendix}

\vspace{-6pt}
\op{
\subsection{Visualization of score components} \label{sec:app-vis-latent}
\vspace{-6pt}

\begin{wrapfigure}{R}{0.15\textwidth}  
  \centering
  \includegraphics[width=0.15\textwidth]{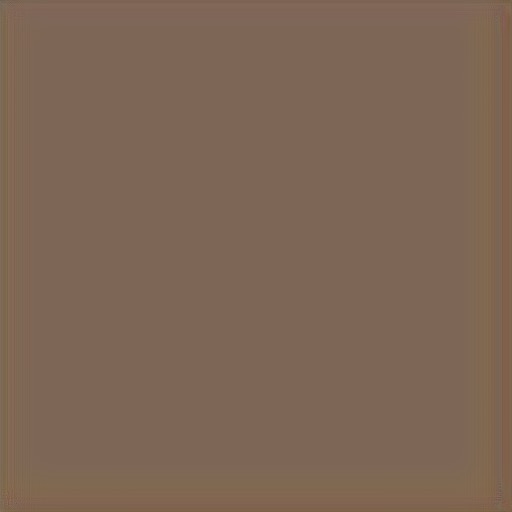}  
  \label{fig:zero-latent}
\end{wrapfigure}

In Figures \ref{fig:d_c}, \ref{fig:dr-dn} and \ref{fig:sds-dr} we visualize the score components, or manipulation of them. Stable Diffusion operates in the latent space of a pretrained autoencoder, and therefore we use the decoder to obtain these images. At a first glance it may be surprising that the decoder can decode such latents, as they are out of the distribution that it was trained on. However, as shown in \citep{linear_latent_approximation} and discussed by \citep{metzer2022latent}, the RGB value at a certain pixel, can be approximated by applying a fixed linear transformation on the corresponding pixel of the latent code. Hence, we conclude that the latent space of Stable Diffusion behaves similarly to the RGB space, and decoding to RGB latent codes that are out of distribution is still meaningful.
We also note that when decoding the zero latent code, we get the brown RGB image shown on the right. Therefore, this color in our visualizations indicates a value of $0$ in the latent code.

}

Additionally, in \figref{fig:dr-dn} we visualize $\dirr$ and $\dirn$ by generating pairs of $\x_{\textrm{ID}}$ (in-domain) and $\x_{\textrm{OOD}}$ (out-of-domain) images. The two ID images were generated using DDIM sampling with the prompts ``a pretty cat on a sand'' and ``a cow in a field''. Subsequently, generate the corresponding OOD images by applying a Gaussian filter to the cat image and a bilateral filter to the cow image.

\vspace{-6pt}
\subsection{Prompts used in the paper} \label{sec:app-prompts}
\vspace{-6pt}

Here we report the prompts used in several of the figures, in a left-to-right, top-to-bottom fashion.

In \figref{fig:3d-generation}: ``A huge Hedgehog", ``a soldier iron decor pen holder",  ``a trunk up statue of an Elephant with Thailand Decoration, side view", ``a photo of a vase with sunflowers", ``a brass statue of a dragonfly" and ``a rainbow colored wings spread parrot".

In \figref{fig:3d_comparison}: ``a eagle catching a snake", ``a golden statue of a fairy angel with white wings", ``a metal gargoyle statue with white wings, ``a silver metal running horse on a table glass", ``a huge Hystrix" and ``a phoenix in golden cage".

\vspace{-6pt}
\subsection{Additional results} \label{sec:app-results}
\vspace{-6pt}

\begin{figure}[h]
    \centering
    \setlength{\tabcolsep}{1pt}
    {\scriptsize
    \begin{tabular}{c c c c c c}
        \includegraphics[width=0.2\linewidth]{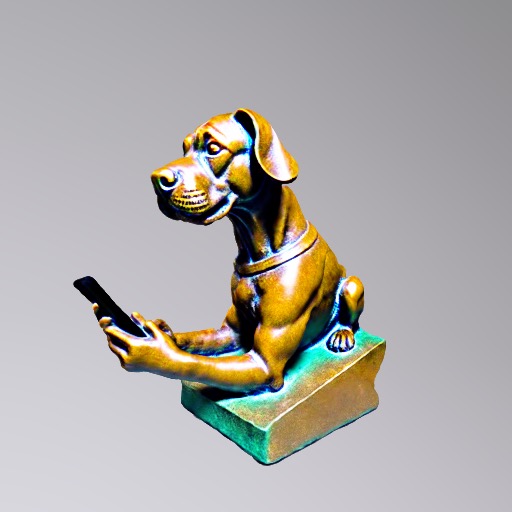} &
        \includegraphics[width=0.2\linewidth]{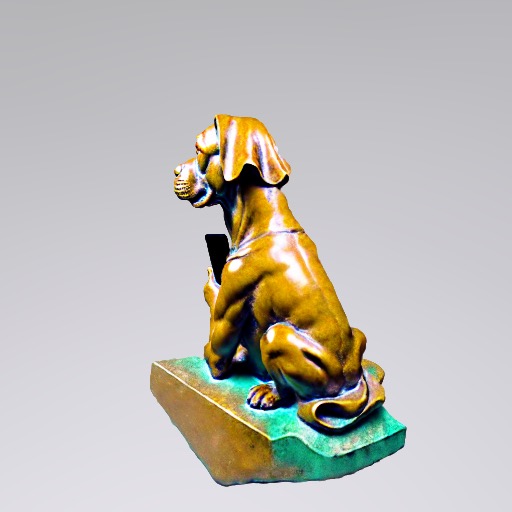} &
        \raisebox{0.7525\height}{%
        \begin{tabular}[b]{c}
            \adjustbox{valign=t}{\includegraphics[width=0.1\linewidth]{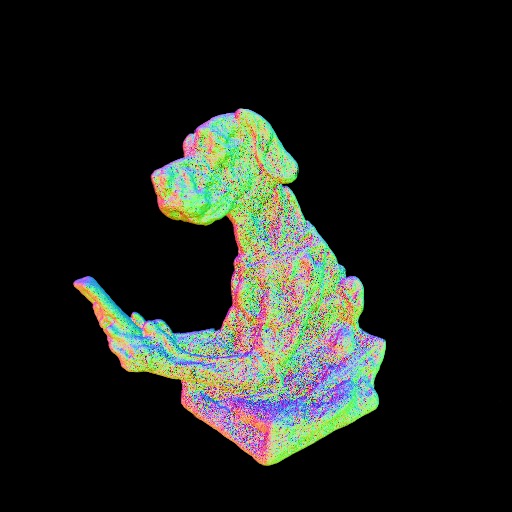}} \\
            \adjustbox{valign=t}{\includegraphics[width=0.1\linewidth]{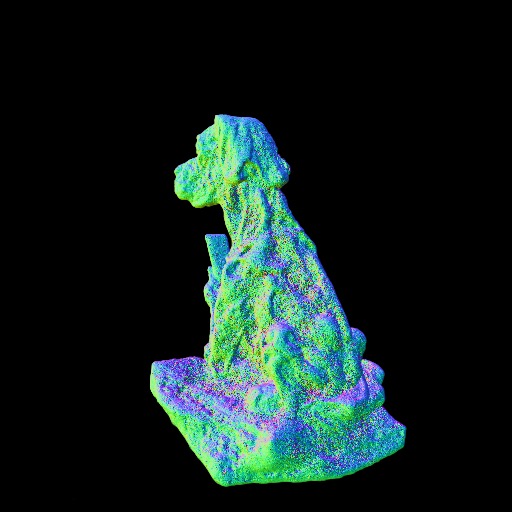}}
        \end{tabular}
        } &
        \includegraphics[width=0.2\linewidth]{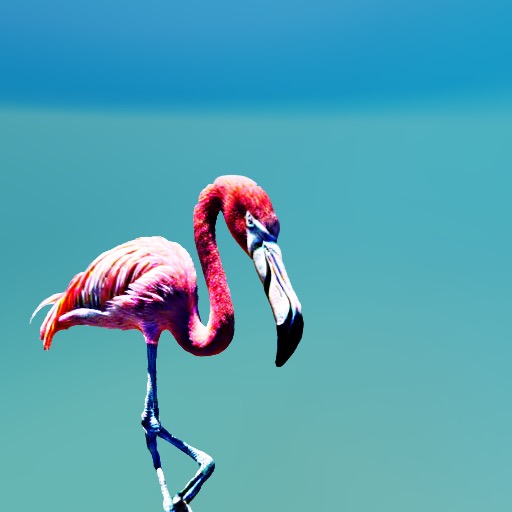} &
        \includegraphics[width=0.2\linewidth]{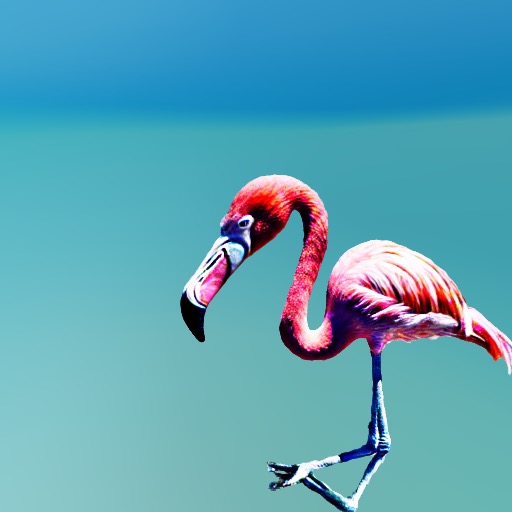} &
        \raisebox{0.7525\height}{%
        \begin{tabular}[b]{c}
            \adjustbox{valign=t}{\includegraphics[width=0.1\linewidth]{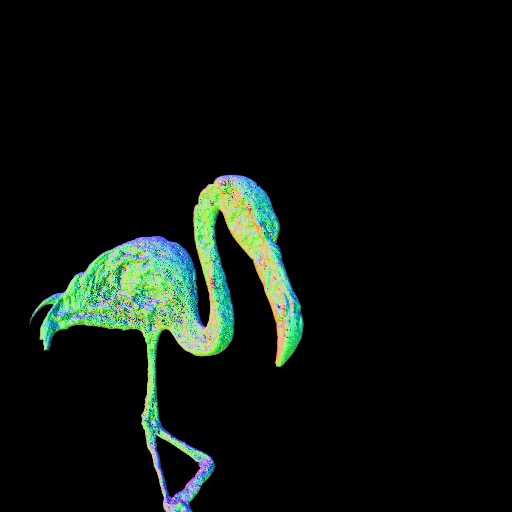}} \\
            \adjustbox{valign=t}{\includegraphics[width=0.1\linewidth]{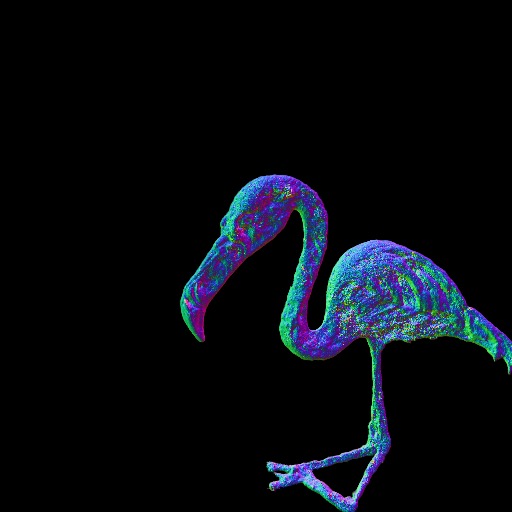}}
        \end{tabular}
        }  \\
        \multicolumn{3}{c}{``Michelangelo style statue of dog reading news on a cellphone''} &
        \multicolumn{3}{c}{``pink flamingo in paradise''} \\
        \includegraphics[width=0.2\linewidth]{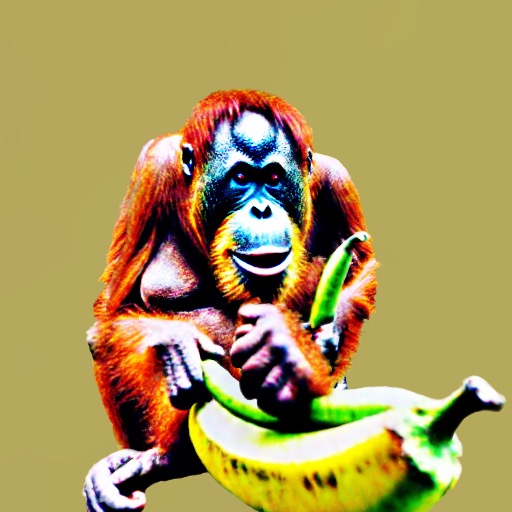} &
        \includegraphics[width=0.2\linewidth]{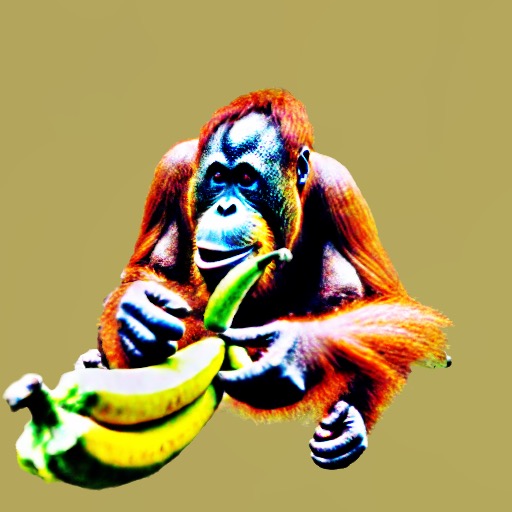} &
        \raisebox{0.7525\height}{%
        \begin{tabular}[b]{c}
            \adjustbox{valign=t}{\includegraphics[width=0.1\linewidth]{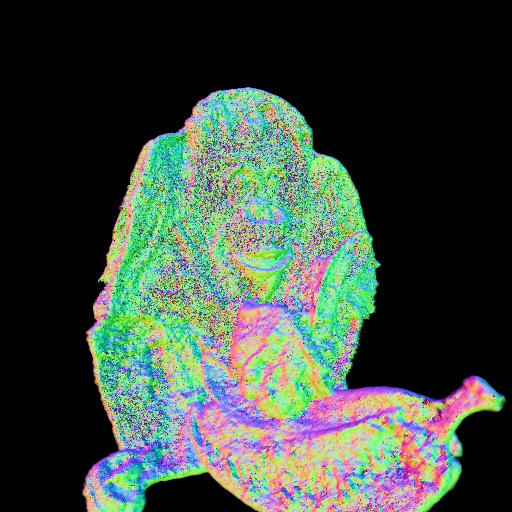}} \\
            \adjustbox{valign=t}{\includegraphics[width=0.1\linewidth]{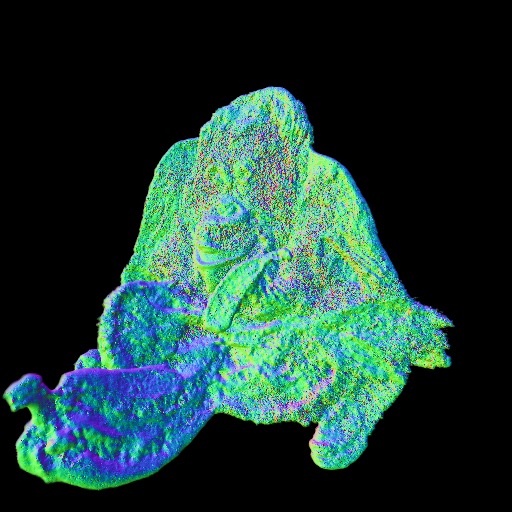}}
        \end{tabular}
        } &
        \includegraphics[width=0.2\linewidth]{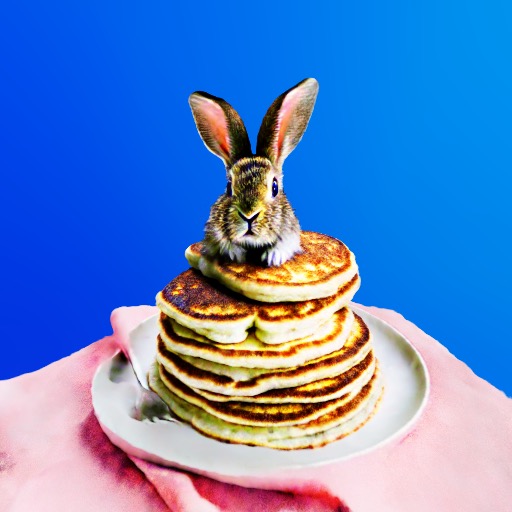} &
        \includegraphics[width=0.2\linewidth]{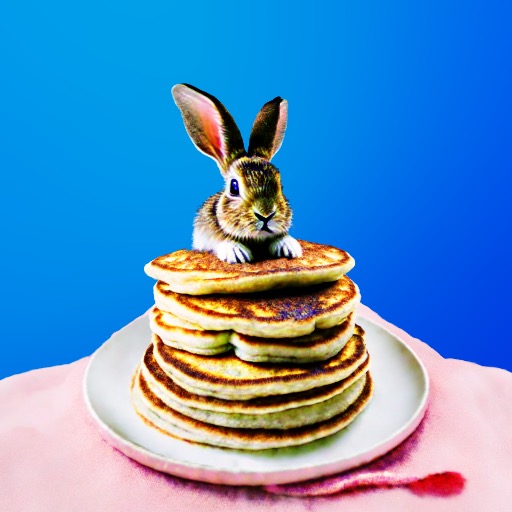} &
        \raisebox{0.7525\height}{%
        \begin{tabular}[b]{c}
            \adjustbox{valign=t}{\includegraphics[width=0.1\linewidth]{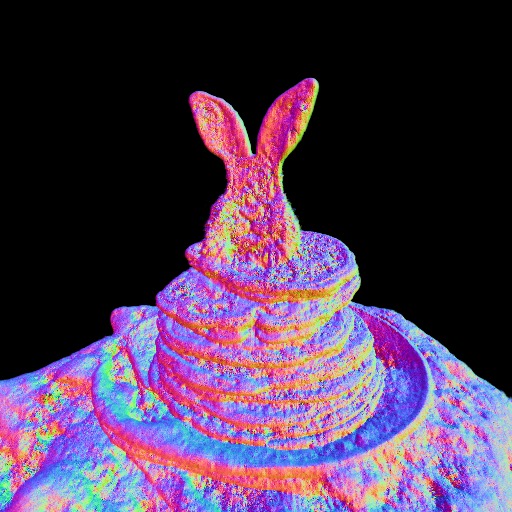}} \\
            \adjustbox{valign=t}{\includegraphics[width=0.1\linewidth]{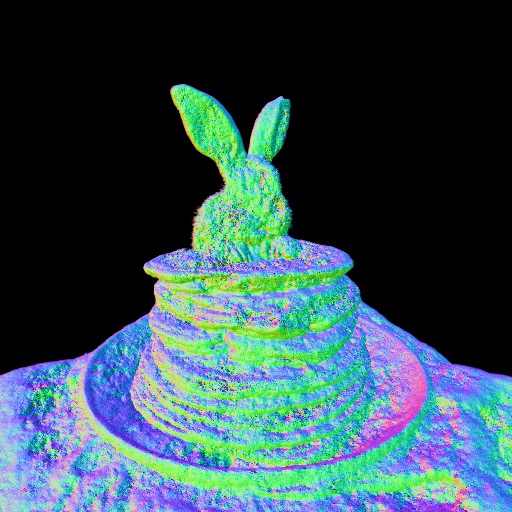}}
        \end{tabular}
        } 
         \\
        \multicolumn{3}{c}{``Orangutan eating a banana''} &
        \multicolumn{3}{c}{``[*] a baby bunny sitting on top of a stack of pancakes''}
        \\        %
        \includegraphics[width=0.2\linewidth]{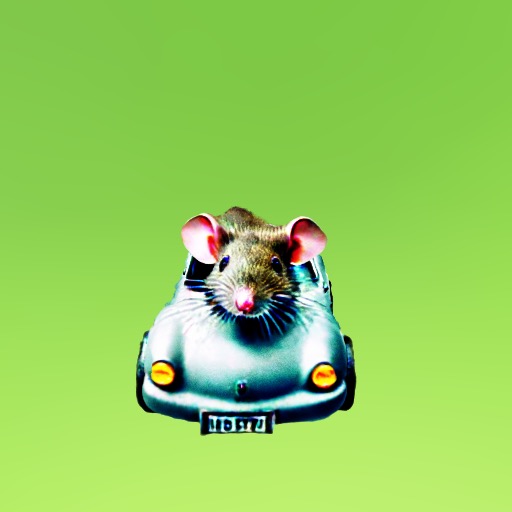} &
        \includegraphics[width=0.2\linewidth]{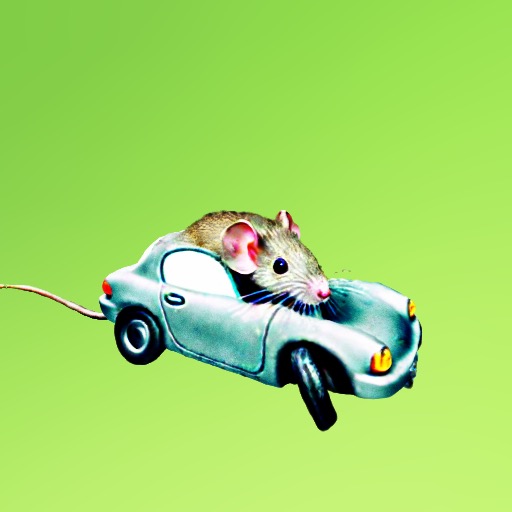} &
        \raisebox{0.7525\height}{%
        \begin{tabular}[b]{c}
            \adjustbox{valign=t}{\includegraphics[width=0.1\linewidth]{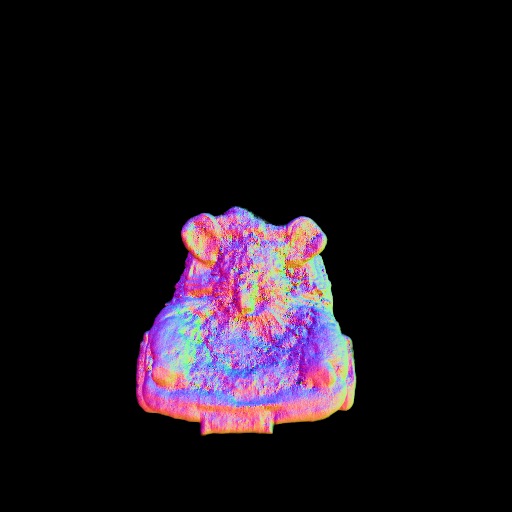}} \\
            \adjustbox{valign=t}{\includegraphics[width=0.1\linewidth]{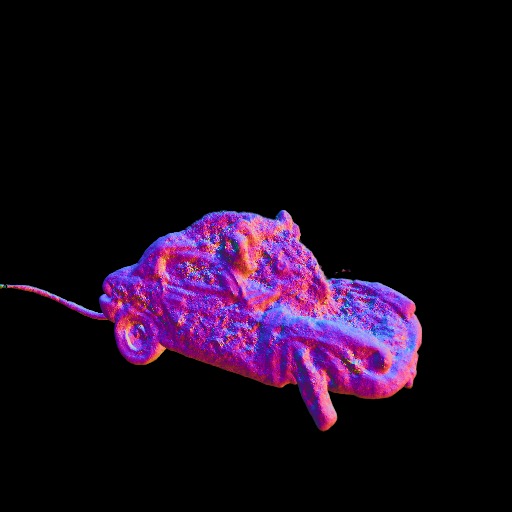}}
        \end{tabular}
        } &
        \includegraphics[width=0.2\linewidth]{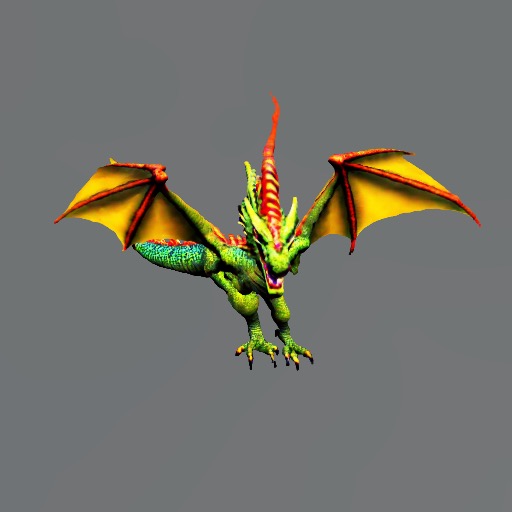} &
        \includegraphics[width=0.2\linewidth]{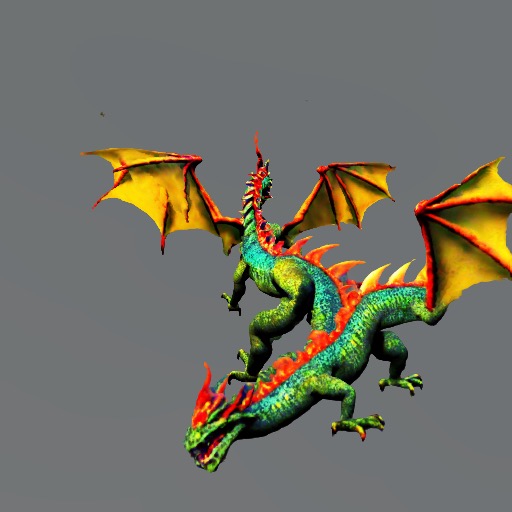} &
        \raisebox{0.7525\height}{%
        \begin{tabular}[b]{c}
            \adjustbox{valign=t}{\includegraphics[width=0.1\linewidth]{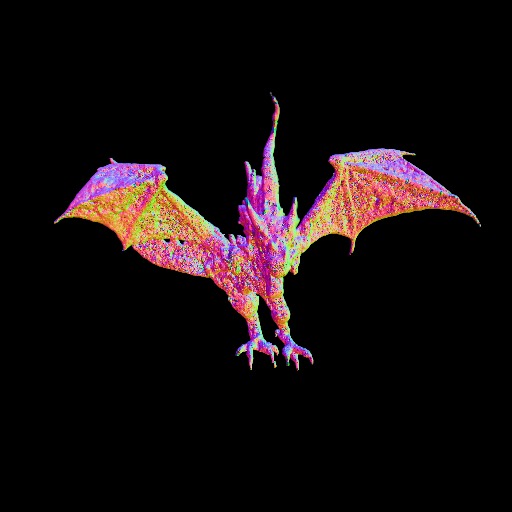}} \\
            \adjustbox{valign=t}{\includegraphics[width=0.1\linewidth]{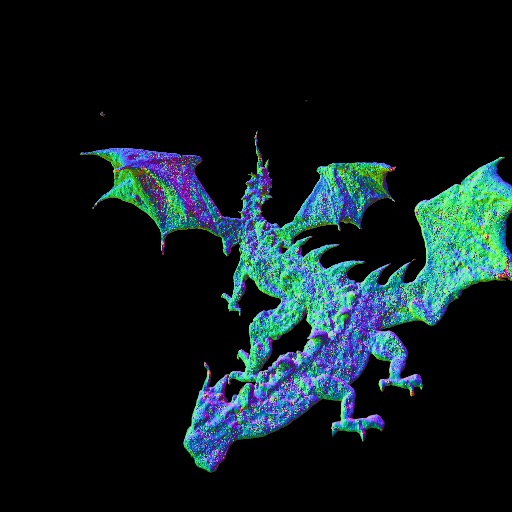}}
        \end{tabular}
        } 
         \\
        \multicolumn{3}{c}{``A rat driving a car''} &
        \multicolumn{3}{c}{``A two headed fire dragon with small wings''}
    \end{tabular}
    }
    \vspace{-12pt}
    \caption{NeRFs optimized with NFSD. [*] ``A zoomed out DSLR photo of'',  [...]-``A wide angle zoomed out DSLR photo of zoomed out view of''.}
    \vspace{-20pt}
    \label{fig:app-3d-generation}
\end{figure}

\begin{figure}
        \centering
    \setlength{\tabcolsep}{1pt}
    {\scriptsize
    \begin{tabular}{c c c c c c}
        \includegraphics[width=0.2\linewidth]{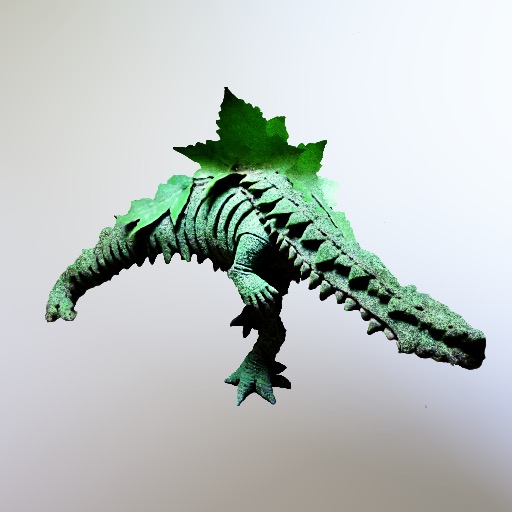} &
        \includegraphics[width=0.2\linewidth]{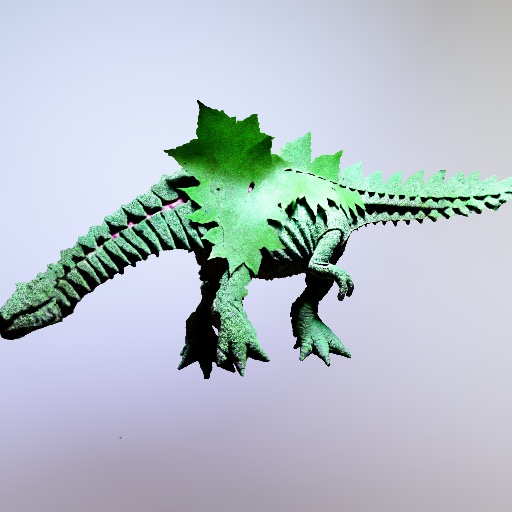} &
        \raisebox{0.7525\height}{%
        \begin{tabular}[b]{c}
            \adjustbox{valign=t}{\includegraphics[width=0.1\linewidth]{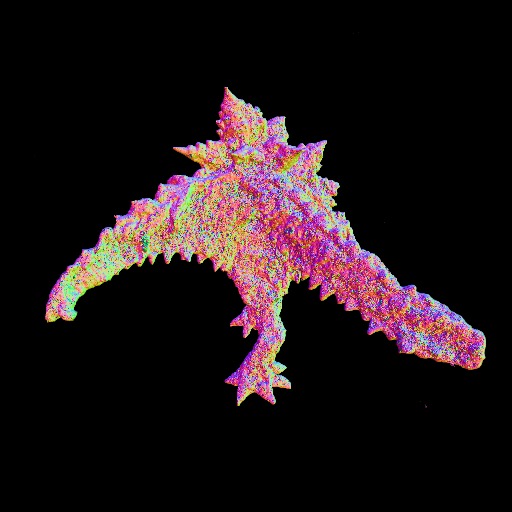}} \\
            \adjustbox{valign=t}{\includegraphics[width=0.1\linewidth]{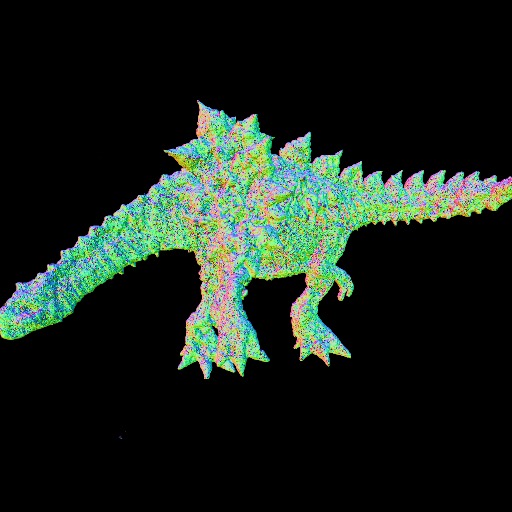}}
        \end{tabular}
        } &
        \includegraphics[width=0.2\linewidth]{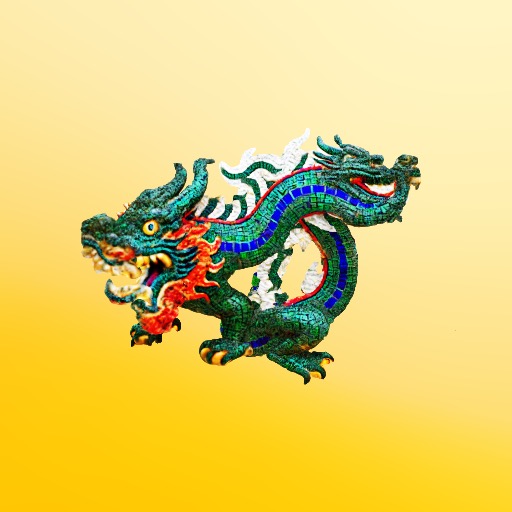} &
        \includegraphics[width=0.2\linewidth]{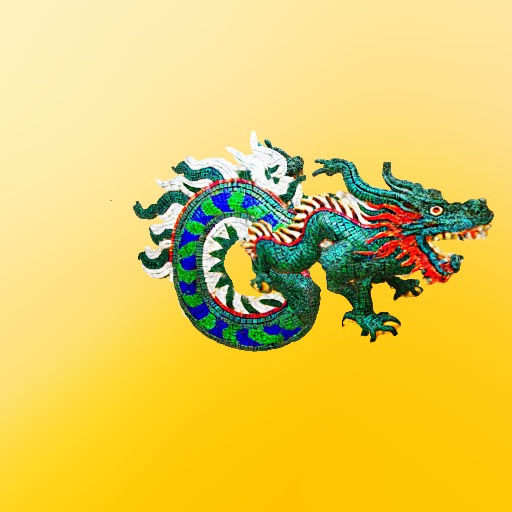} &
        \raisebox{0.7525\height}{%
        \begin{tabular}[b]{c}
            \adjustbox{valign=t}{\includegraphics[width=0.1\linewidth]{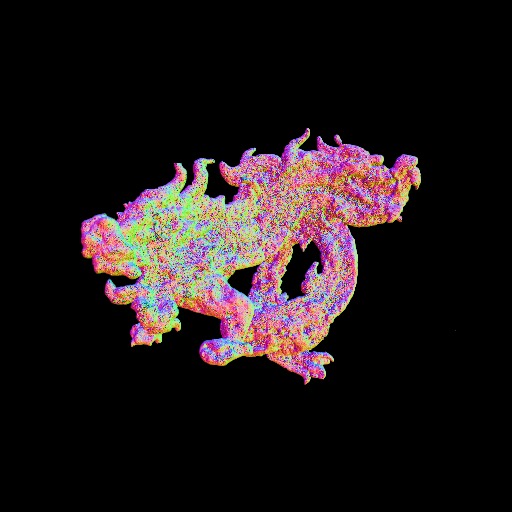}} \\
            \adjustbox{valign=t}{\includegraphics[width=0.1\linewidth]{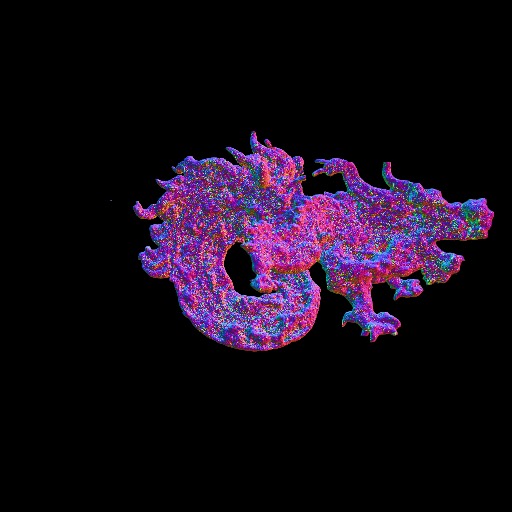}}
        \end{tabular}
        } 
        
        \\
        \multicolumn{3}{c}{``A highly detailed statue of a dinosaur made of leaves''} &
        \multicolumn{3}{c}{``Decorative mosaic of the mythological Chinese dragon''}
        \\

        \includegraphics[width=0.2\linewidth]{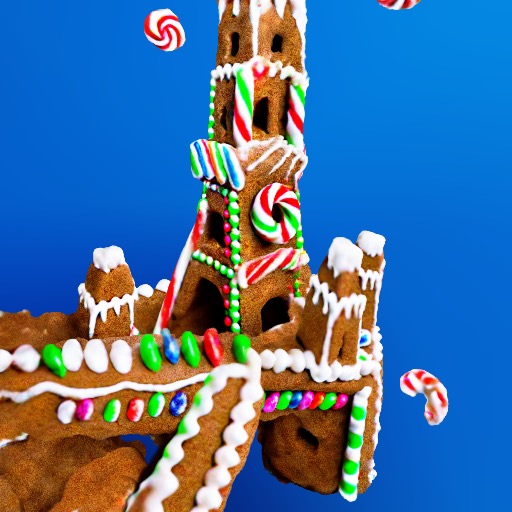} &
        \includegraphics[width=0.2\linewidth]{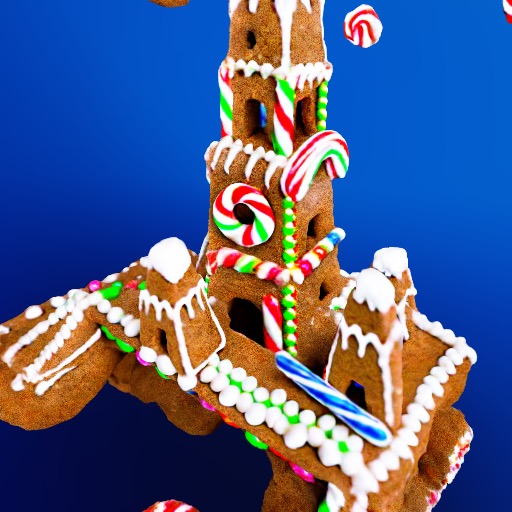} &
        \raisebox{0.7525\height}{%
        \begin{tabular}[b]{c}
            \adjustbox{valign=t}{\includegraphics[width=0.1\linewidth]{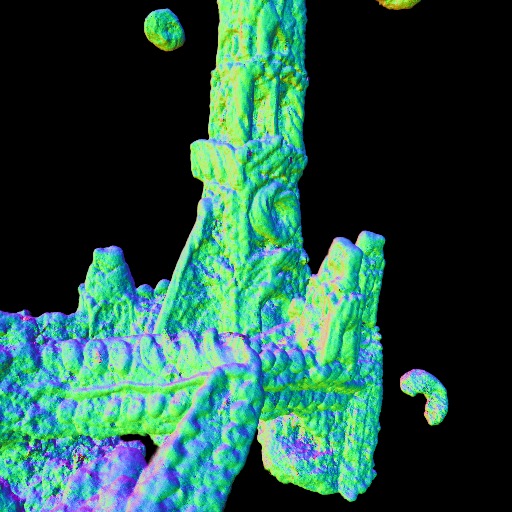}} \\
            \adjustbox{valign=t}{\includegraphics[width=0.1\linewidth]{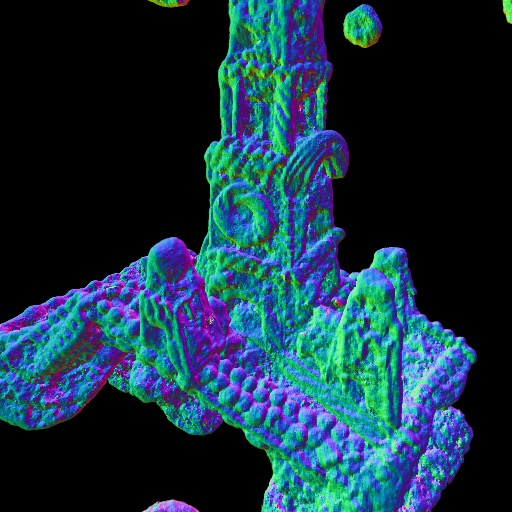}}
        \end{tabular}
        } &
        \includegraphics[width=0.2\linewidth]{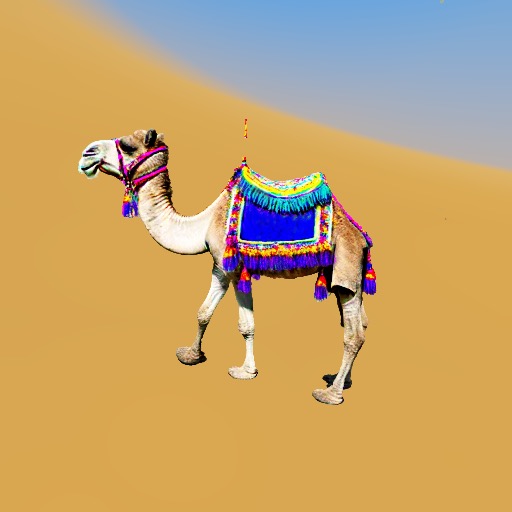} &
        \includegraphics[width=0.2\linewidth]{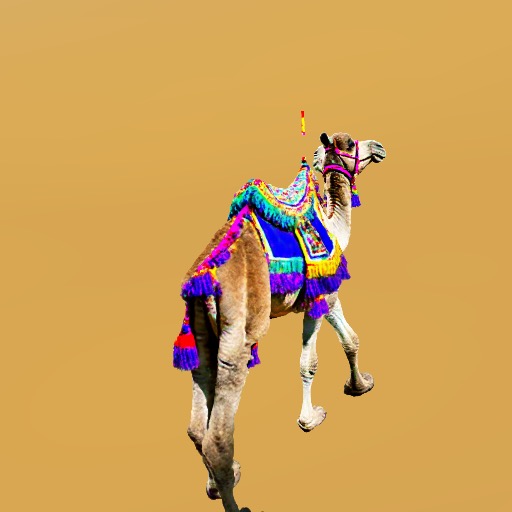} &
        \raisebox{0.7525\height}{%
        \begin{tabular}[b]{c}
            \adjustbox{valign=t}{\includegraphics[width=0.1\linewidth]{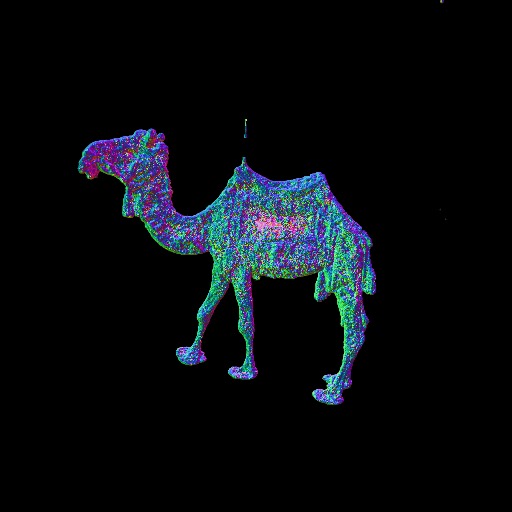}} \\
            \adjustbox{valign=t}{\includegraphics[width=0.1\linewidth]{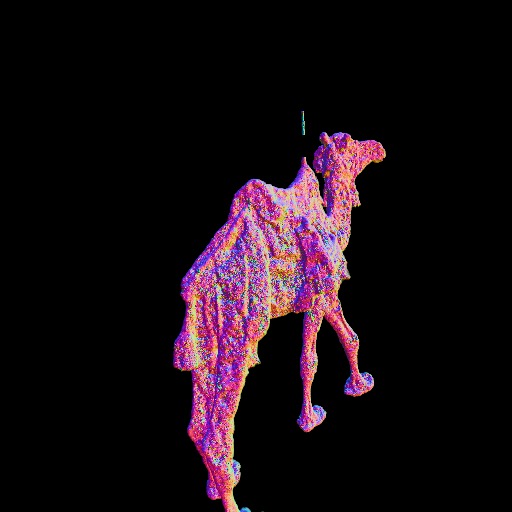}}
        \end{tabular}
        } 
        \\
        \multicolumn{3}{c}{``[...] Tower Bridge made out of gingerbread and candy''} & 
        \multicolumn{3}{c}{``A camel with a colorful saddle''} \\
        \includegraphics[width=0.2\linewidth]{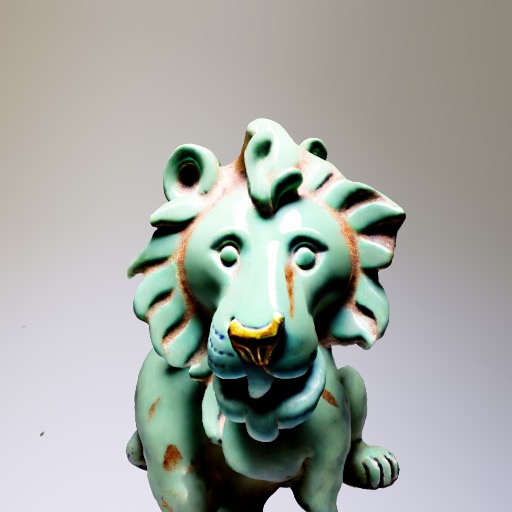} &
        \includegraphics[width=0.2\linewidth]{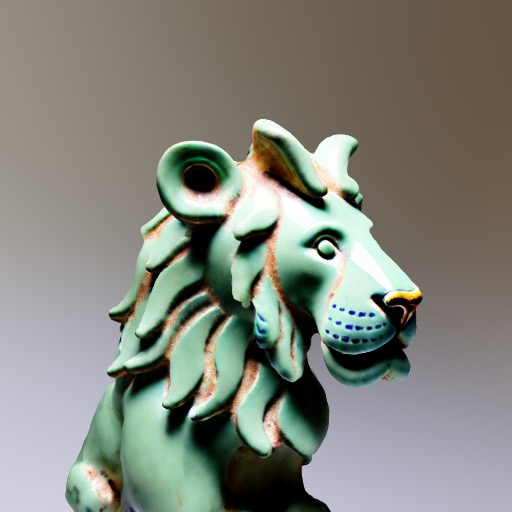} &
        \raisebox{0.7525\height}{%
        \begin{tabular}[b]{c}
            \adjustbox{valign=t}{\includegraphics[width=0.1\linewidth]{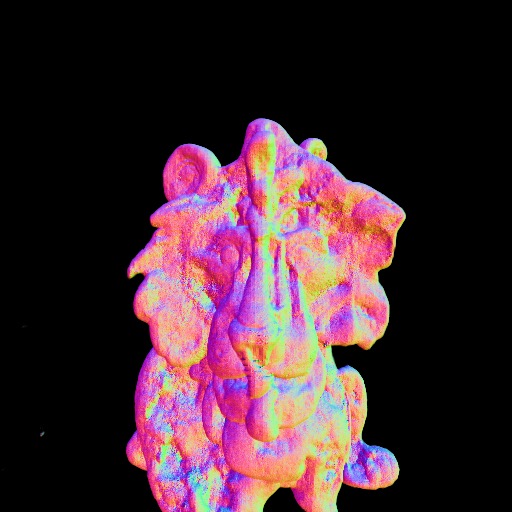}} \\
            \adjustbox{valign=t}{\includegraphics[width=0.1\linewidth]{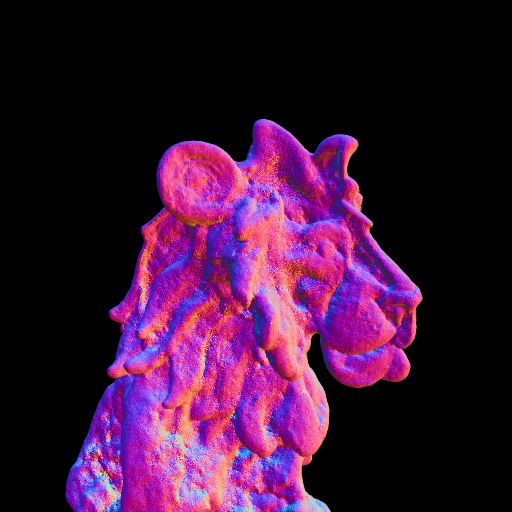}}
        \end{tabular}
        } &
        \includegraphics[width=0.2\linewidth]{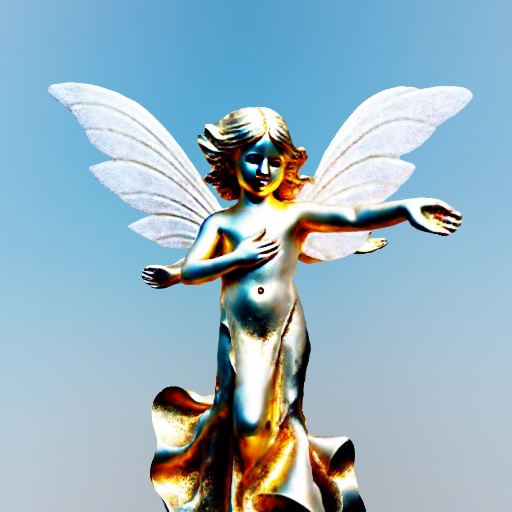} &
        \includegraphics[width=0.2\linewidth]{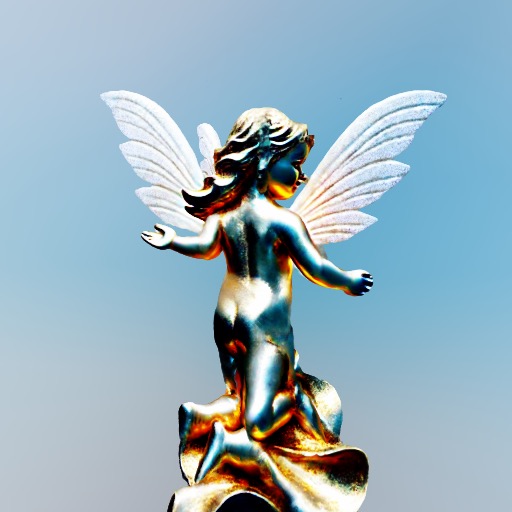} &
        \raisebox{0.7525\height}{%
        \begin{tabular}[b]{c}
            \adjustbox{valign=t}{\includegraphics[width=0.1\linewidth]{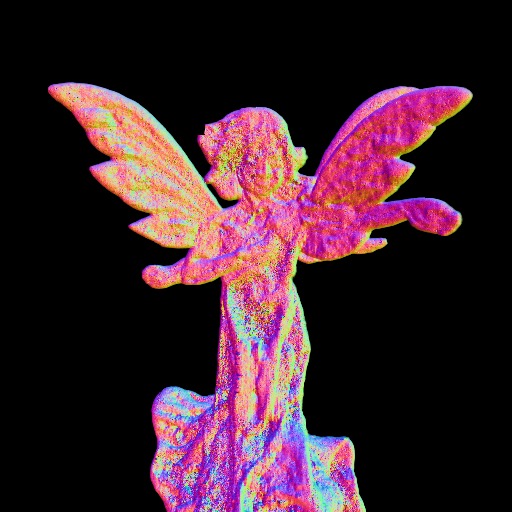}} \\
            \adjustbox{valign=t}{\includegraphics[width=0.1\linewidth]{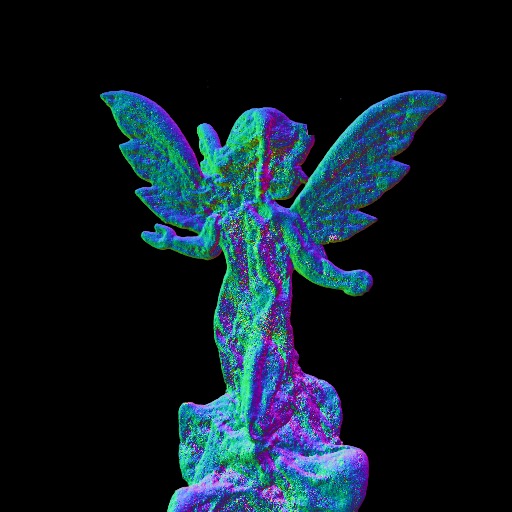}}
        \end{tabular}
        } 
        \\
        \multicolumn{3}{c}{``a ceramic lion''} &
        \multicolumn{3}{c}{``A golden statue of a fairy angel with white wings''}
        \\
        \includegraphics[width=0.2\linewidth]{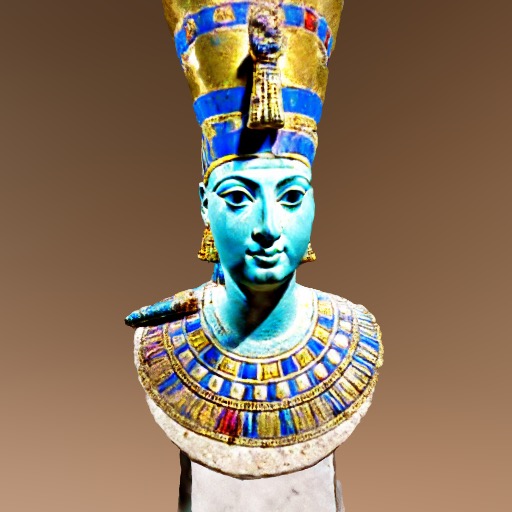} &
        \includegraphics[width=0.2\linewidth]{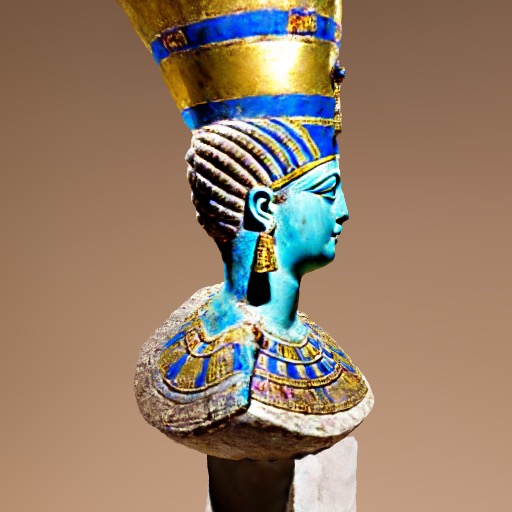} &
        \raisebox{0.7525\height}{%
        \begin{tabular}[b]{c}
            \adjustbox{valign=t}{\includegraphics[width=0.1\linewidth]{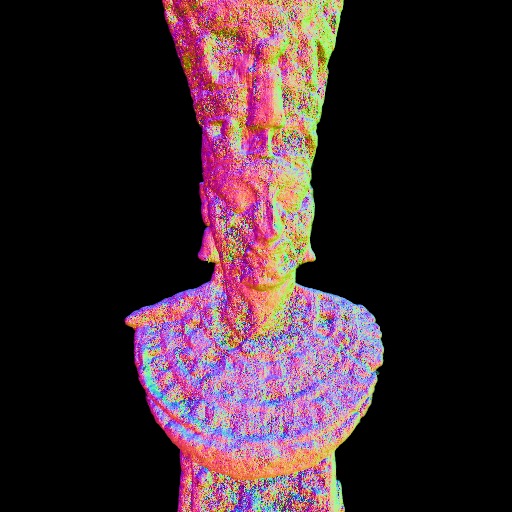}} \\
            \adjustbox{valign=t}{\includegraphics[width=0.1\linewidth]{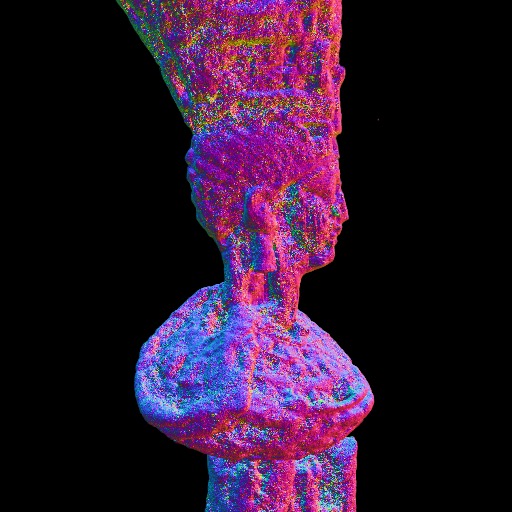}}
        \end{tabular}
        } &
        \includegraphics[width=0.2\linewidth]{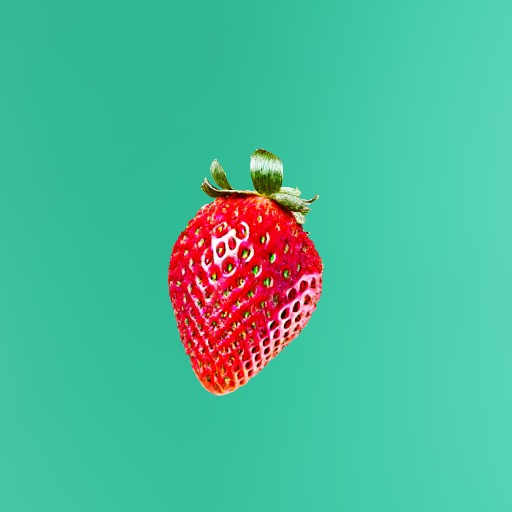} &
        \includegraphics[width=0.2\linewidth]{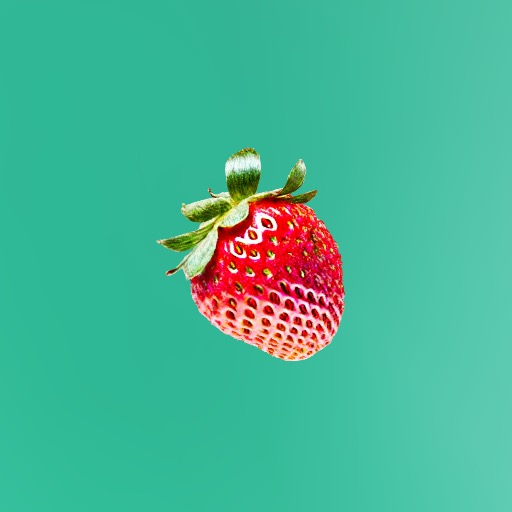} &
        \raisebox{0.7525\height}{%
        \begin{tabular}[b]{c}
            \adjustbox{valign=t}{\includegraphics[width=0.1\linewidth]{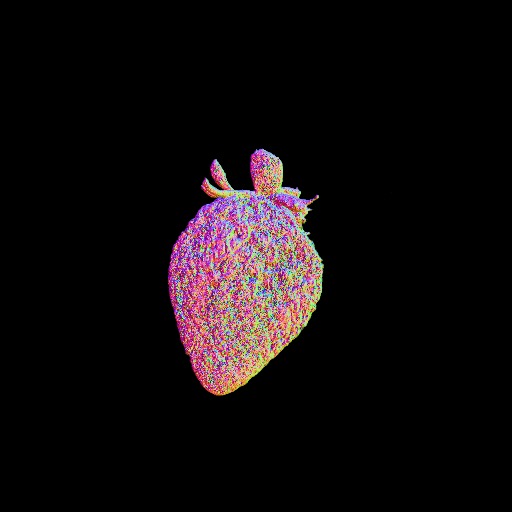}} \\
            \adjustbox{valign=t}{\includegraphics[width=0.1\linewidth]{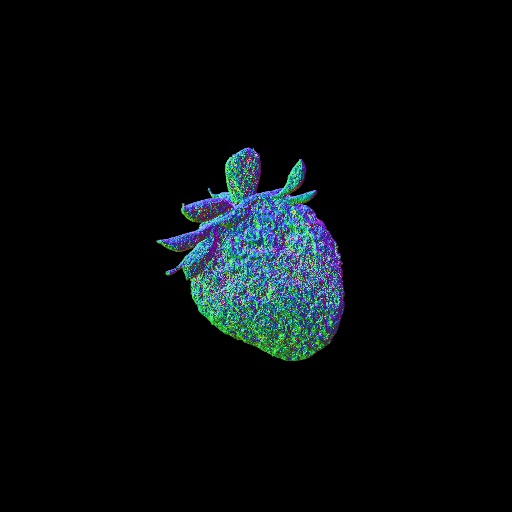}}
        \end{tabular}
        } 
        \\
        \multicolumn{3}{c}{``A sculpture of cleopatra with ceremonial decoration''} &
        \multicolumn{3}{c}{``A ripe strawberry''}
        \\
        \includegraphics[width=0.2\linewidth]{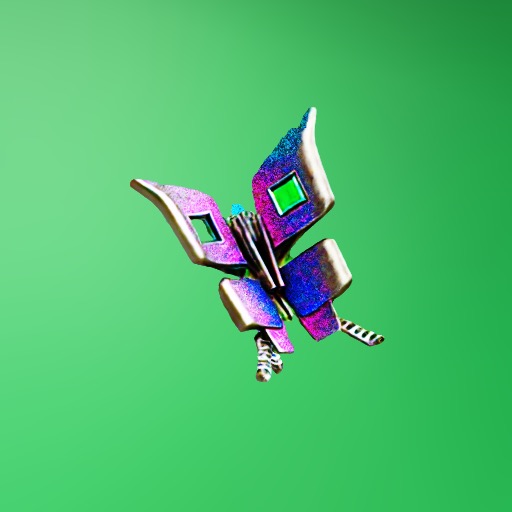} &
        \includegraphics[width=0.2\linewidth]{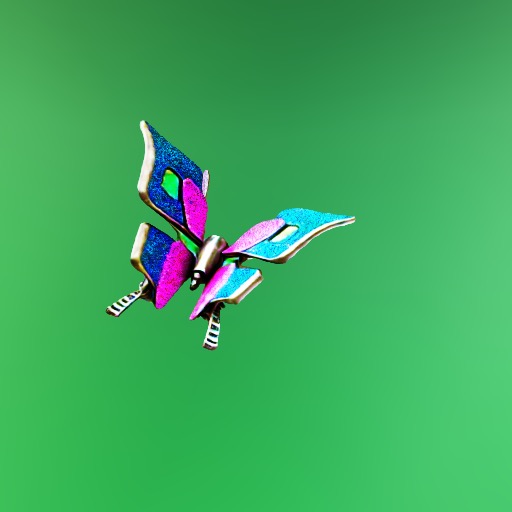} &
        \raisebox{0.7525\height}{%
        \begin{tabular}[b]{c}
            \adjustbox{valign=t}{\includegraphics[width=0.1\linewidth]{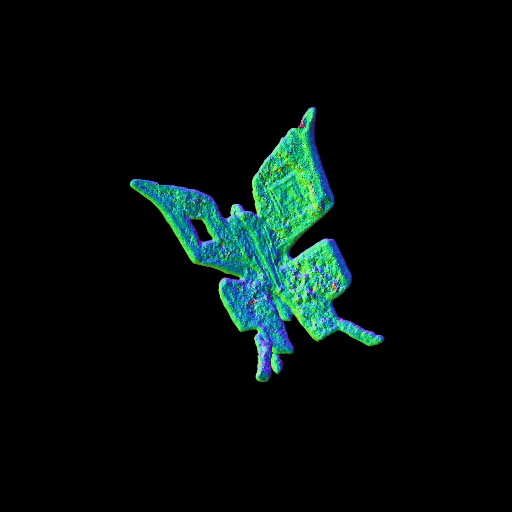}} \\
            \adjustbox{valign=t}{\includegraphics[width=0.1\linewidth]{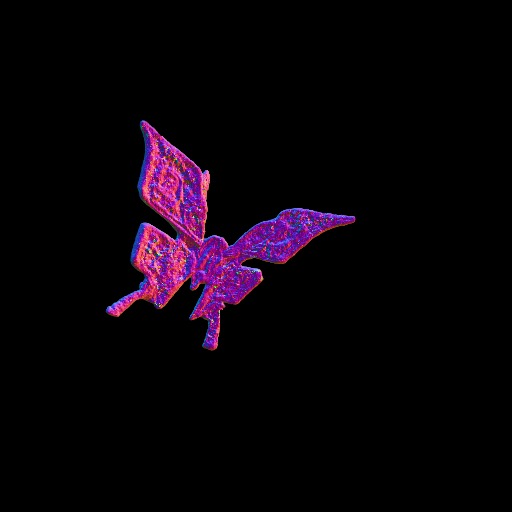}}
        \end{tabular}
        } &
        \includegraphics[width=0.2\linewidth]{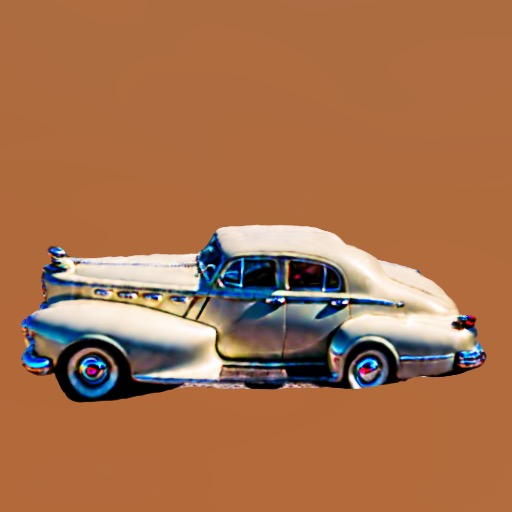} &
        \includegraphics[width=0.2\linewidth]{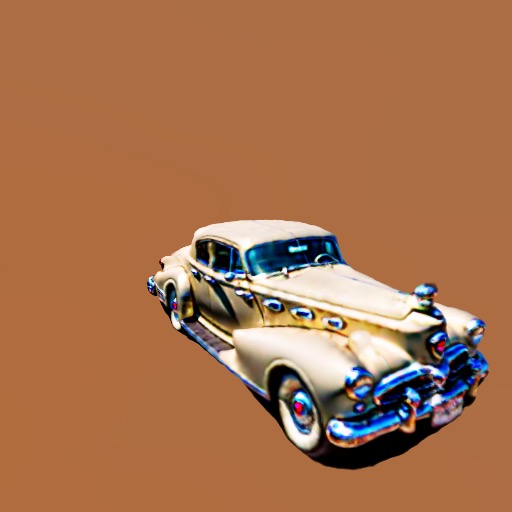} &
        \raisebox{0.7525\height}{%
        \begin{tabular}[b]{c}
            \adjustbox{valign=t}{\includegraphics[width=0.1\linewidth]{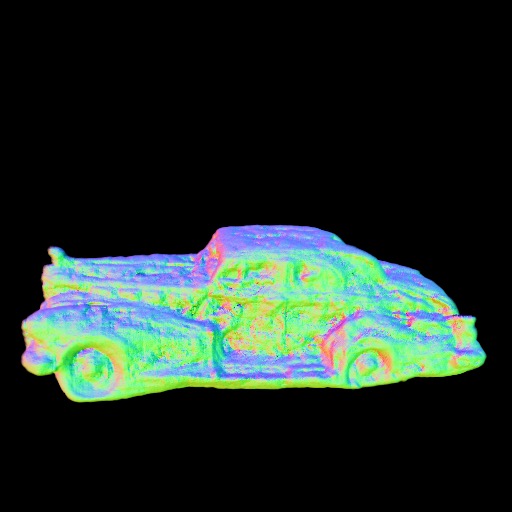}} \\
            \adjustbox{valign=t}{\includegraphics[width=0.1\linewidth]{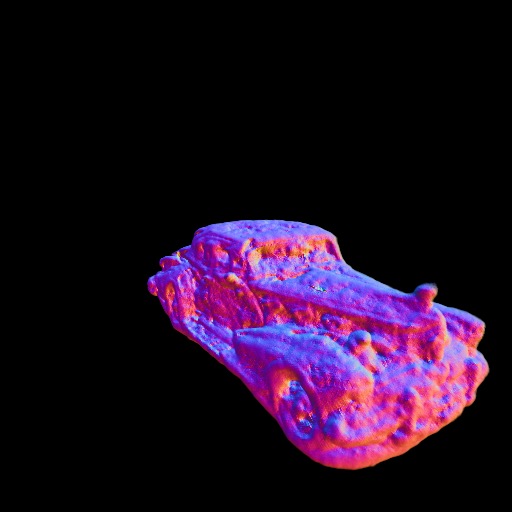}}
        \end{tabular}
        } 
        \\
        \multicolumn{3}{c}{``A robotic butterfly on a metal flower''} &
        \multicolumn{3}{c}{``A DSLR photo of a classic Packard car''}
        \\
        \includegraphics[width=0.2\linewidth]{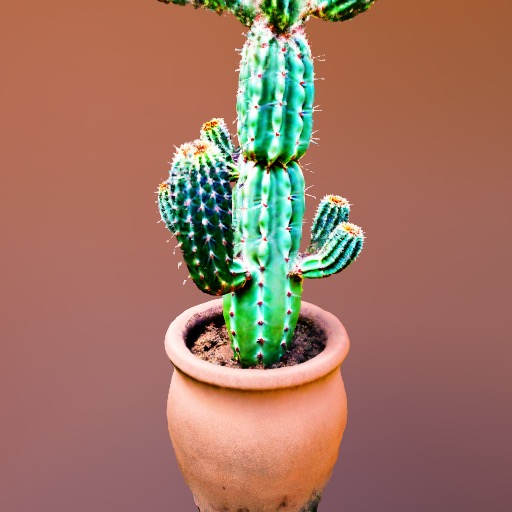} &
        \includegraphics[width=0.2\linewidth]{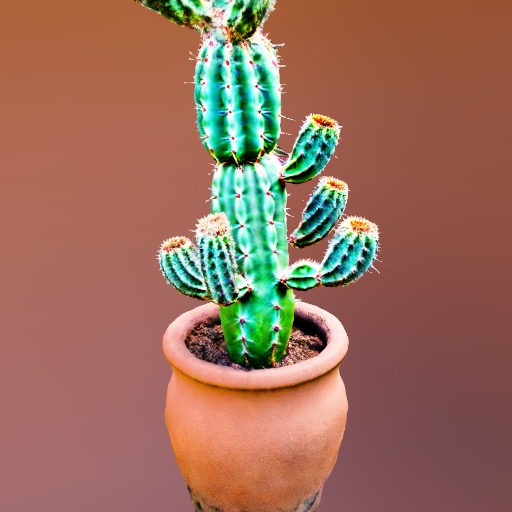} &
        \raisebox{0.7525\height}{%
        \begin{tabular}[b]{c}
            \adjustbox{valign=t}{\includegraphics[width=0.1\linewidth]{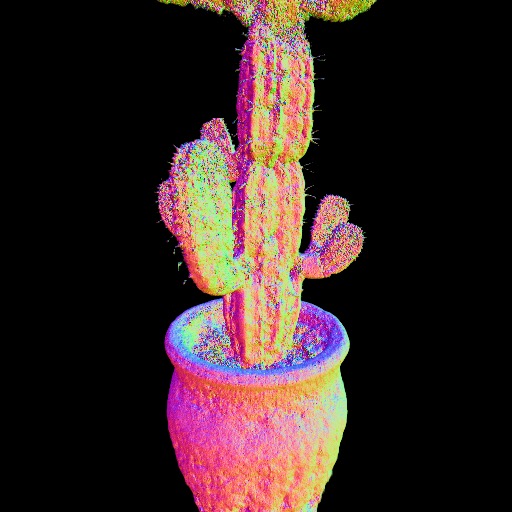}} \\
            \adjustbox{valign=t}{\includegraphics[width=0.1\linewidth]{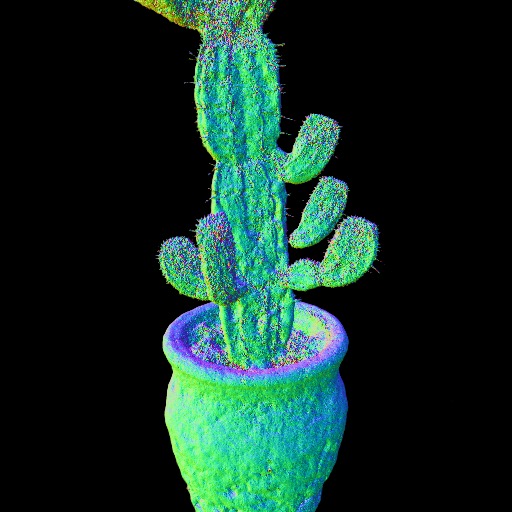}}
        \end{tabular}
        } &
        \includegraphics[width=0.2\linewidth]{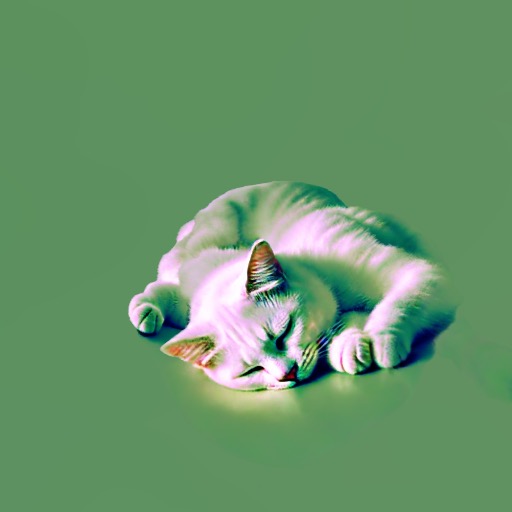} &
        \includegraphics[width=0.2\linewidth]{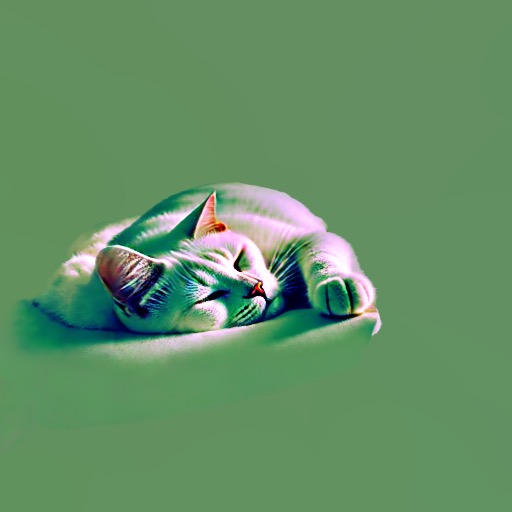} &
        \raisebox{0.7525\height}{%
        \begin{tabular}[b]{c}
            \adjustbox{valign=t}{\includegraphics[width=0.1\linewidth]{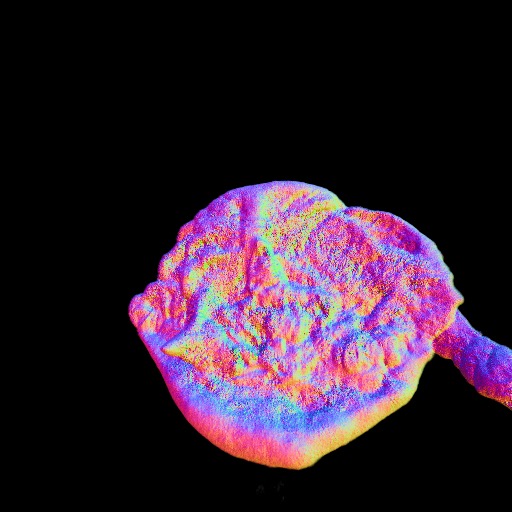}} \\
            \adjustbox{valign=t}{\includegraphics[width=0.1\linewidth]{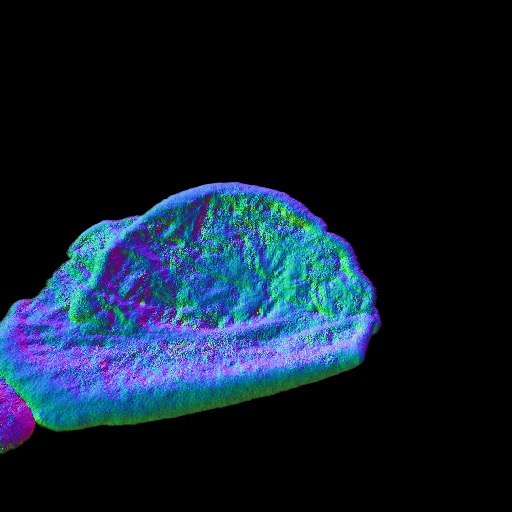}}
        \end{tabular}
        } 
        \\
        \multicolumn{3}{c}{``A small saguaro cactus planted in a clay pot''} &
        \multicolumn{3}{c}{``A white cat sleeping''}
        \\
        \includegraphics[width=0.2\linewidth]{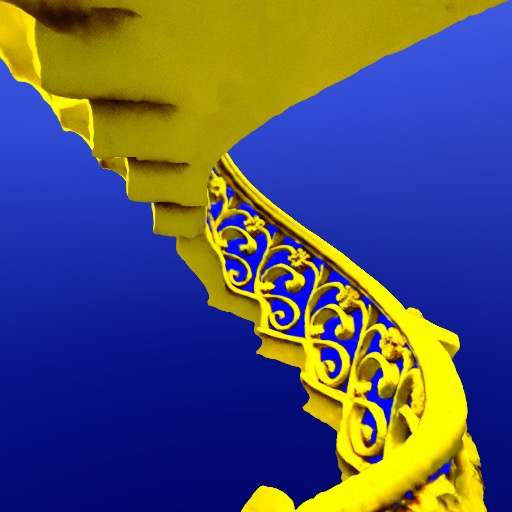} &
        \includegraphics[width=0.2\linewidth]{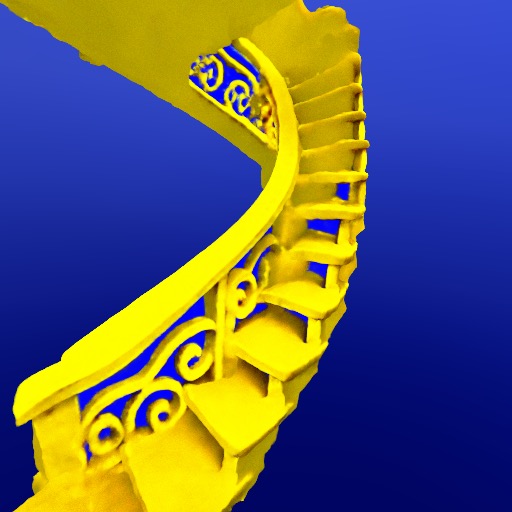} &
        \raisebox{0.7525\height}{%
        \begin{tabular}[b]{c}
            \adjustbox{valign=t}{\includegraphics[width=0.1\linewidth]{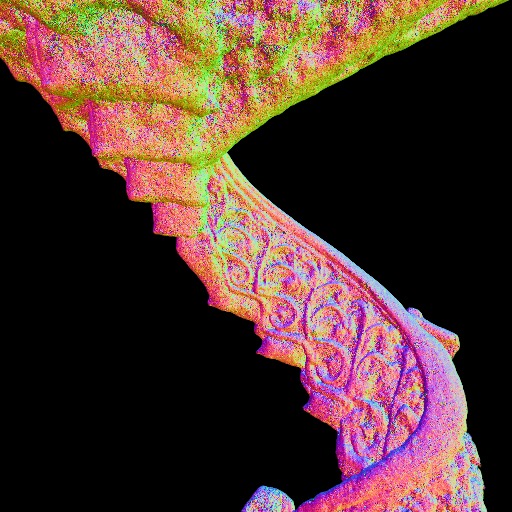}} \\
            \adjustbox{valign=t}{\includegraphics[width=0.1\linewidth]{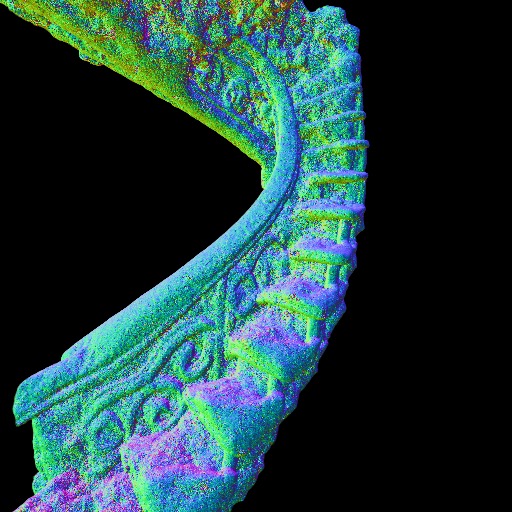}}
        \end{tabular}
        } &
        \includegraphics[width=0.2\linewidth]{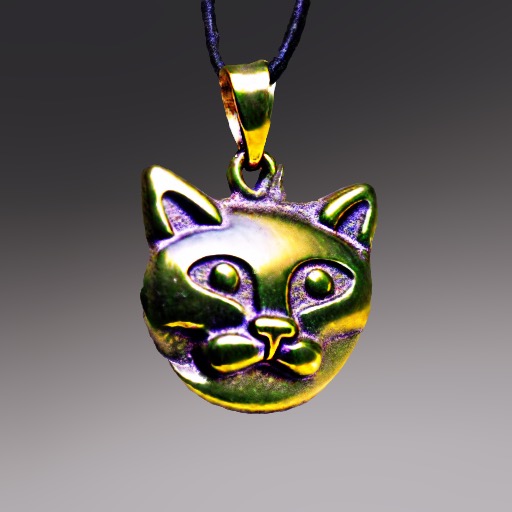} &
        \includegraphics[width=0.2\linewidth]{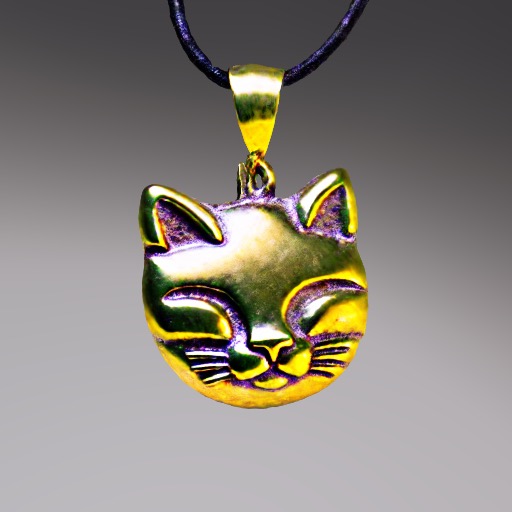} &
        \raisebox{0.7525\height}{%
        \begin{tabular}[b]{c}
            \adjustbox{valign=t}{\includegraphics[width=0.1\linewidth]{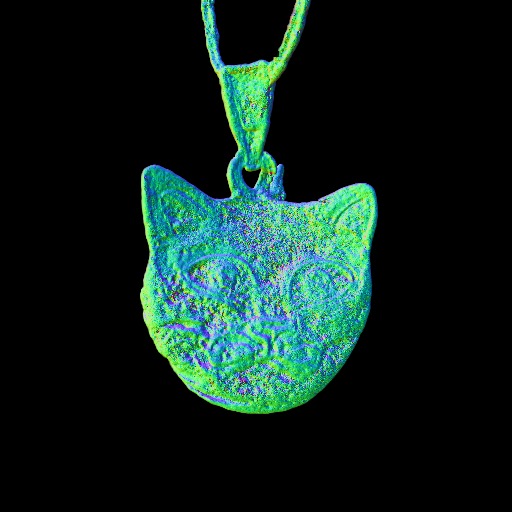}} \\
            \adjustbox{valign=t}{\includegraphics[width=0.1\linewidth]{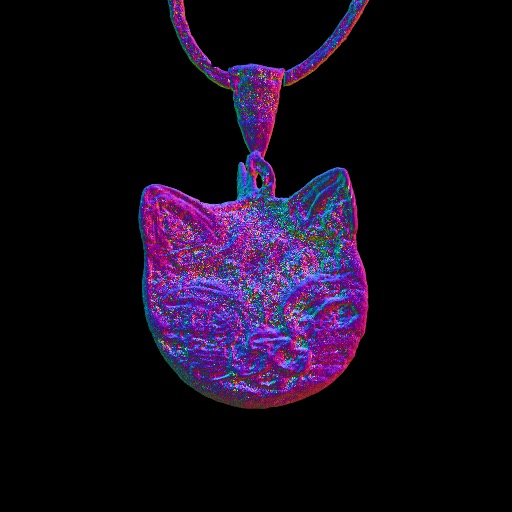}}
        \end{tabular}
        } 
        \\
        \multicolumn{3}{c}{``Railing tangga mewah''} &
        \multicolumn{3}{c}{``A tumbaga pendant depicting a cat''}
        \\

    \end{tabular}
    }
    \vspace{-8pt}
    \caption{NeRFs optimized with NFSD. [*] ``A zoomed out DSLR photo of'',  [...]-``A wide angle zoomed out DSLR photo of zoomed out view of''.}
    \label{fig:app-3d-generation-2}
\end{figure}

In Figures \ref{fig:app-3d-generation} and \ref{fig:app-3d-generation-2} we present additional results for text-to-NeRF generation using our NFSD.


\clearpage
\subsection{More Comparisons with Related Methods} \label{sec:app-comparison}
\begin{figure}[h]
    \centering
    \setlength{\tabcolsep}{1pt}
    \begin{tabular}{C{0.16\linewidth} C{0.16\linewidth} C{0.16\linewidth} C{0.16\linewidth} C{0.16\linewidth} C{0.16\linewidth} C{0.16\linewidth}}
        \includegraphics[width=\linewidth]{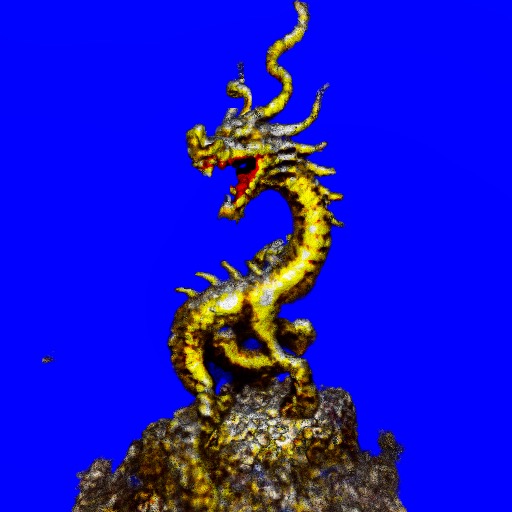} &
        \includegraphics[width=\linewidth]{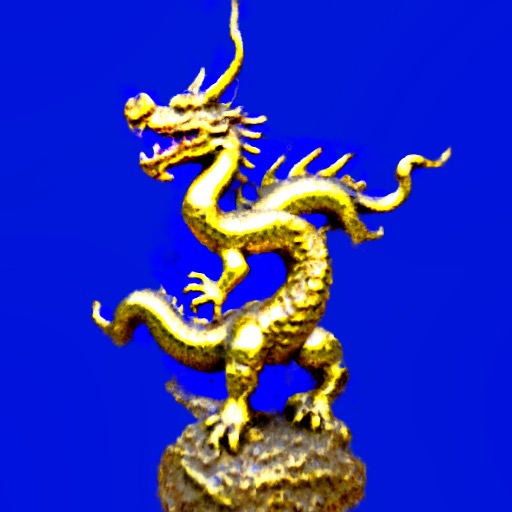} &
        \includegraphics[width=\linewidth]{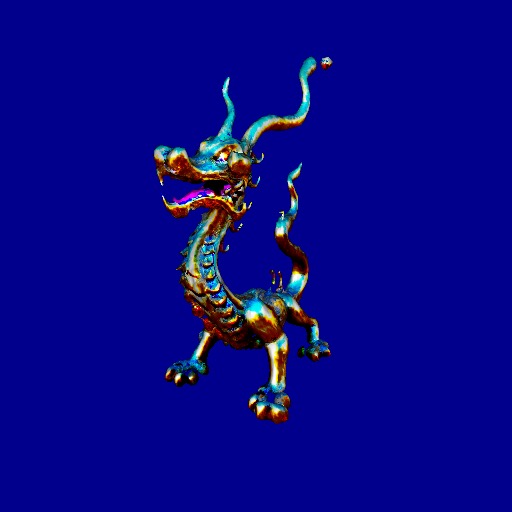} &
        \includegraphics[width=\linewidth]{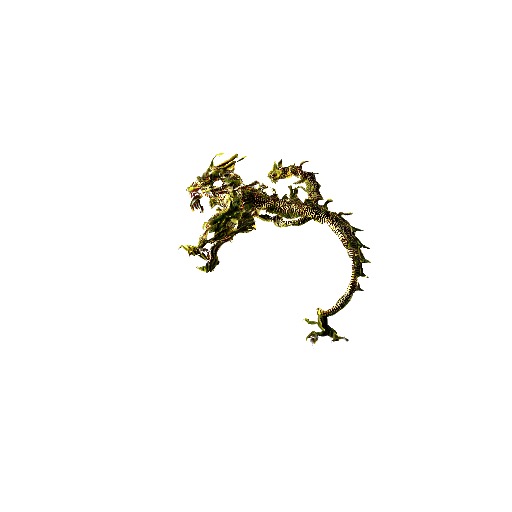} &
        \includegraphics[width=\linewidth]{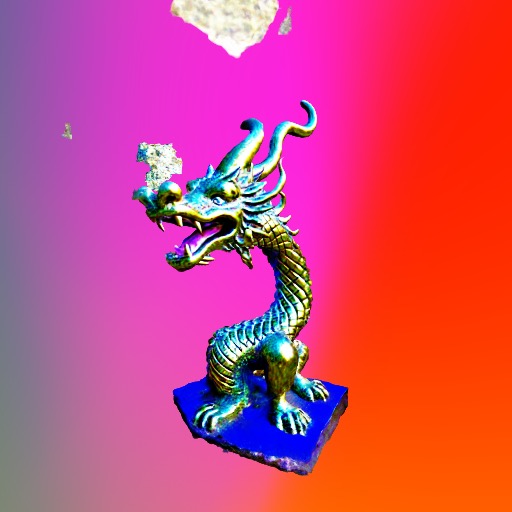} &
        \includegraphics[width=\linewidth]{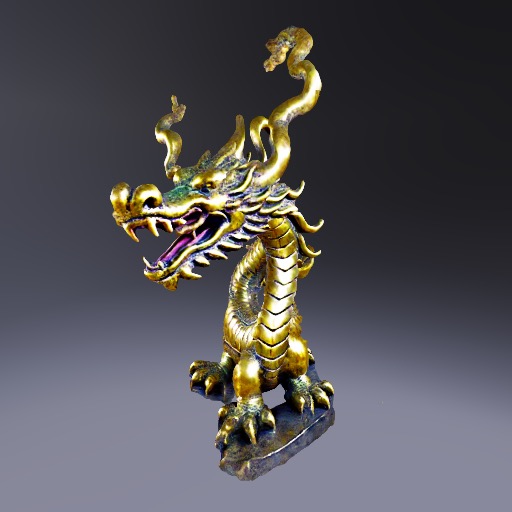} &
        \\
        \includegraphics[width=\linewidth]{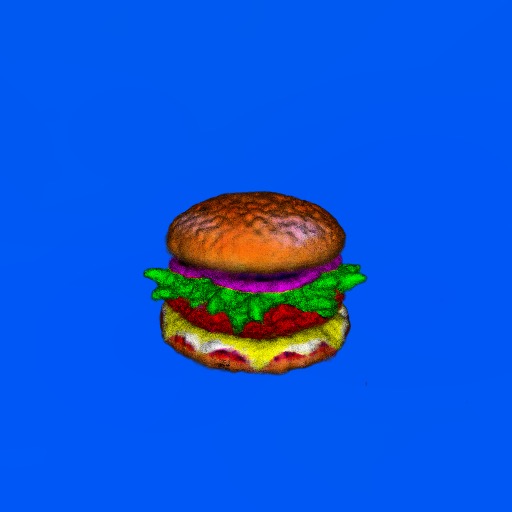} &
        \includegraphics[width=\linewidth]{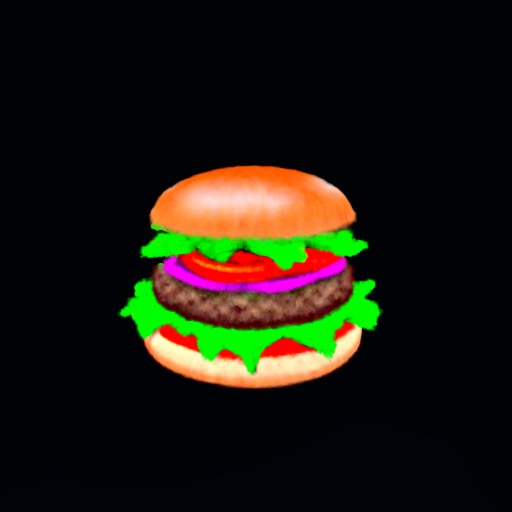} &
        \includegraphics[width=\linewidth]{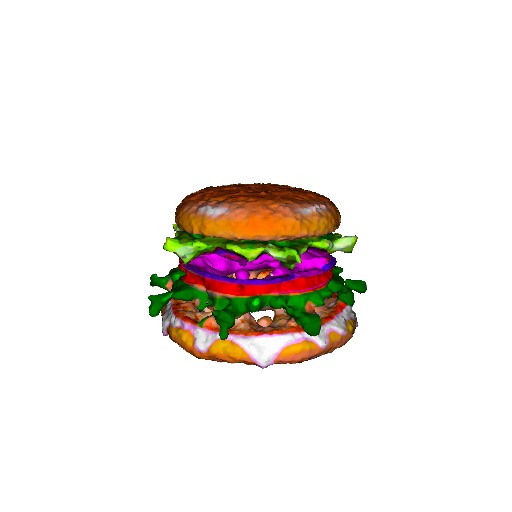} &
        \includegraphics[width=\linewidth]{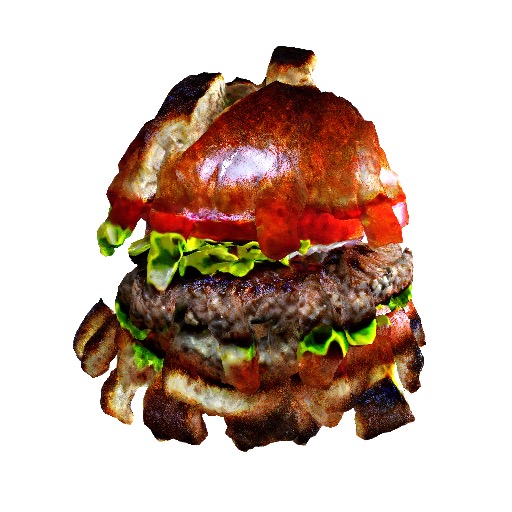} &
        \includegraphics[width=\linewidth]{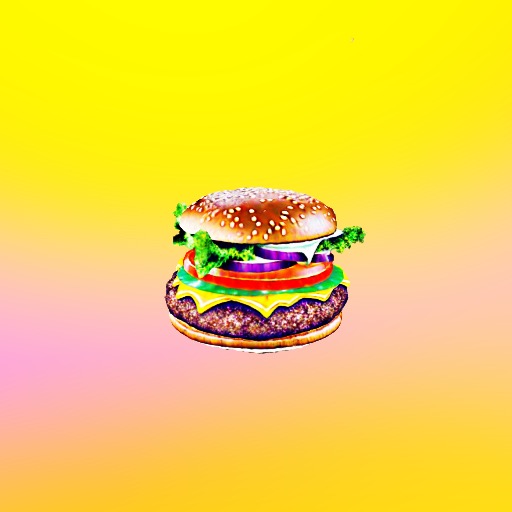} &
        \includegraphics[width=\linewidth]{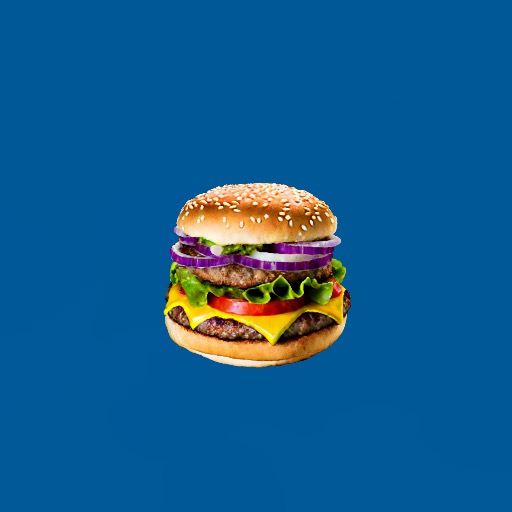} &
        \\
        \includegraphics[width=\linewidth]{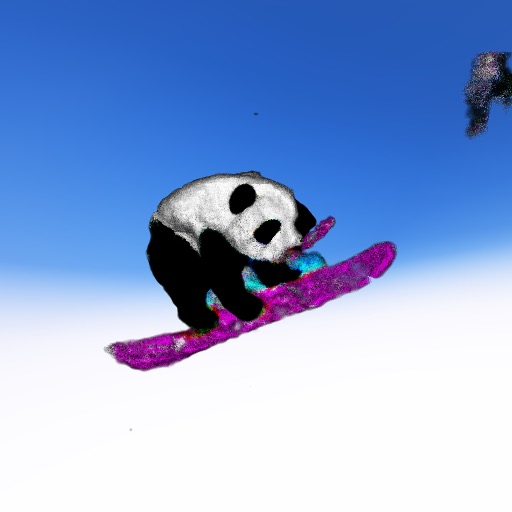} &
        \includegraphics[width=\linewidth]{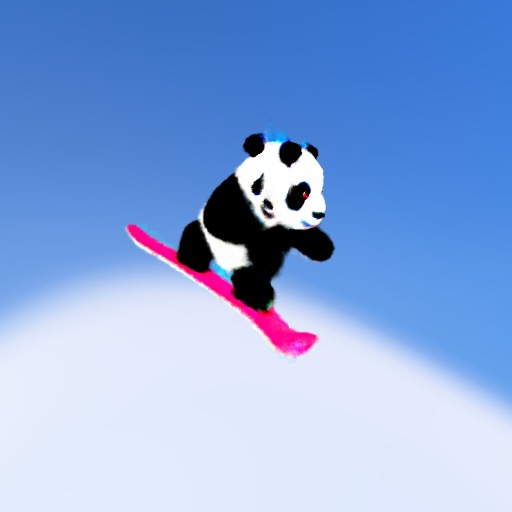} &
        \includegraphics[width=\linewidth]{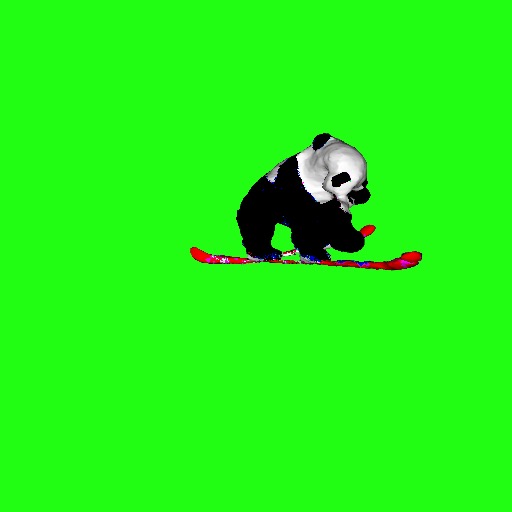} &
        \includegraphics[width=\linewidth]{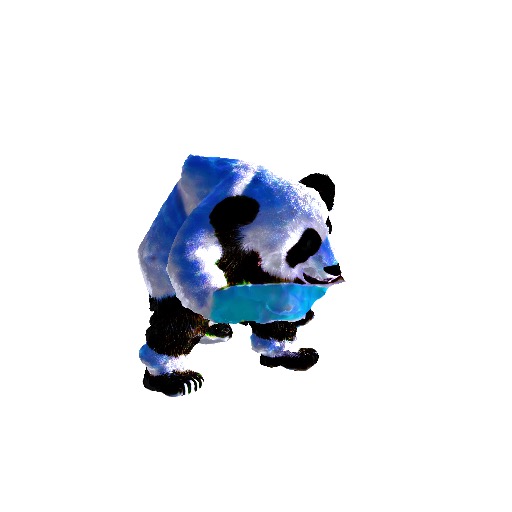} &
        \includegraphics[width=\linewidth]{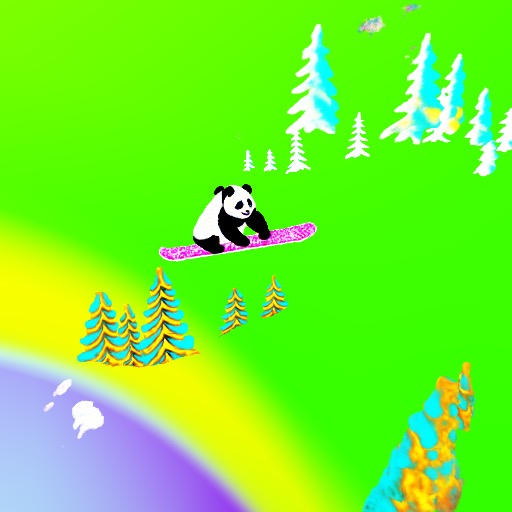} &
        \includegraphics[width=\linewidth]{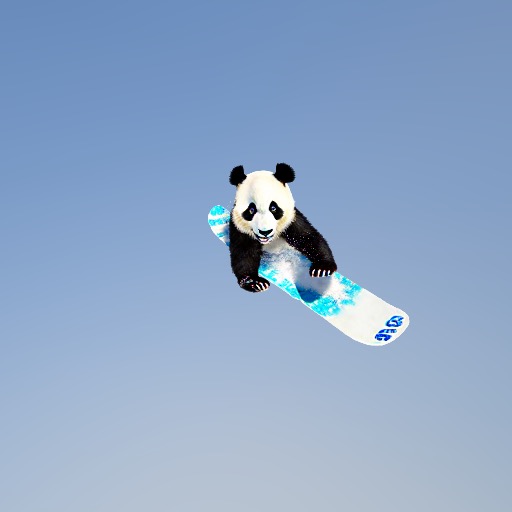} &
        \\
        DreamFusion & Latent-NeRF & Magic3D & Fantasia3D & ProlificDreamer & NFSD (ours)
    \end{tabular}
    \caption{Comparison of NFSD with other methods using the threestudio~\citep{threestudio2023} implementation.}
    \label{fig:app-method-comparison}
\end{figure}

\op{As mentioned in the main paper, methods in different papers were implemented differently. For example, the diffusion model used in each of the methods if different, the NeRF implementation may be different, etc. Hence, we present results obtained by running the unified threestudio~\citep{threestudio2023} framework implementation for all the methods. For all the methods we use Stable Diffusion 2.1-base and run them on a single GPU. As can be seen in \figref{fig:app-method-comparison}, for some of the methods the results look better with threestudio (e.g., DreamFusion), but for other methods the results may seem worse. For Magic3D~\citep{lin2023magic3d} the gap may be caused by the diffusion model, while for Fantasia3D~\citep{Chen_2023_ICCV} it may be caused by the amount of resources. The results reported by the authors of Fantasia3D were obtained by optimizing on 8 GPUs while we use a single one. Overall, our method is comparable or better than all the methods, exhibiting high resolution and detailed features.
}

\begin{figure}
    \centering
    \setlength{\tabcolsep}{1pt}
    \begin{tabular}{C{0.19\linewidth} C{0.19\linewidth} C{0.19\linewidth} C{0.19\linewidth} C{0.19\linewidth}}
        \hline \\[-8pt]
        \multicolumn{5}{c}{``A marble bust of a mouse''} \\ 
        \includegraphics[width=\linewidth]{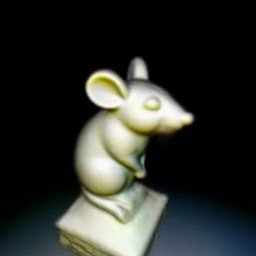} &
        \includegraphics[width=\linewidth]{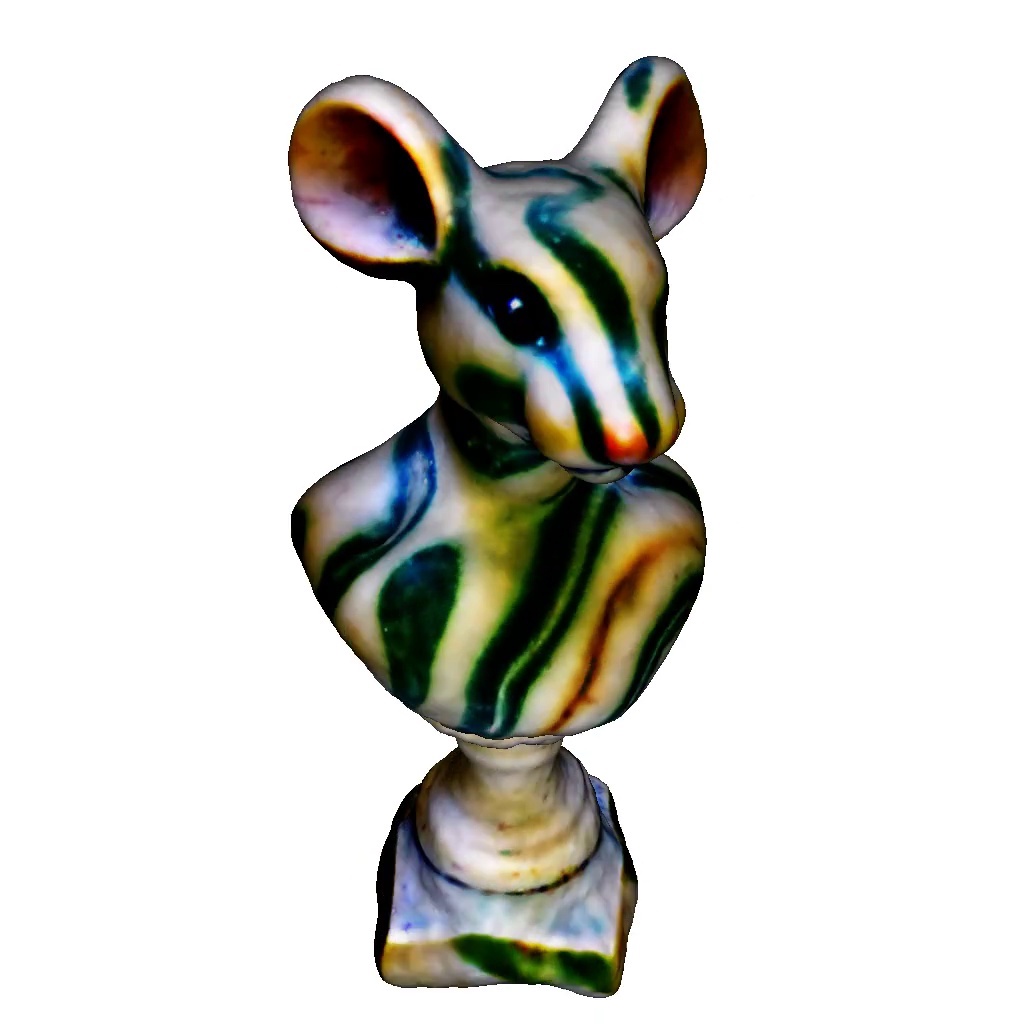} &
        \includegraphics[width=\linewidth]{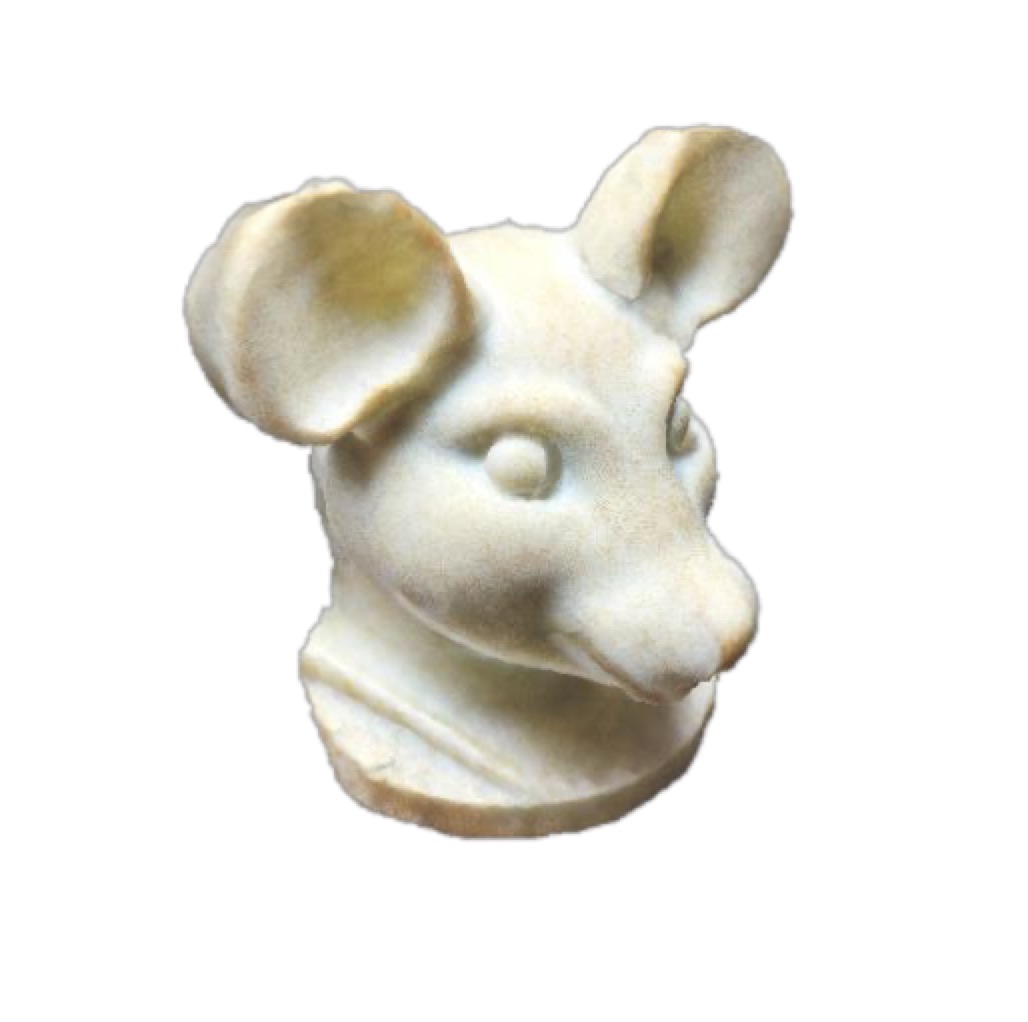} &
        \includegraphics[width=\linewidth]{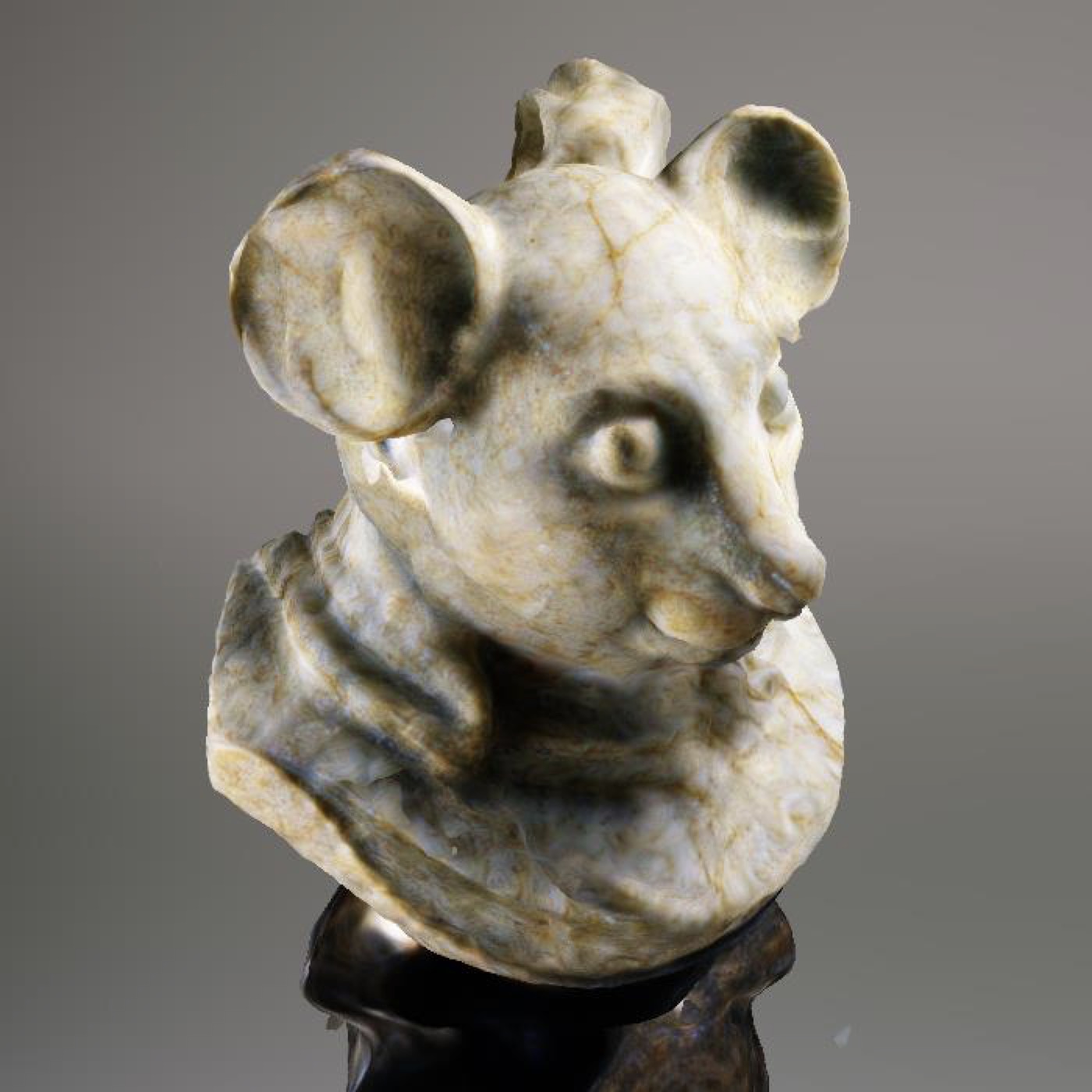} &
        \includegraphics[width=\linewidth]{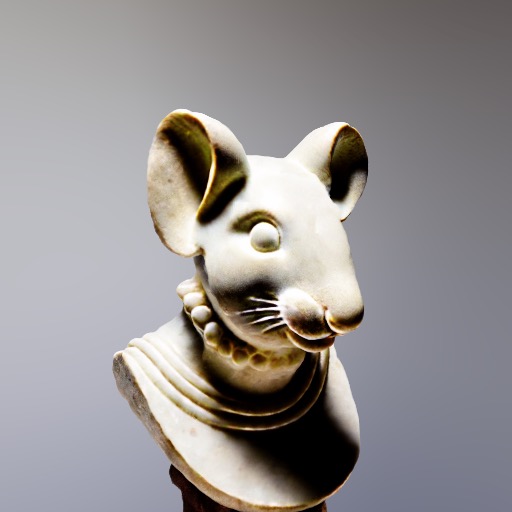} 
        \\
        DreamFusion & Magic3D & Fantasia3D & ProlificDreamer & NFSD (ours) \\
        \\[-9pt]
        \hline
        \hline \\[-8pt]
        \multicolumn{5}{c}{``A car made of sushi''} \\ 
        \includegraphics[width=\linewidth]{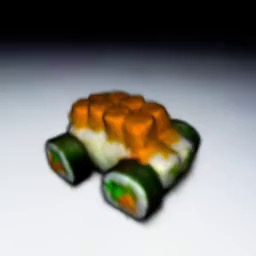} &
        \includegraphics[width=\linewidth]{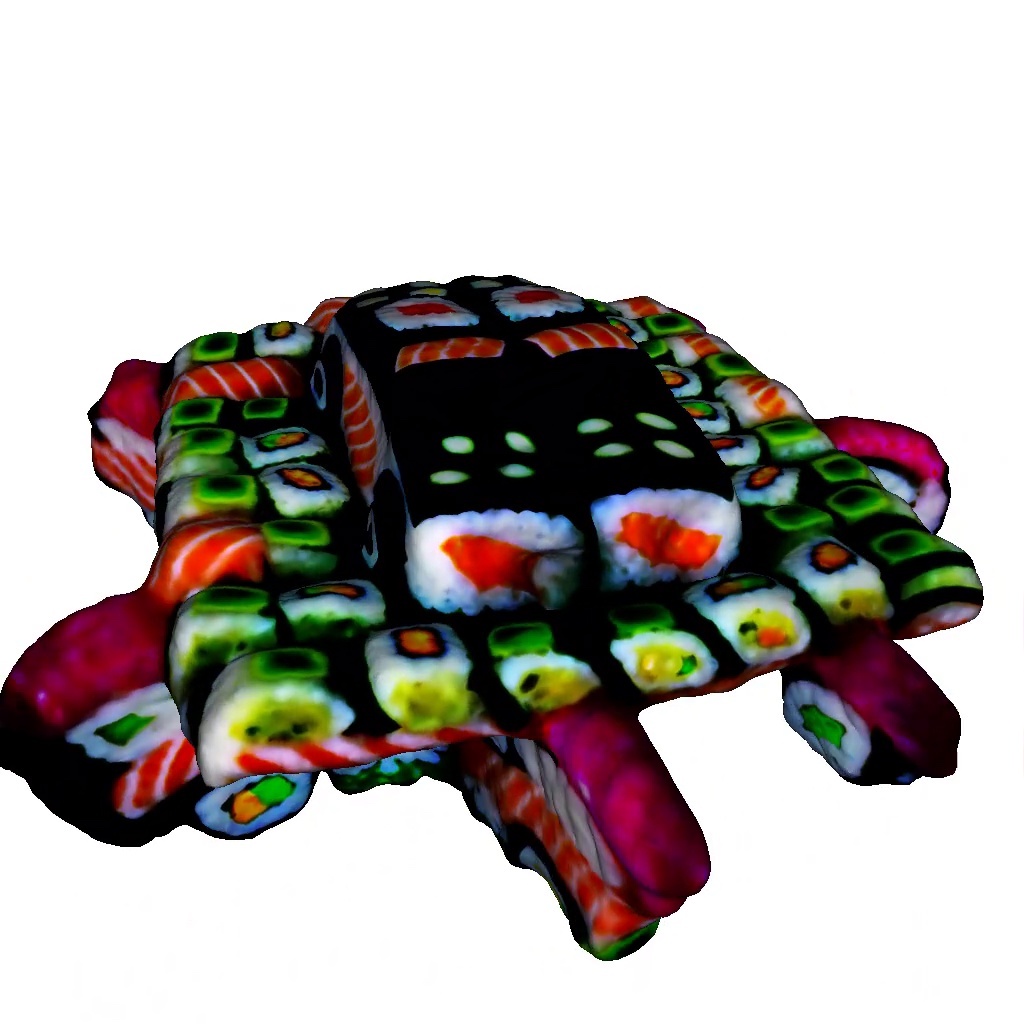} &
        \includegraphics[width=\linewidth]{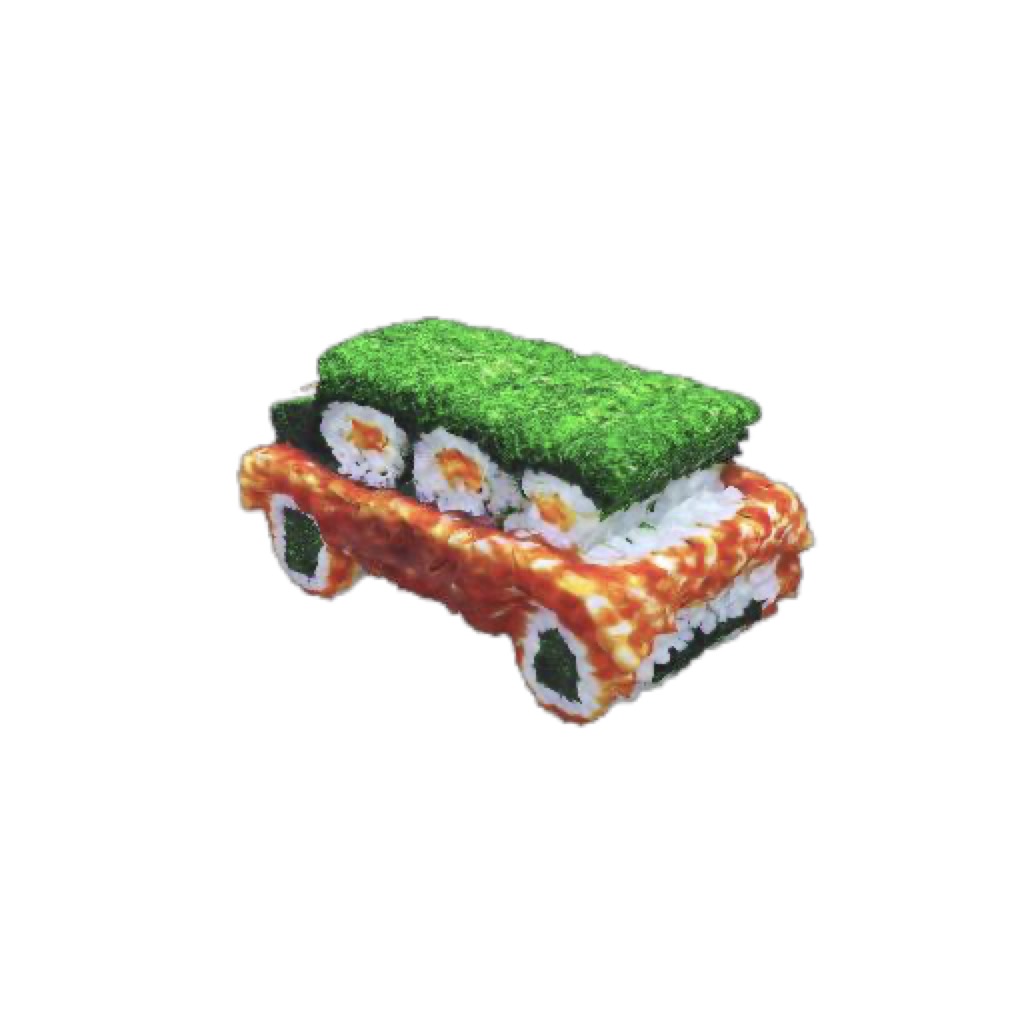} &
        \includegraphics[width=\linewidth]{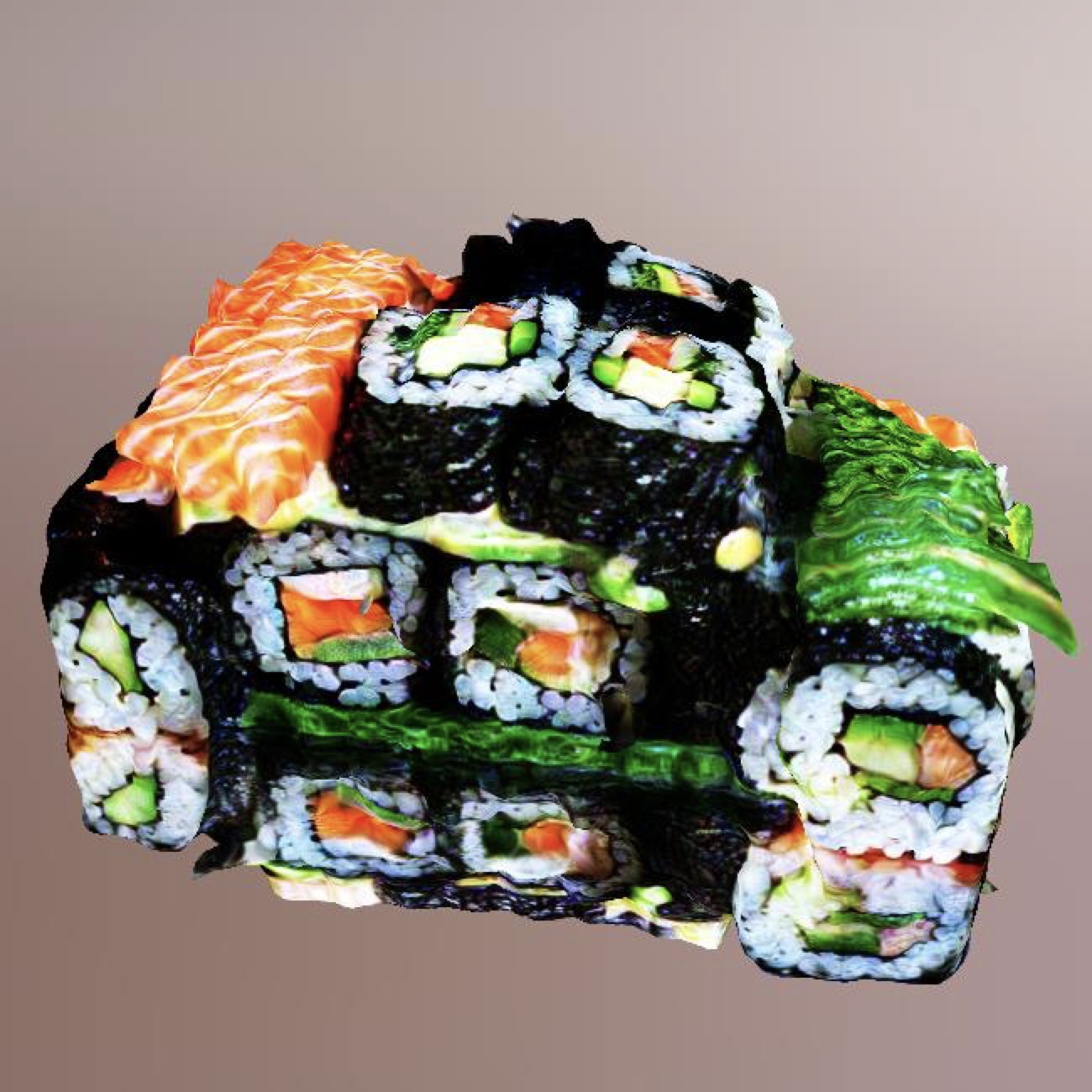} &
        \includegraphics[width=\linewidth]{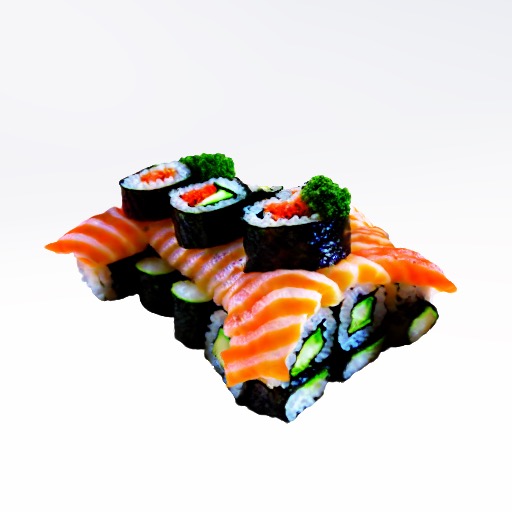} 
        \\
        DreamFusion & Magic3D & Fantasia3D & ProlificDreamer & NFSD (ours) \\
        \\[-9pt]
        \hline
    \end{tabular}
    \caption{Comparison of NFSD with other methods using results obtained from the original papers.}
    \label{fig:app-comp-orig-1a}
\end{figure}

\begin{figure}
    \centering
    \setlength{\tabcolsep}{1pt}
    \begin{tabular}{C{0.24\linewidth} C{0.24\linewidth} C{0.24\linewidth} C{0.24\linewidth}}
        \hline \\[-8pt]
        \multicolumn{4}{c}{``A ripe strawberry''} \\ 
        \includegraphics[width=\linewidth]{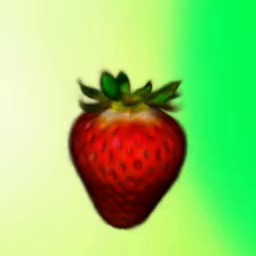} &
        \includegraphics[width=\linewidth]{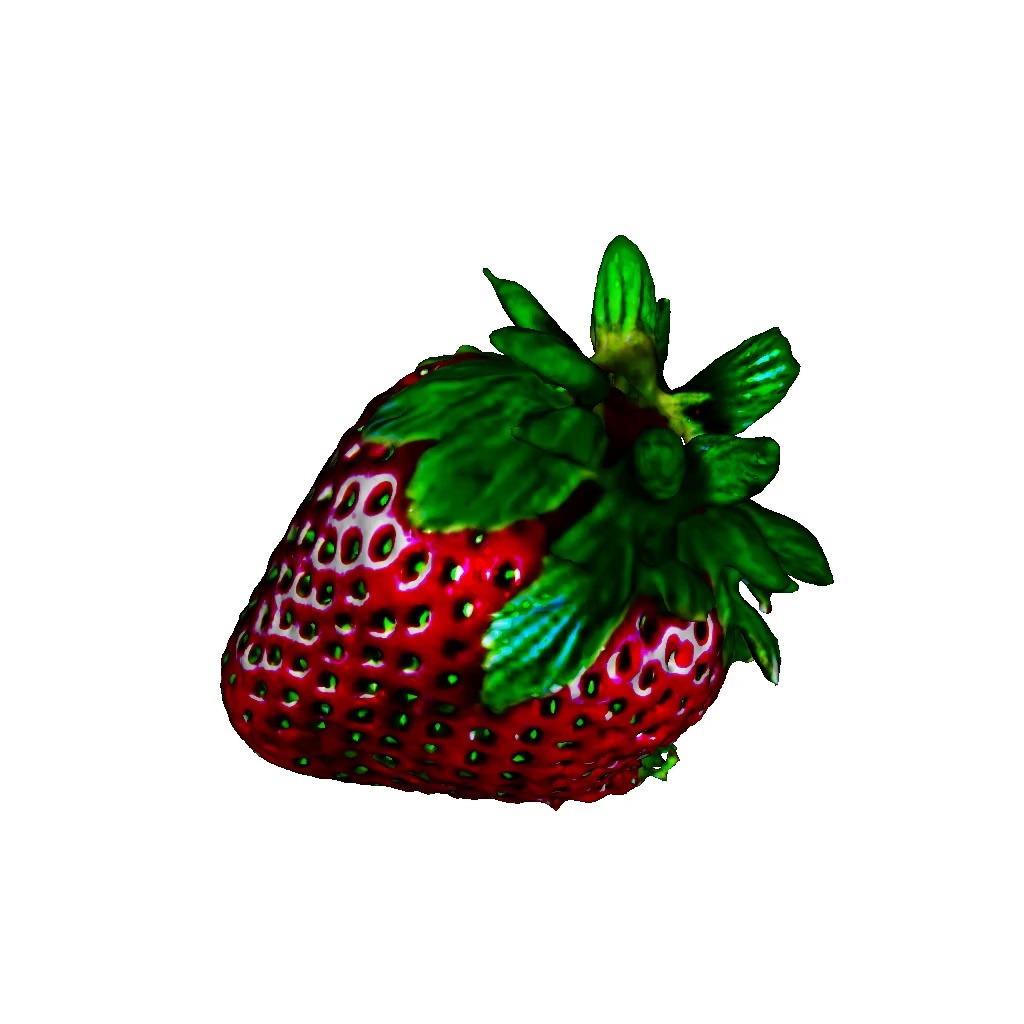} &
        \includegraphics[width=\linewidth]{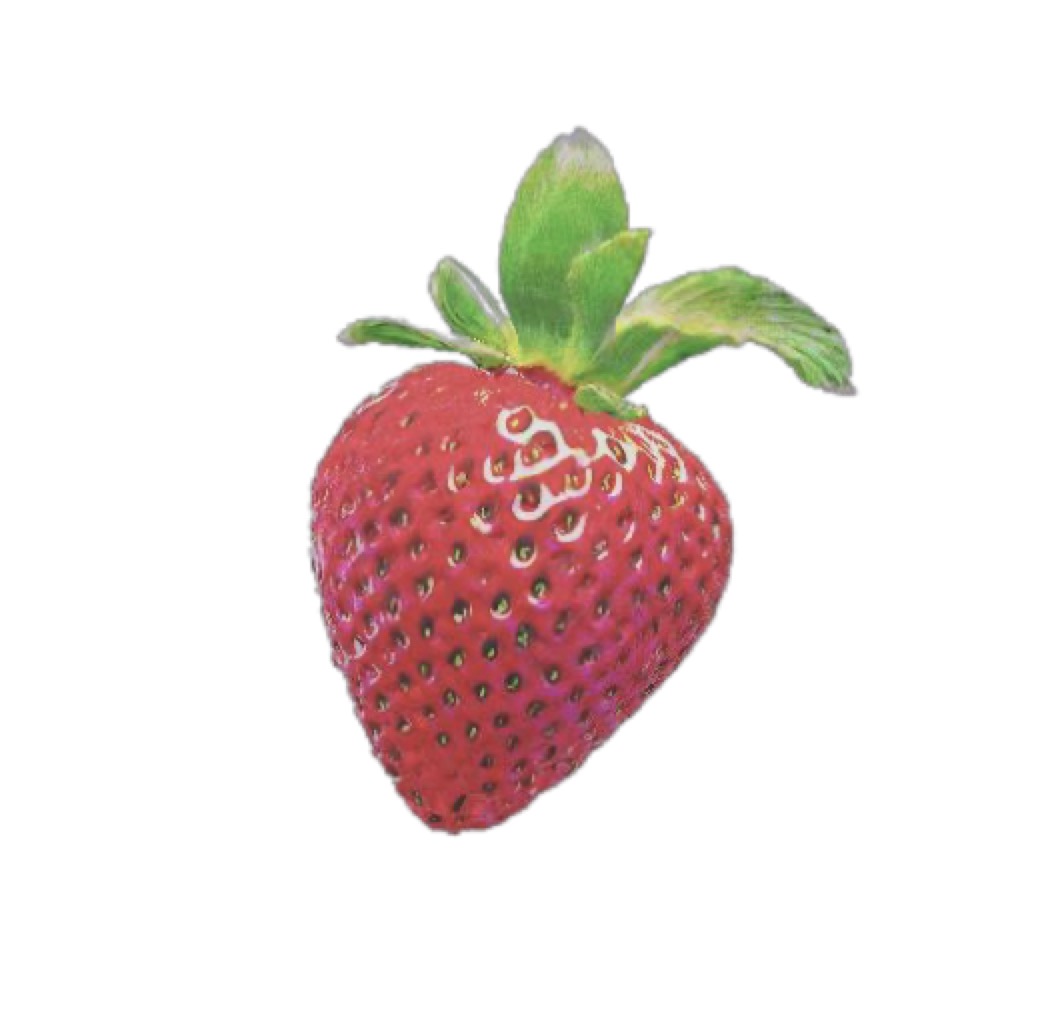} &
        \includegraphics[width=\linewidth]{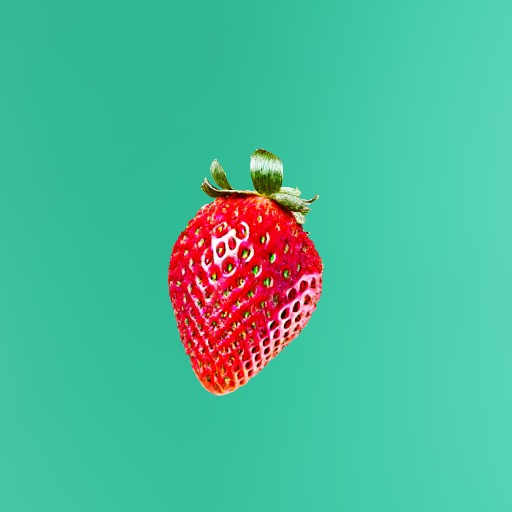} 
        \\
        DreamFusion & Magic3D & Fantasia3D  & NFSD (ours) \\
        \\[-9pt]
        \hline
        \hline \\[-8pt]
        \multicolumn{4}{c}{``A small saguaro cactus planted in a clay pot''} \\ 
        \includegraphics[width=\linewidth]{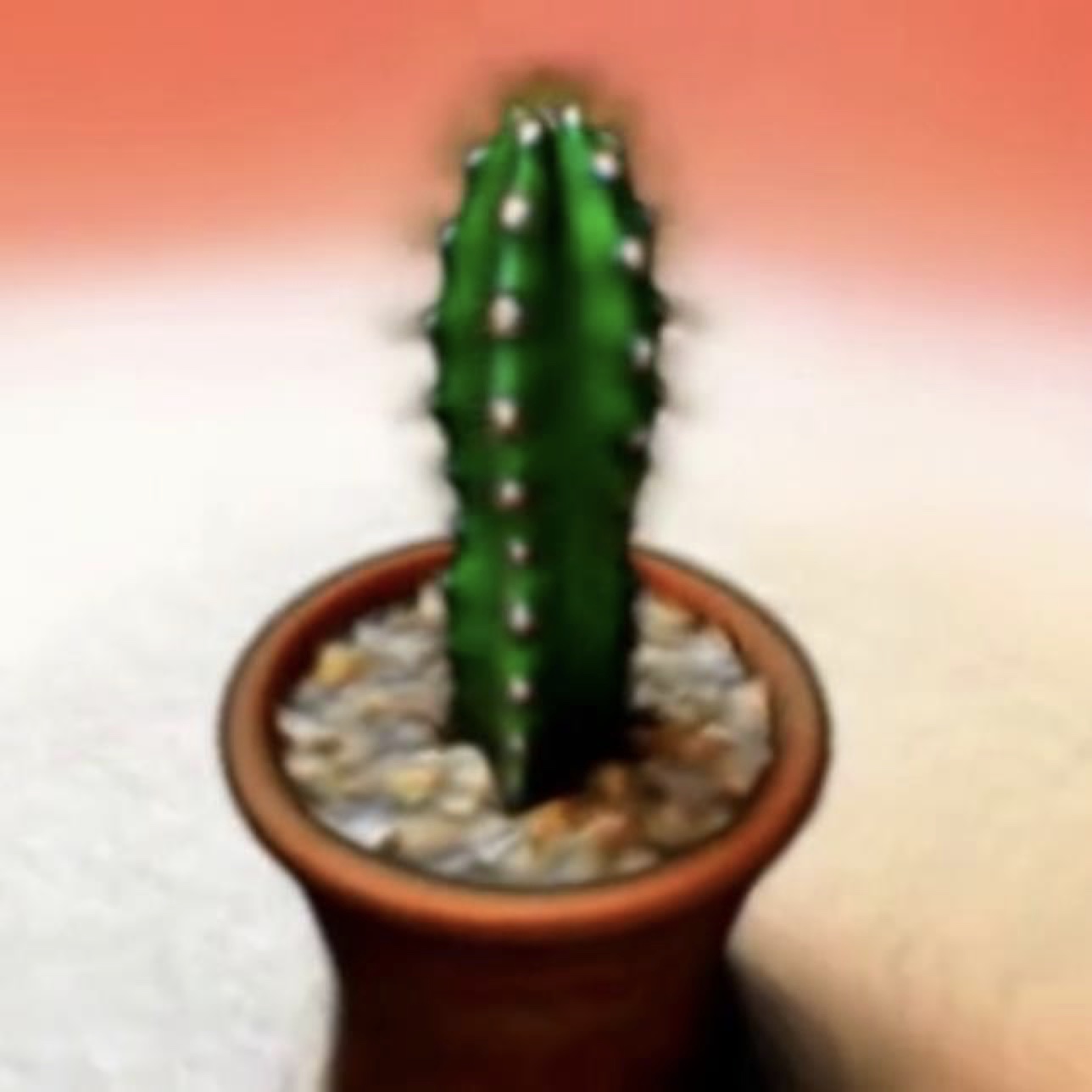} &
        \includegraphics[width=\linewidth]{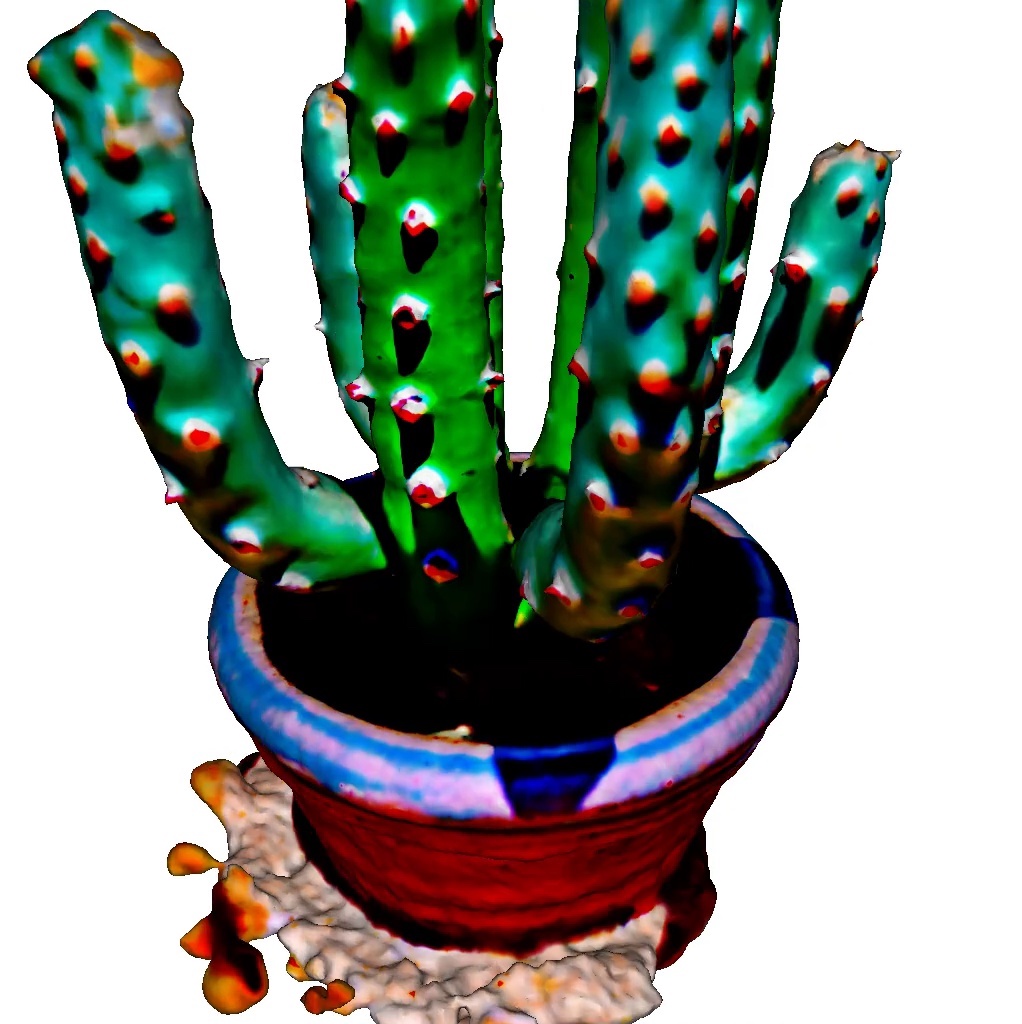} &
        \includegraphics[width=\linewidth]{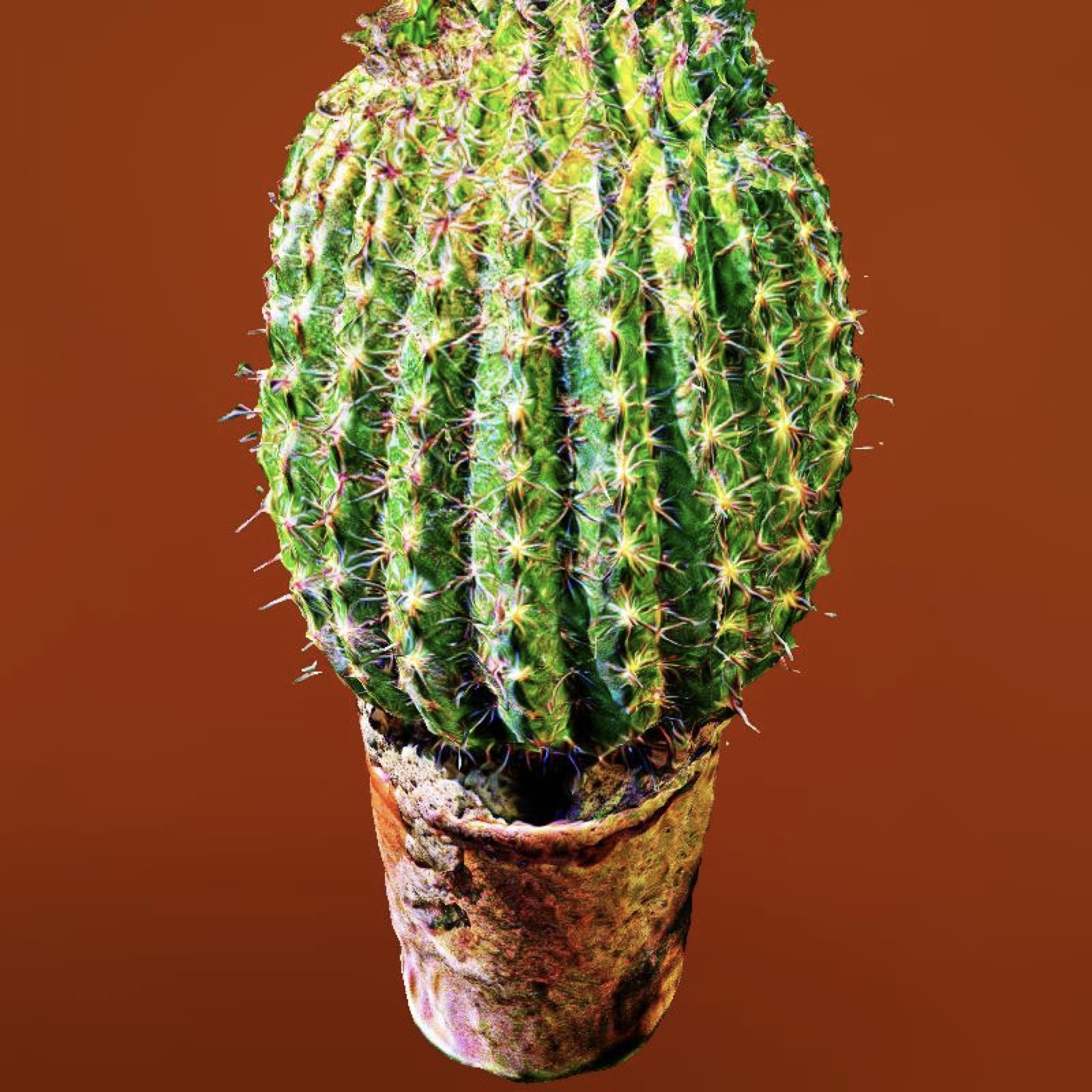} &
        \includegraphics[width=\linewidth]{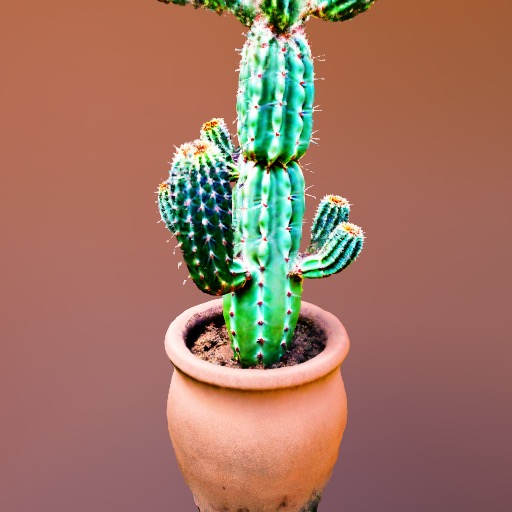} 
        \\
        DreamFusion & Magic3D & ProlificDreamer  & NFSD (ours) \\
        \\[-9pt]
        \hline
        \hline \\[-8pt]
        \multicolumn{4}{c}{``A rabbit, animated movie character, high detail 3d mode''} \\ 
        \includegraphics[width=\linewidth]{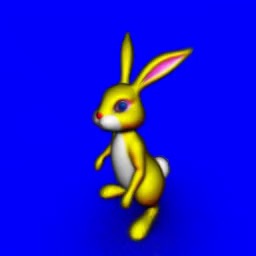} &
        \includegraphics[width=\linewidth]{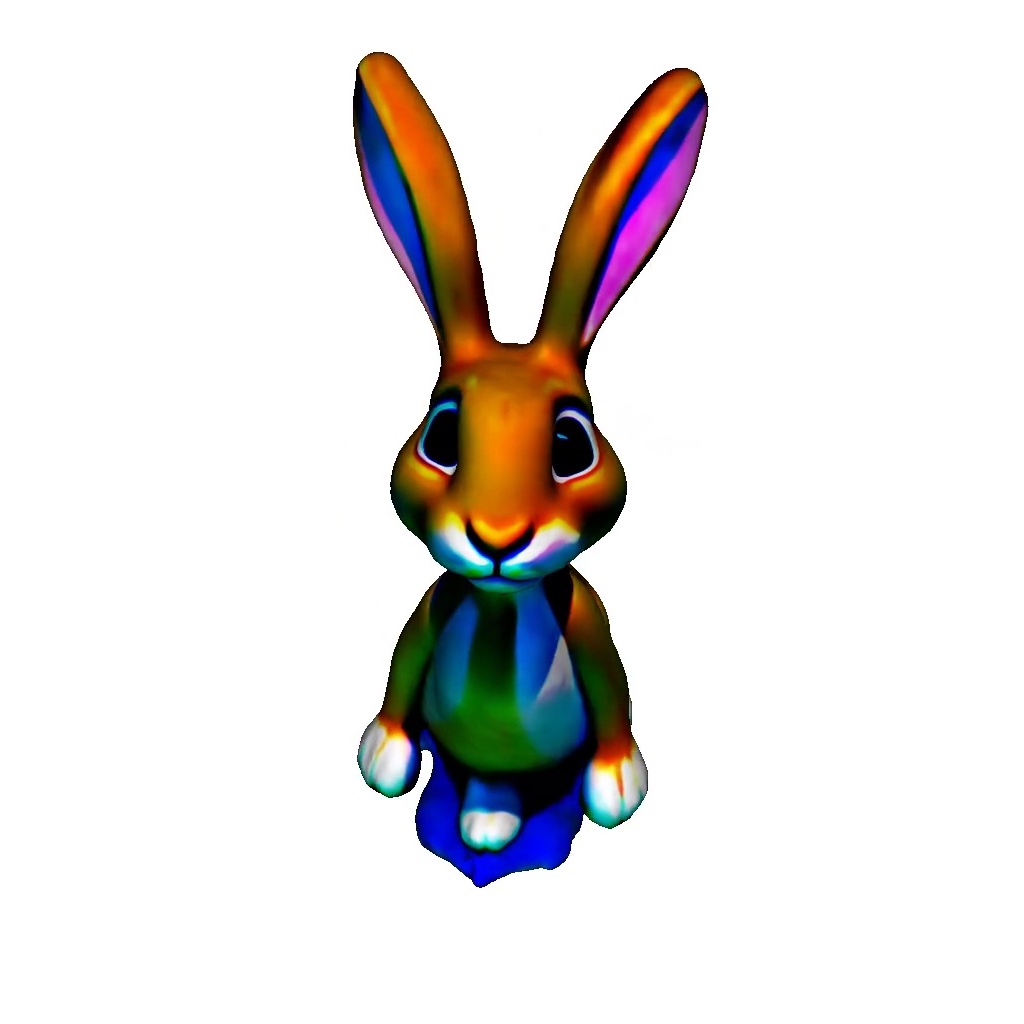} &
        \includegraphics[width=\linewidth]{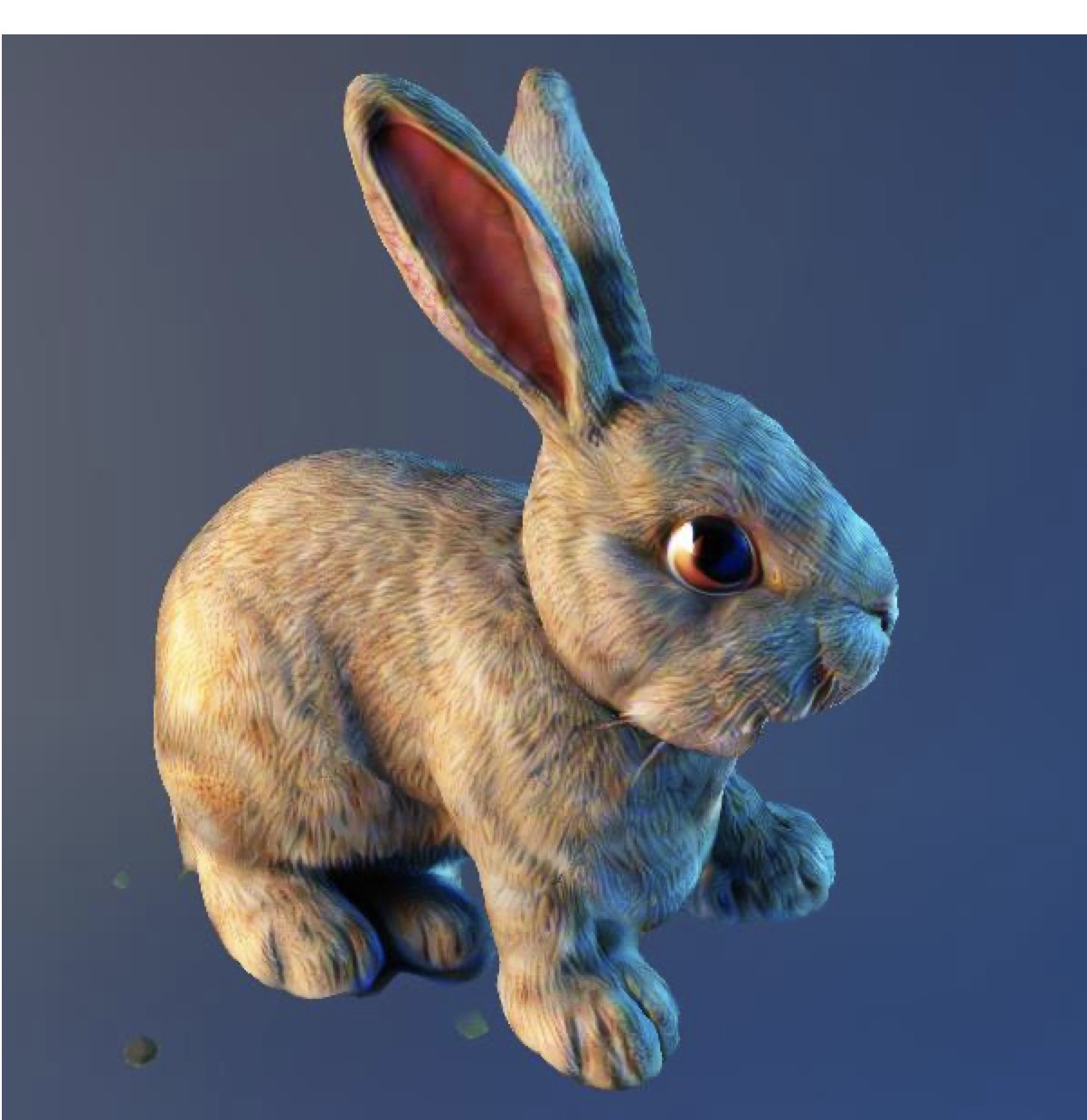} &
        \includegraphics[width=\linewidth]{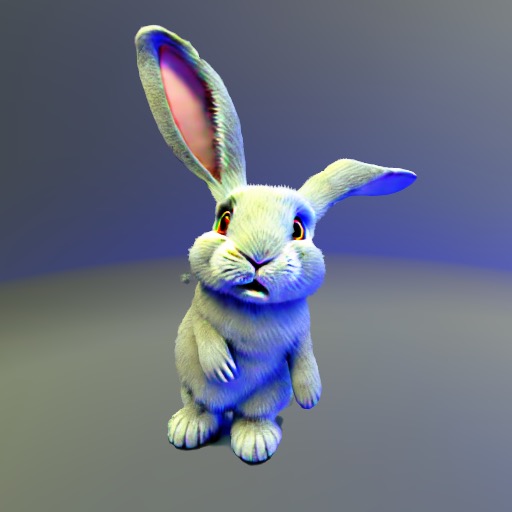} 
        \\
        DreamFusion & Magic3D & ProlificDreamer  & NFSD (ours) \\
        \\[-9pt]
        \hline
    \end{tabular}
    \caption{Comparison of NFSD with other methods using results obtained from the original papers.}
    \label{fig:app-comp-orig-1}
\end{figure}

\begin{figure}
    \centering
    \setlength{\tabcolsep}{1pt}
    \begin{tabular}{C{0.24\linewidth} C{0.24\linewidth} C{0.24\linewidth} C{0.24\linewidth}}
        \hline \\[-8pt]
        \multicolumn{4}{c}{``A stack of pancakes covered in maple syrup''} \\ 
        \includegraphics[width=\linewidth]{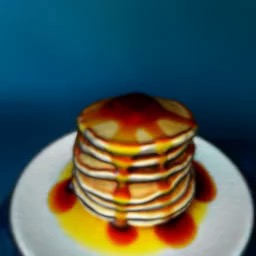} &
        \includegraphics[width=\linewidth]{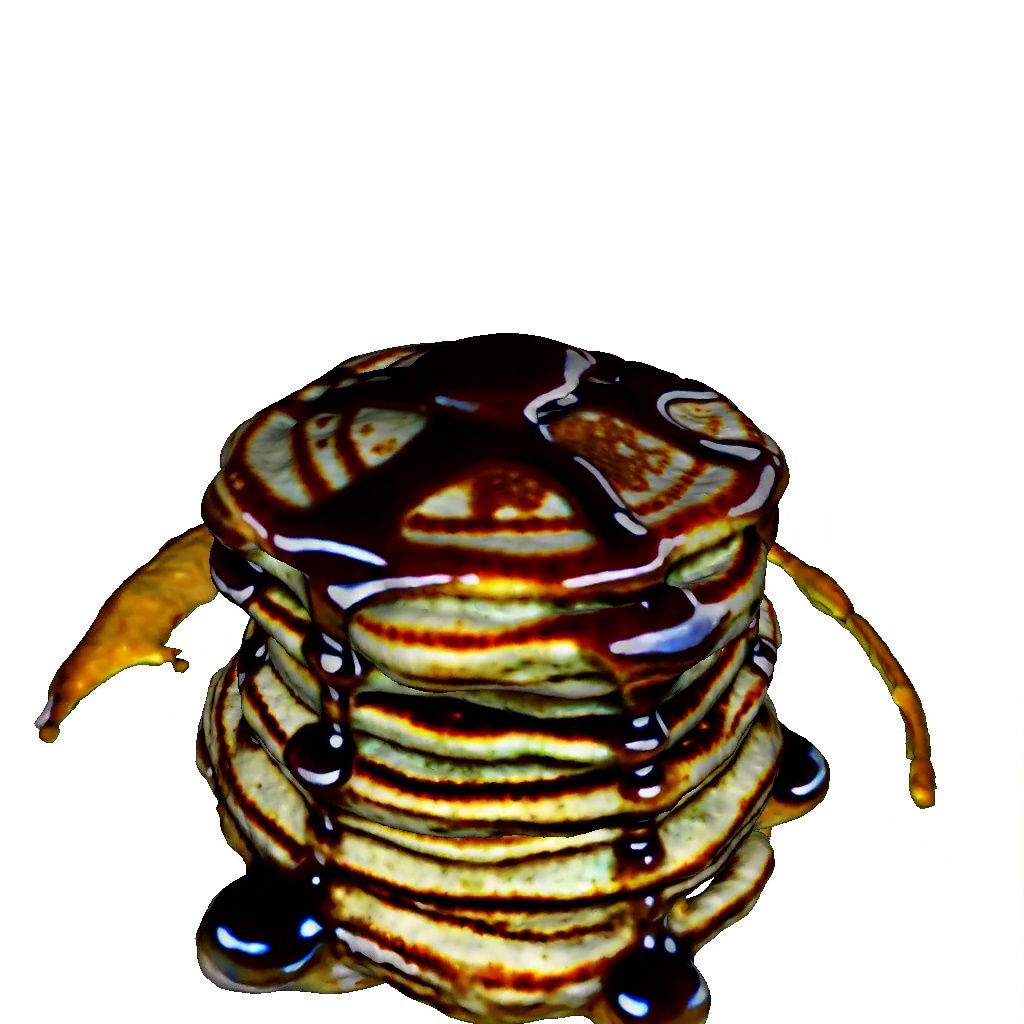} &
        \includegraphics[width=\linewidth]{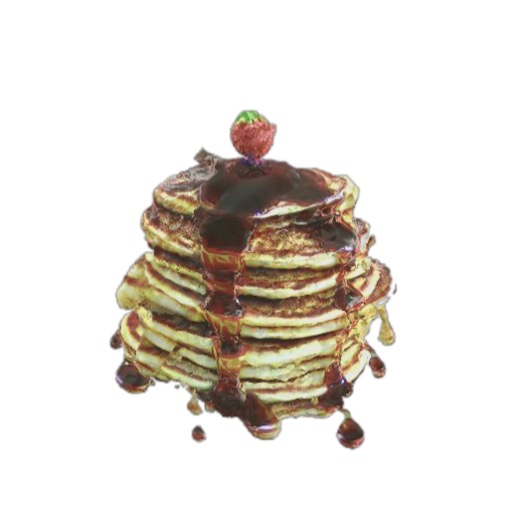} &
        \includegraphics[width=\linewidth]{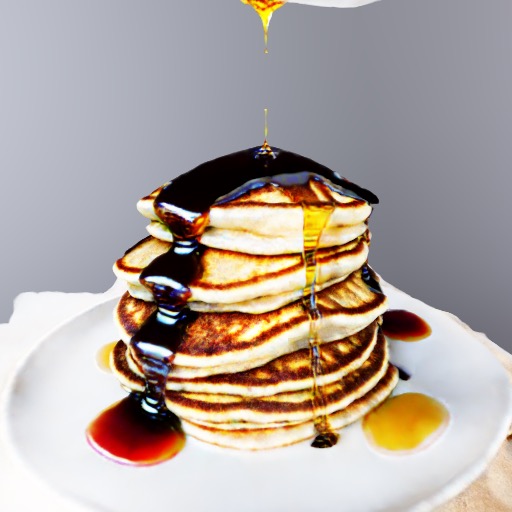} 
        \\
        DreamFusion & Magic3D & Fantasia3D  & NFSD (ours) \\
        \\[-9pt]
        \hline
        \hline \\[-8pt]
        \multicolumn{4}{c}{``A delicious croissan''} \\ 
        \includegraphics[width=\linewidth]{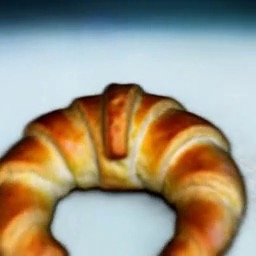} &
        \includegraphics[width=\linewidth]{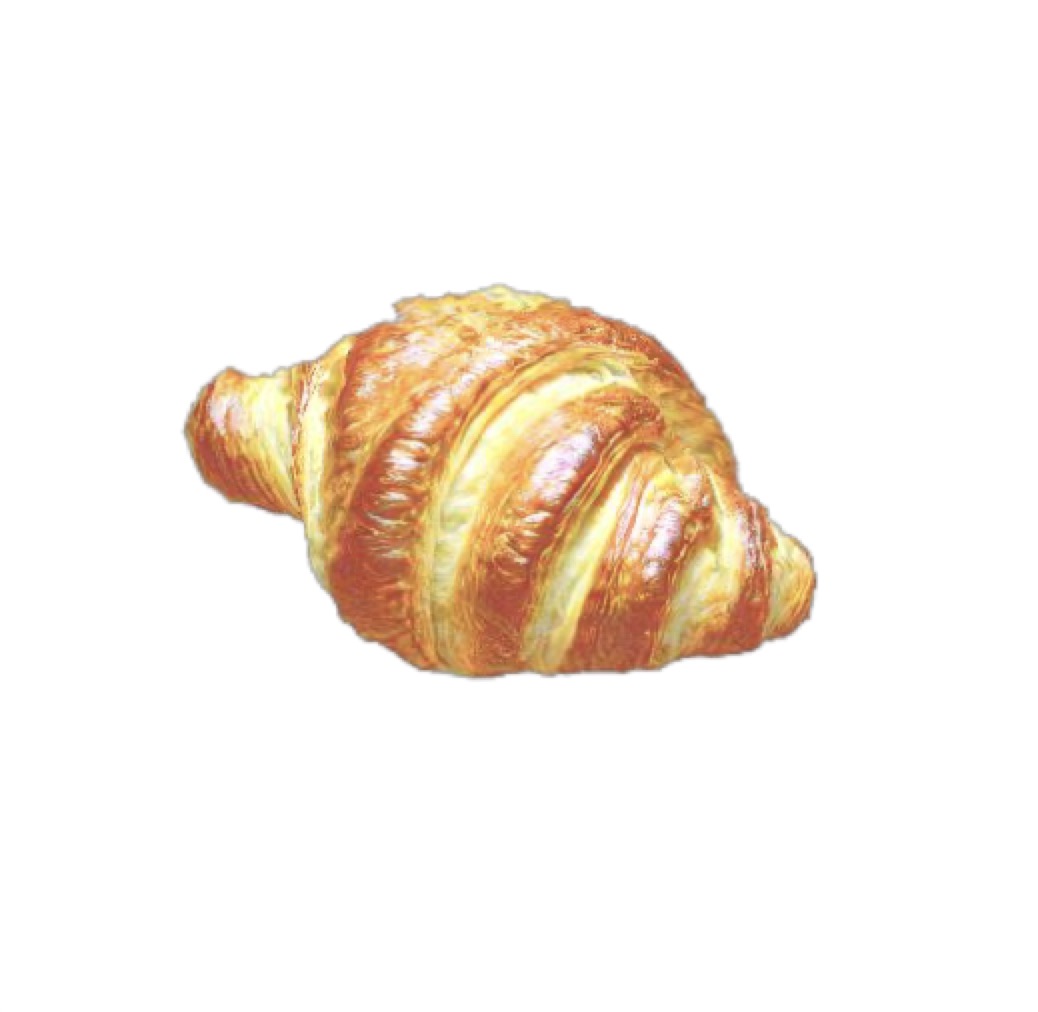} &
        \includegraphics[width=\linewidth]{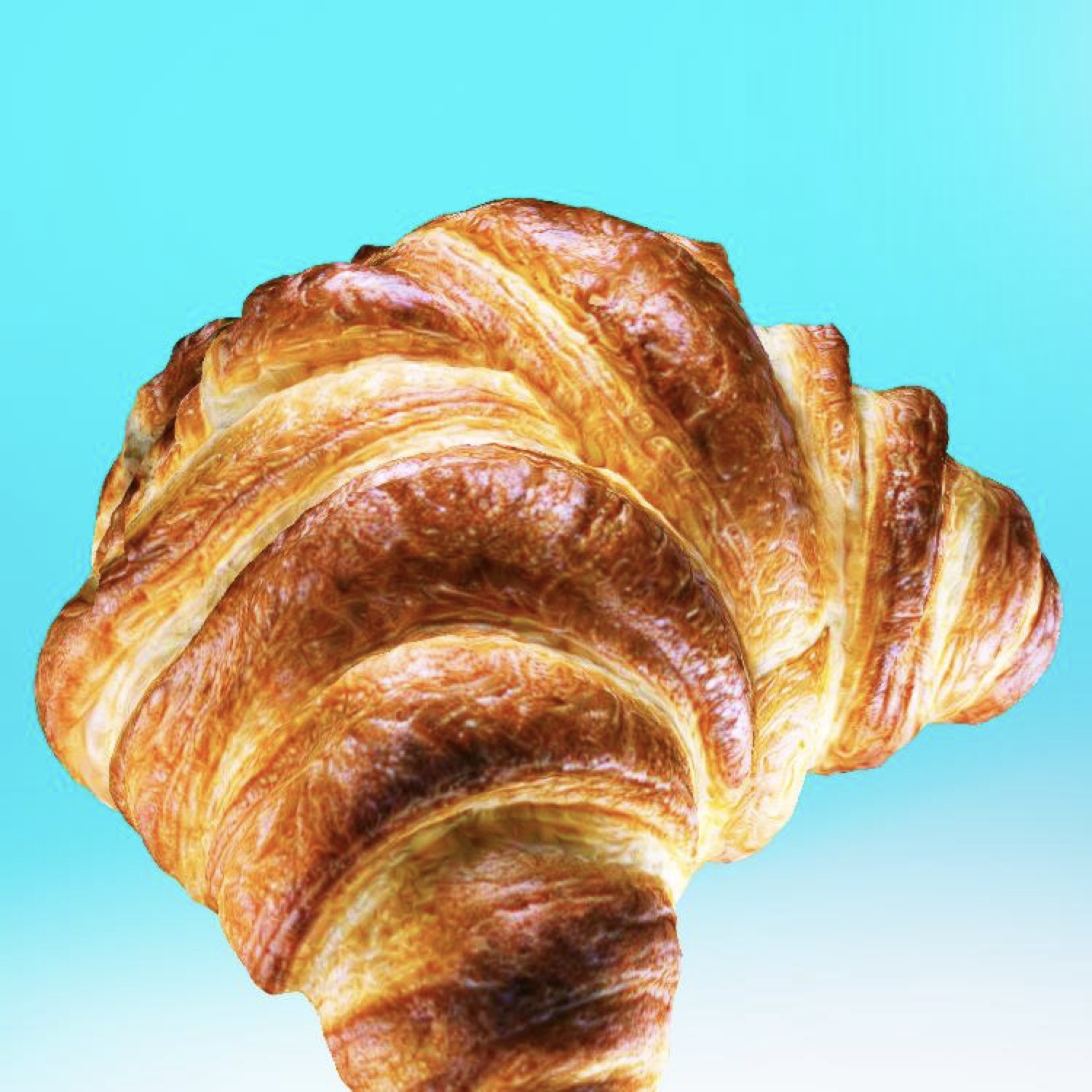} &
        \includegraphics[width=\linewidth]{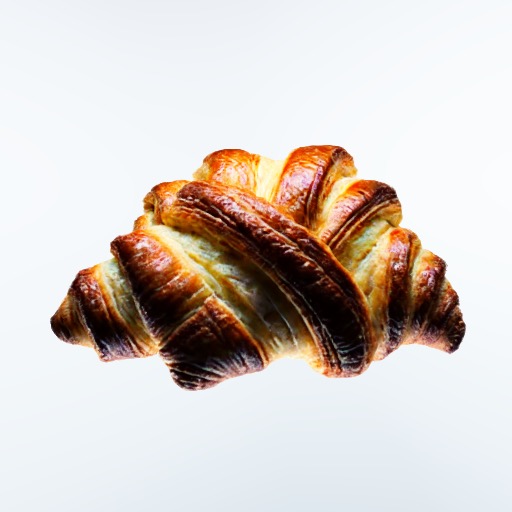} 
        \\
        DreamFusion & Fantasia3D & ProlificDreamer  & NFSD (ours) \\
        \\[-9pt]
        \hline
        \hline \\[-8pt]
        \multicolumn{4}{c}{``An ice cream sundae''} \\ 
        \includegraphics[width=\linewidth]{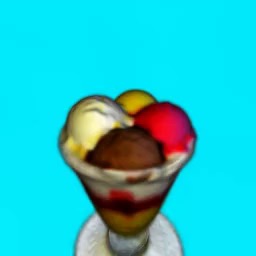} &
        \includegraphics[width=\linewidth]{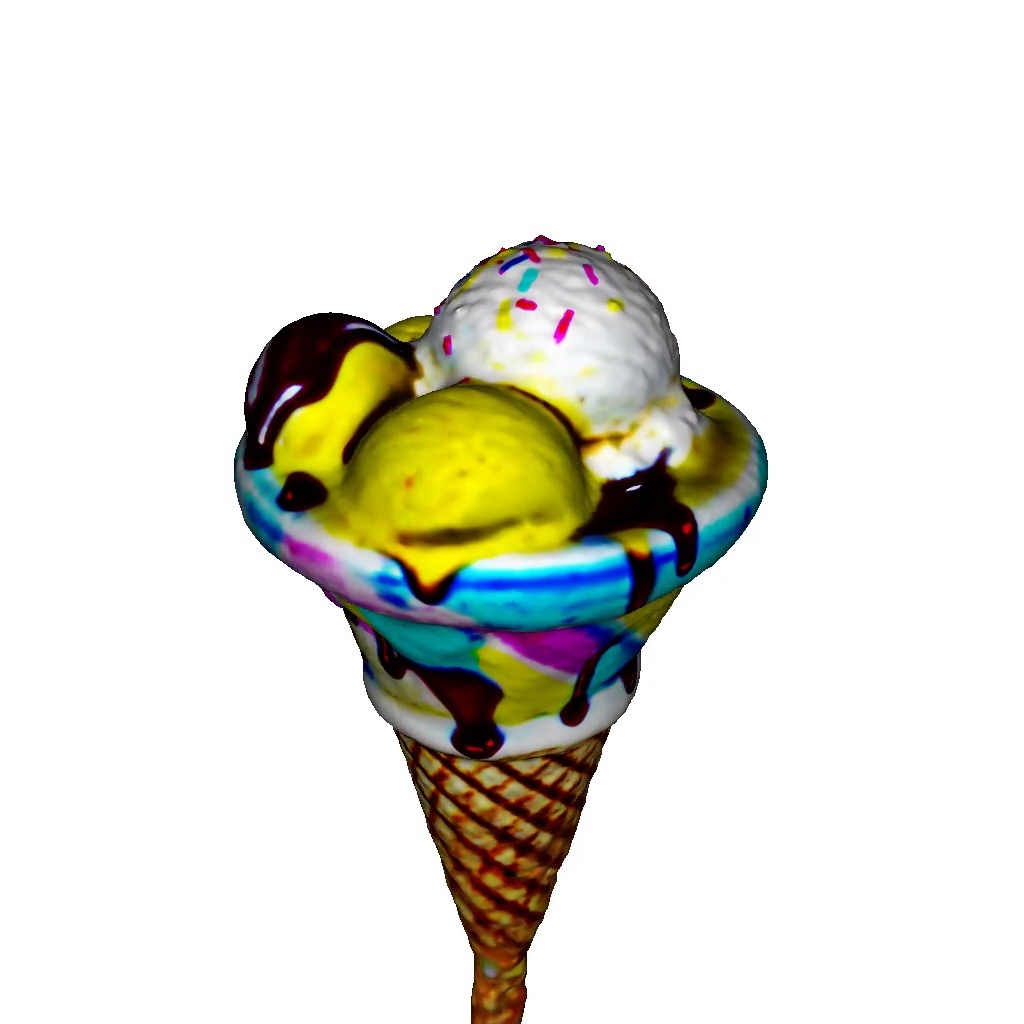} &
        \includegraphics[width=\linewidth]{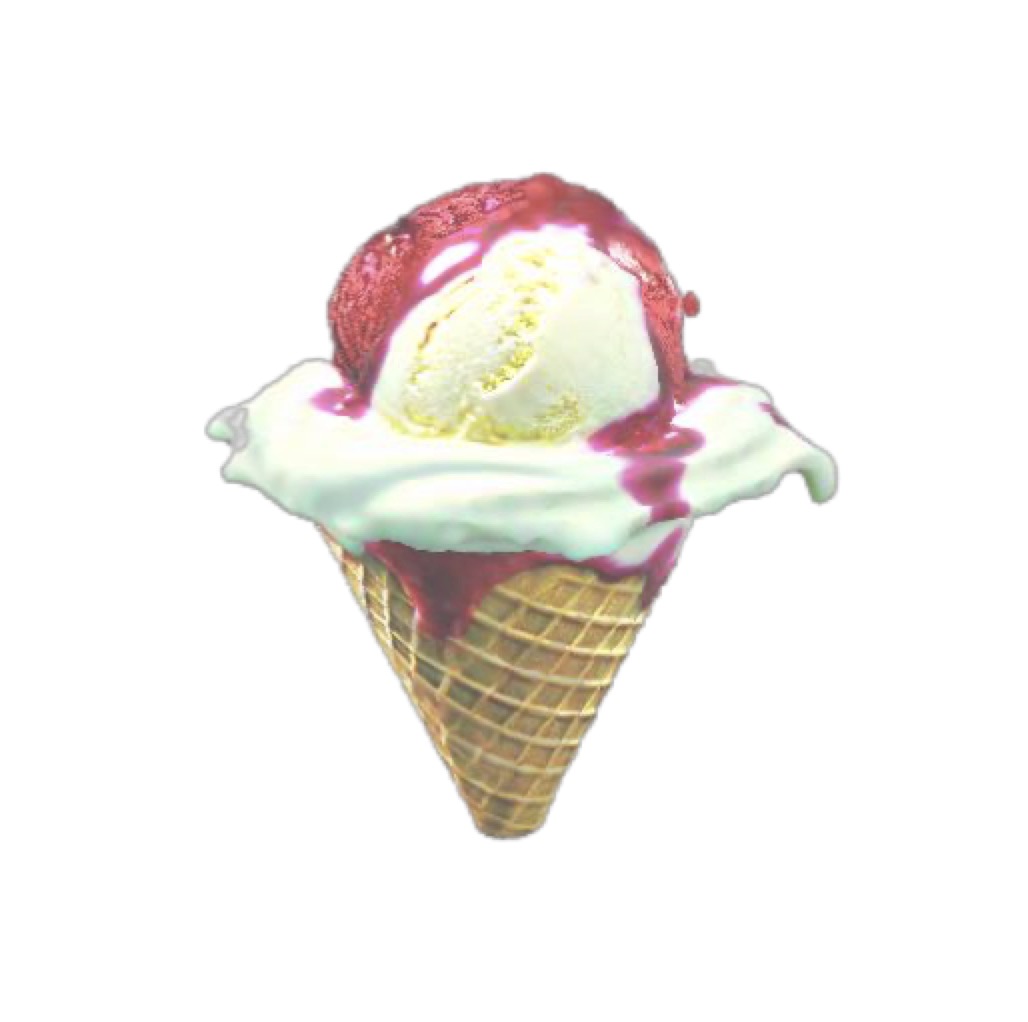} &
        \includegraphics[width=\linewidth]{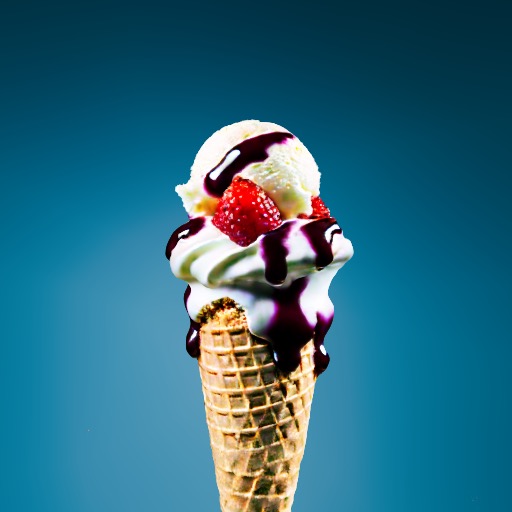} 
        \\
        DreamFusion & Magic3D & Fantasia3D  & NFSD (ours) \\
        \\[-9pt]
        \hline
                \hline \\[-8pt]
        \multicolumn{4}{c}{``A baby bunny sitting on top of a stack of pancakes''} \\ 
        \includegraphics[width=\linewidth]{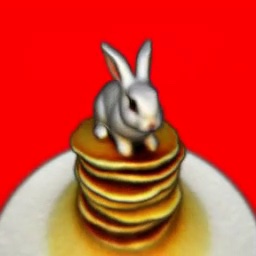} &
        \includegraphics[width=\linewidth]{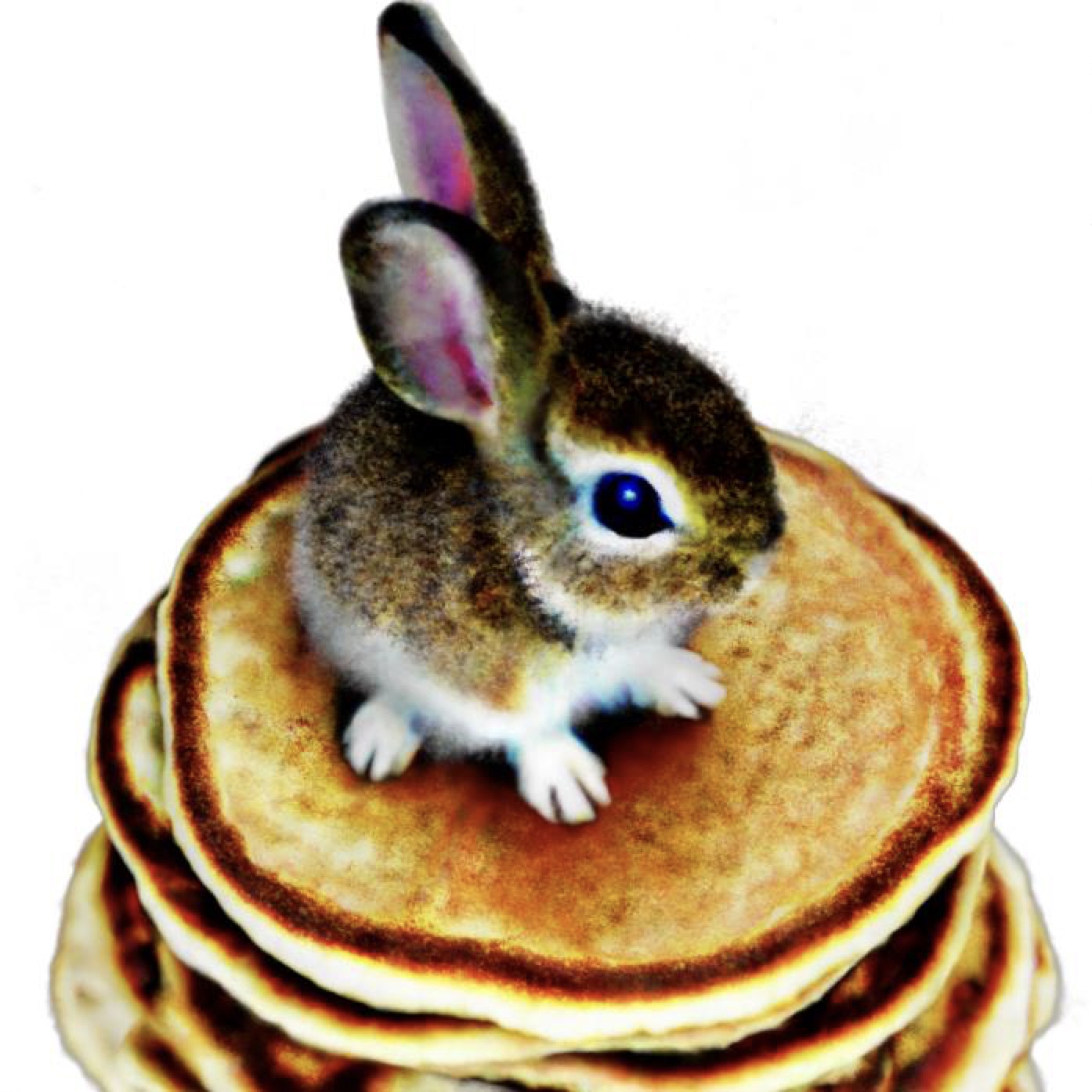} &
        \includegraphics[width=\linewidth]{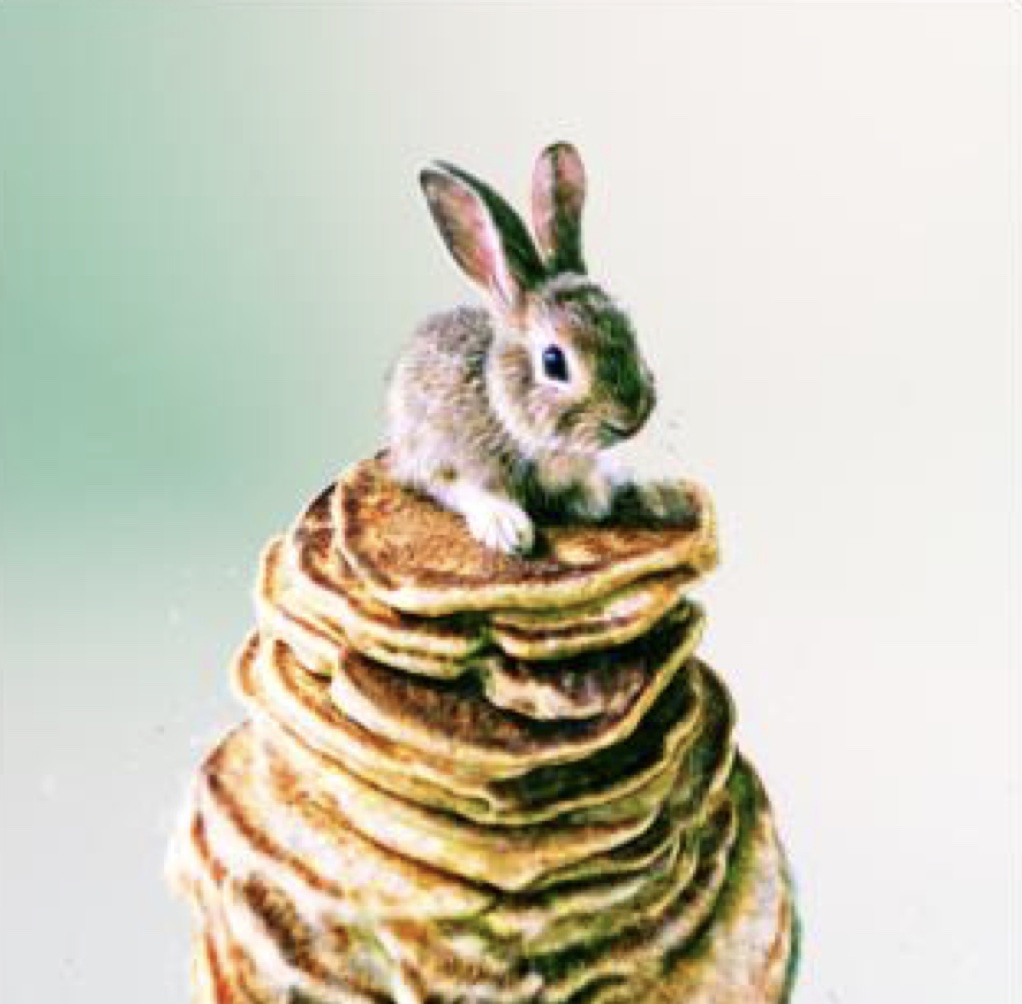} &
        \includegraphics[width=\linewidth]{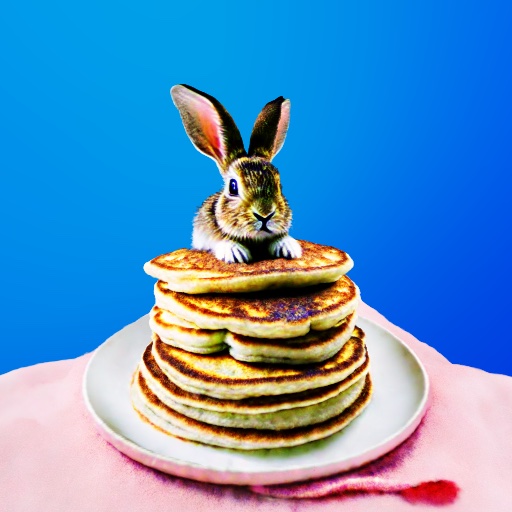} 
        \\
        DreamFusion & Magic3D & ProlificDreamer  & NFSD (ours) \\
        \\[-9pt]
        \hline
         \hline \\[-8pt]
        \multicolumn{4}{c}{``A hamburger''} \\ 
        \includegraphics[width=\linewidth]{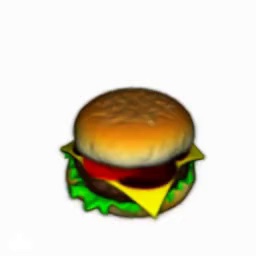} &
        \includegraphics[width=\linewidth]{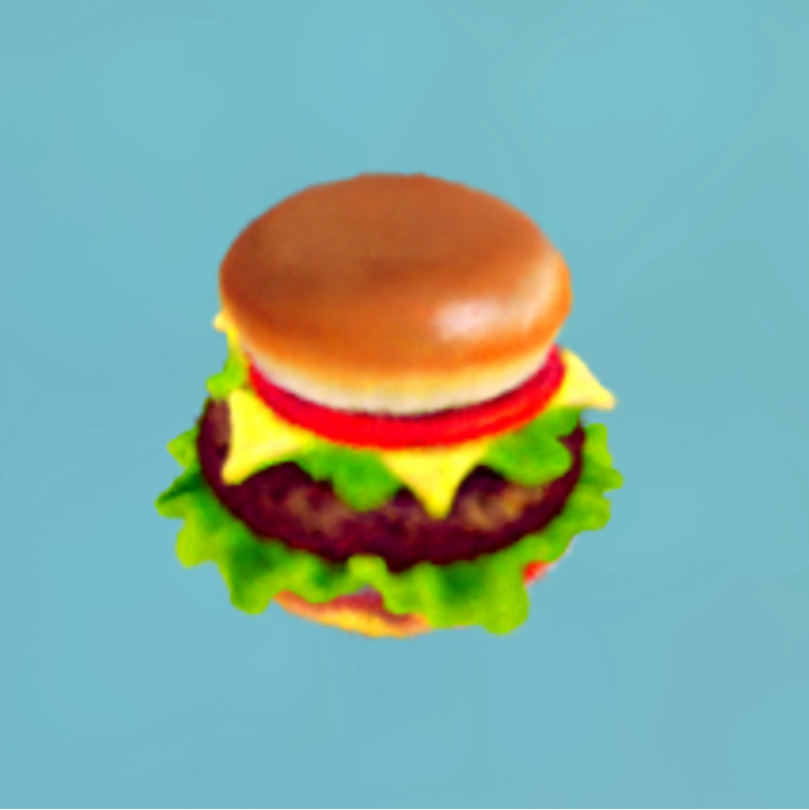} &
        \includegraphics[width=\linewidth]{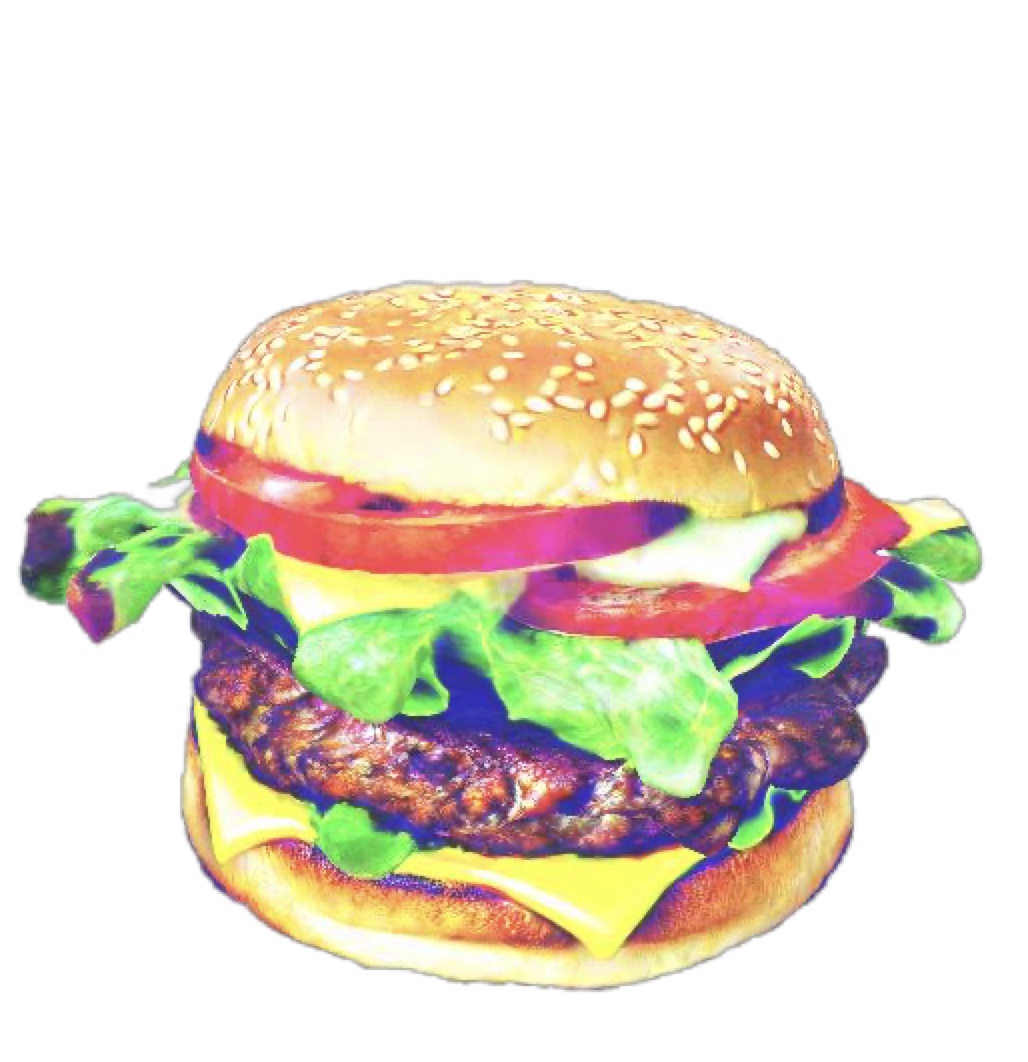} &
        \includegraphics[width=\linewidth]{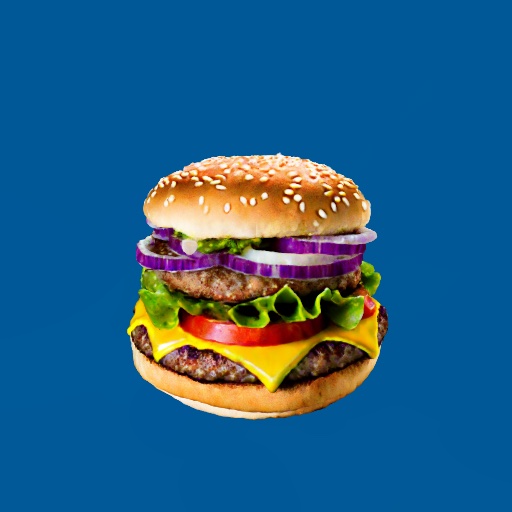} 
        \\
        DreamFusion & LatentNerf & Fantasia  & NFSD (ours) \\
        \\[-9pt]
        \hline
    \end{tabular}
    \caption{Comparison of NFSD with other methods using results obtained from the original papers.}
    \label{fig:app-comp-orig-2a}
\end{figure}

\begin{figure}
    \centering
    \setlength{\tabcolsep}{1pt}
    \begin{tabular}{C{0.33\linewidth} C{0.33\linewidth} C{0.33\linewidth}}
        \hline \\[-8pt]
        \multicolumn{3}{c}{``Baby dragon hatching out of a stone egg''} \\ 
        \includegraphics[width=\linewidth]{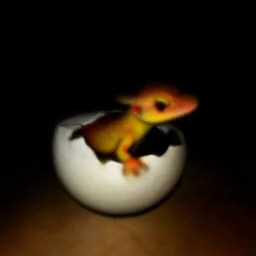} &
        \includegraphics[width=\linewidth]{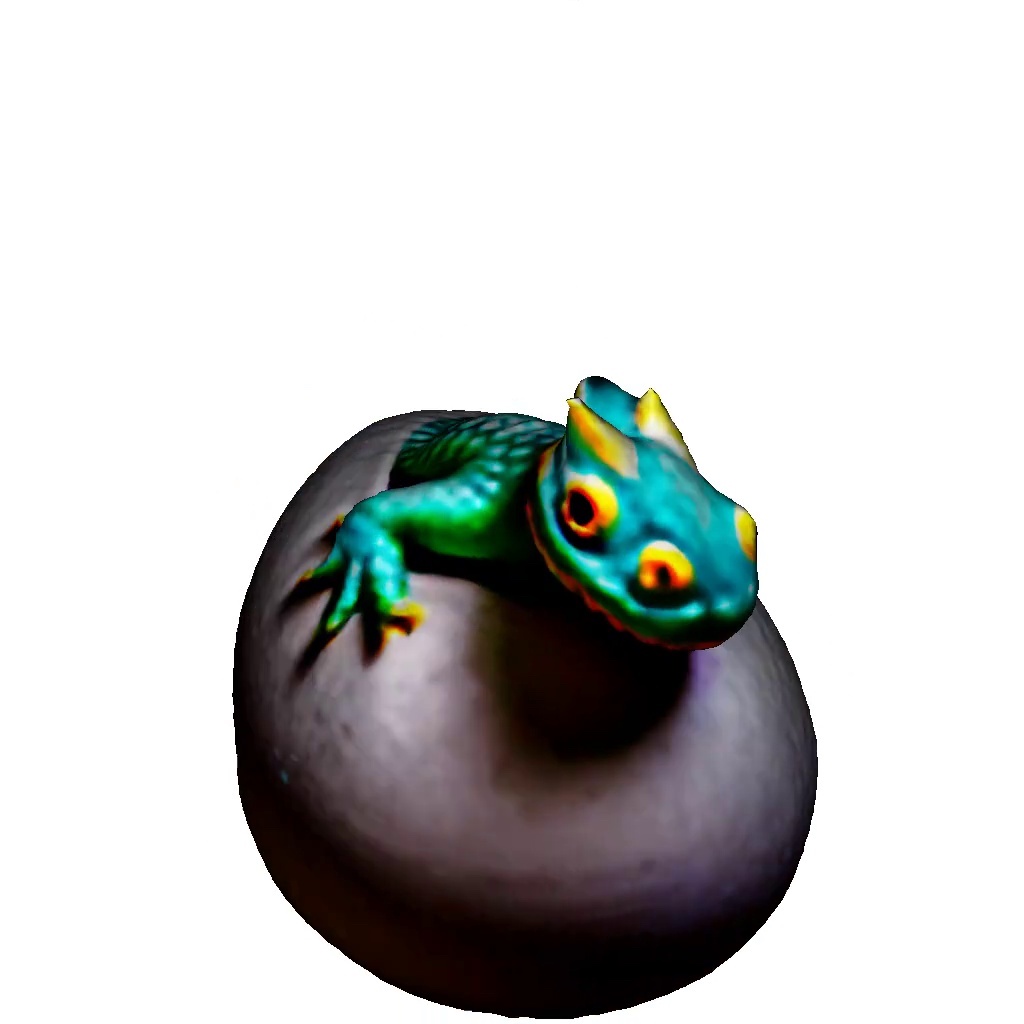} &
        \includegraphics[width=\linewidth]{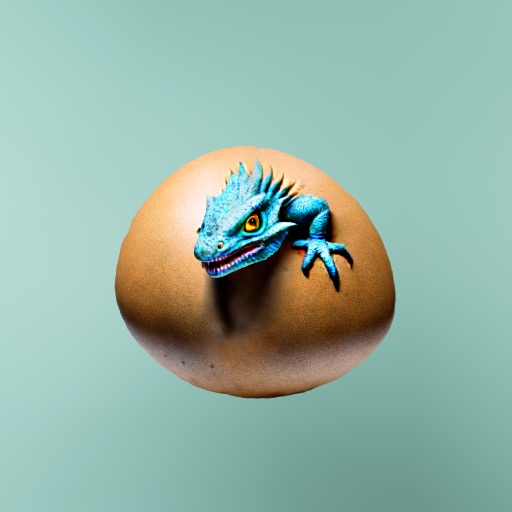} 
        \\
        DreamFusion & Magic3D  & NFSD (ours) \\
        \\[-9pt]
        \hline
        \hline \\[-8pt]
        \multicolumn{3}{c}{``An iguana holding a balloon''} \\ 
        \includegraphics[width=\linewidth]{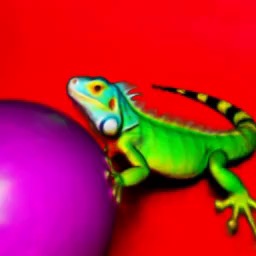} &
        \includegraphics[width=\linewidth]{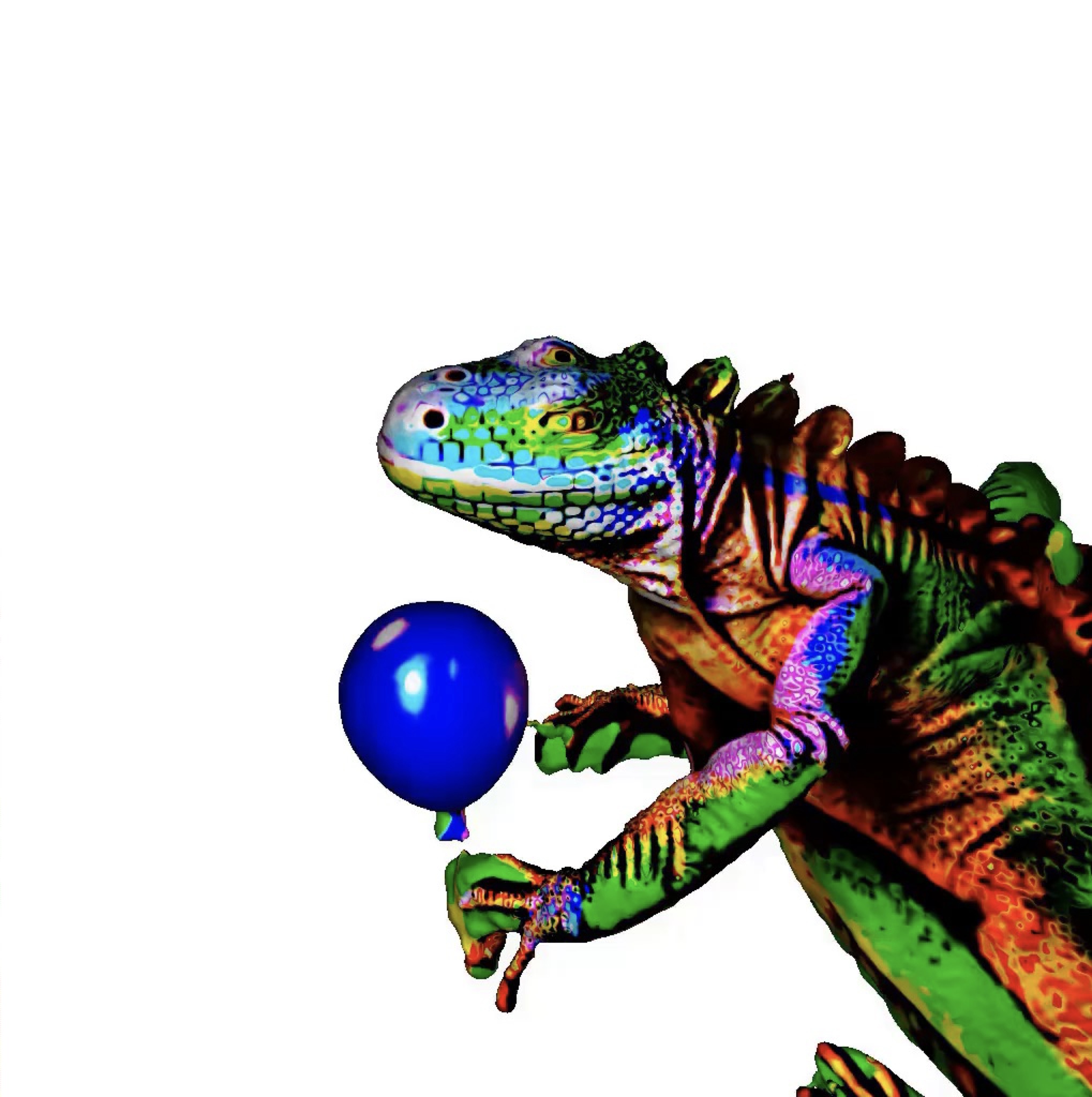} &
        \includegraphics[width=\linewidth]{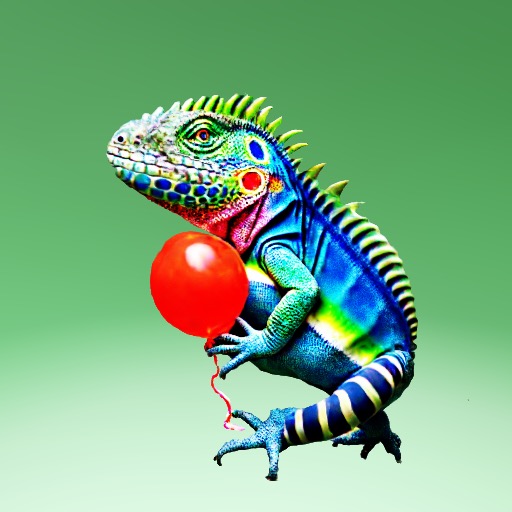} 
        \\
        DreamFusion & Magic3D  & NFSD (ours) \\
        \\[-9pt]
        \hline
        \hline \\[-8pt]
        \multicolumn{3}{c}{``A blue tulip''} \\ 
        \includegraphics[width=\linewidth]{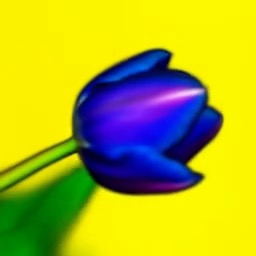} &
        \includegraphics[width=\linewidth]{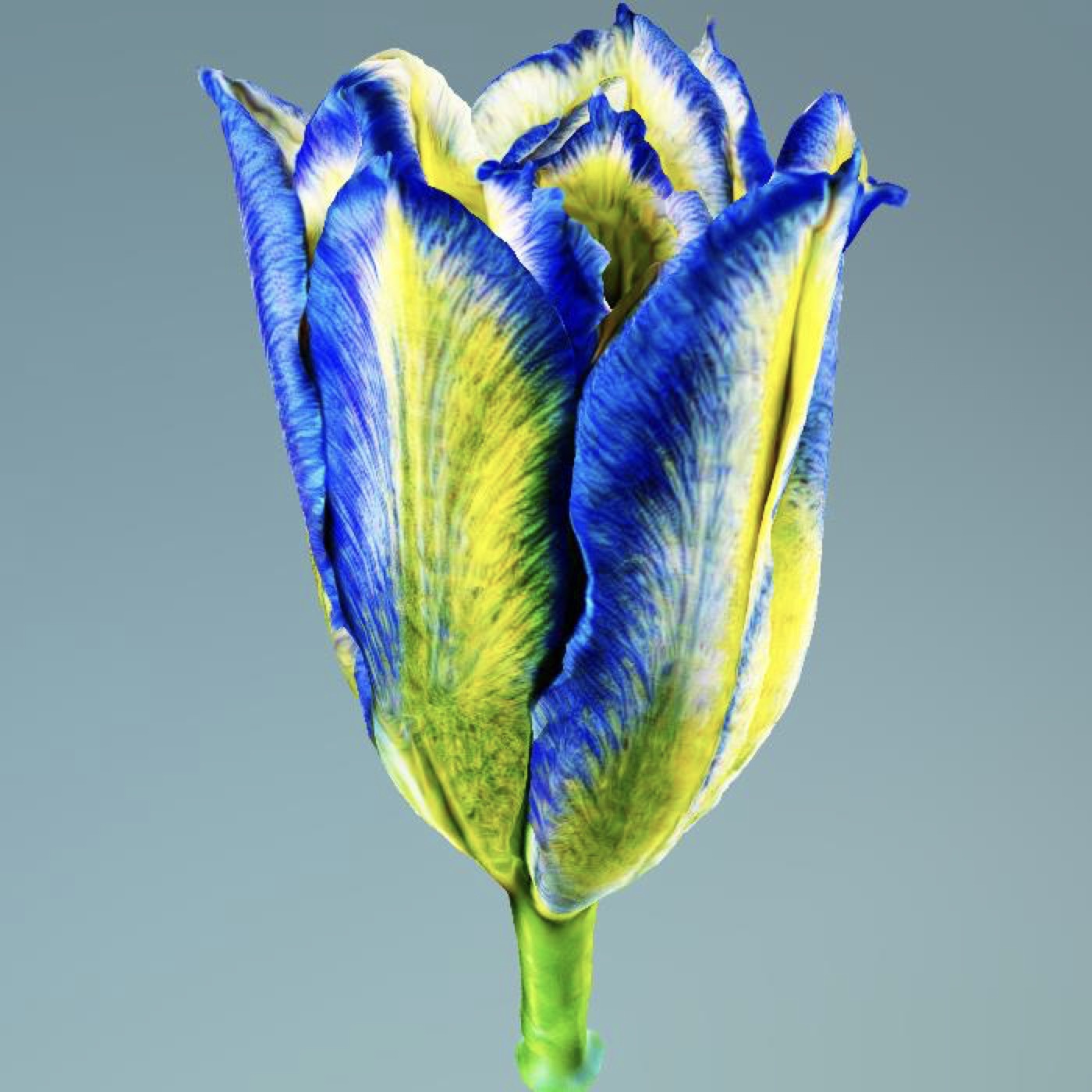} &
        \includegraphics[width=\linewidth]{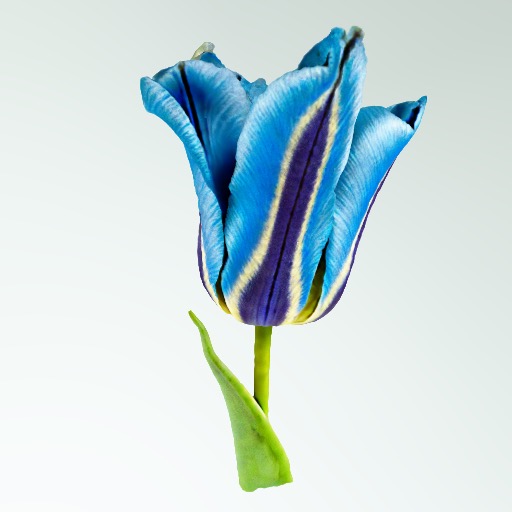} 
        \\
        DreamFusion & ProlificDreamer  & NFSD (ours) \\
        \\[-9pt]
        \hline
        \hline \\[-8pt]
        \multicolumn{3}{c}{``a cauldron full of gold coins''} \\ 
        \includegraphics[width=\linewidth]{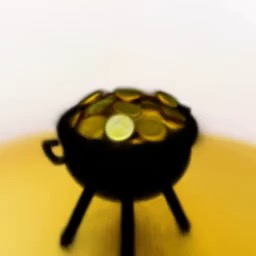} &
        \includegraphics[width=\linewidth]{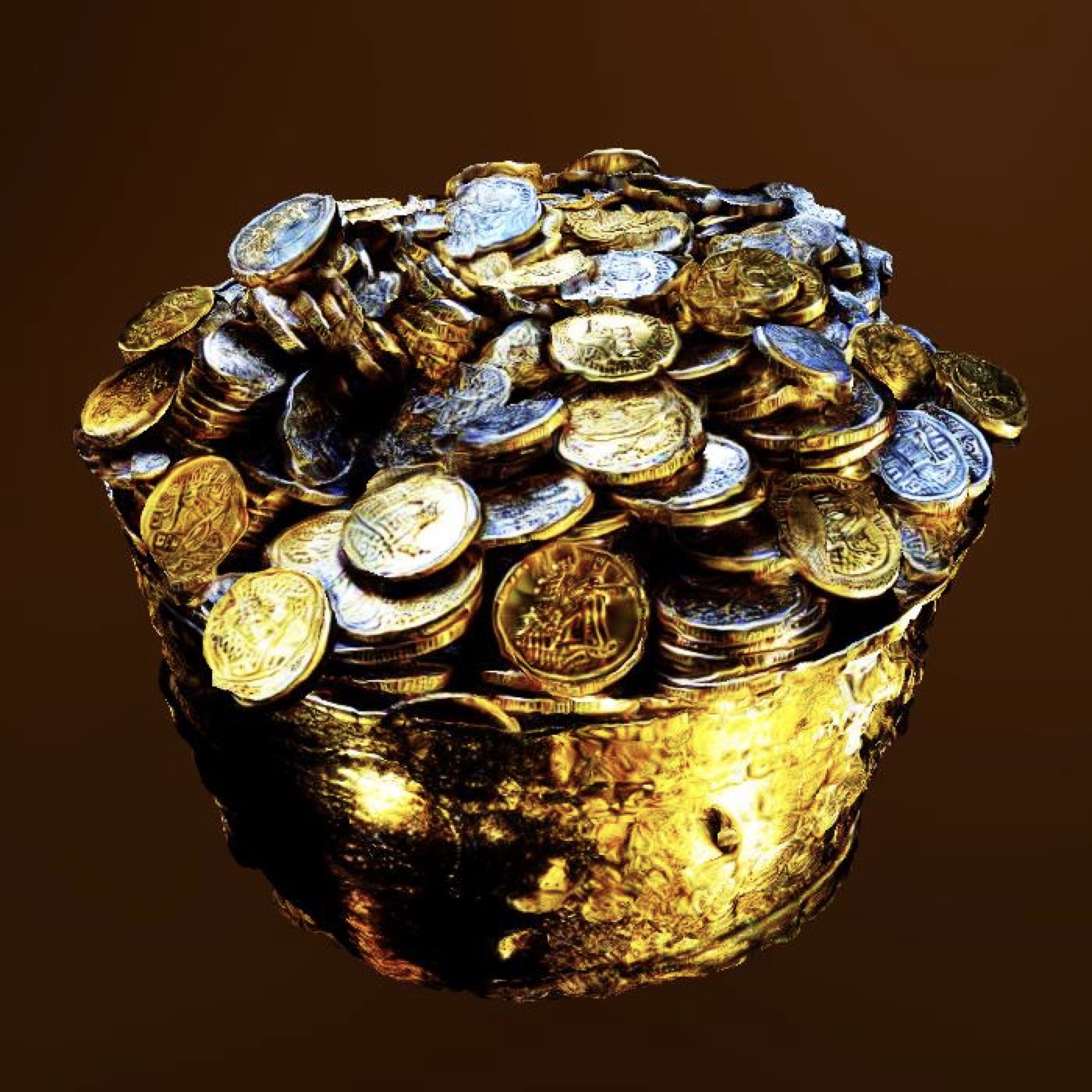} &
        \includegraphics[width=\linewidth]{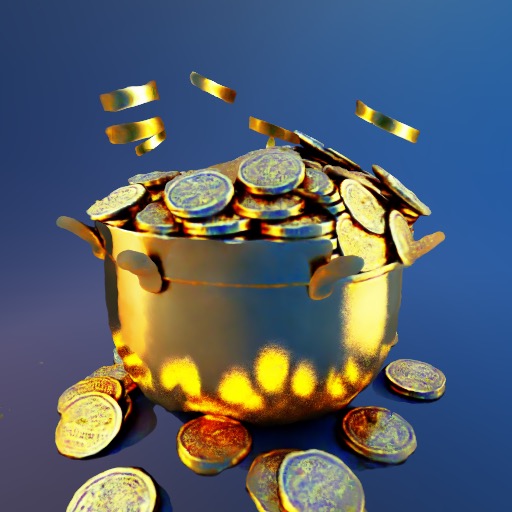} 
        \\
        DreamFusion & ProlificDreamer  & NFSD (ours) \\
        \\[-9pt]
        \hline
    \end{tabular}
    \vspace{-10pt}
    \caption{Comparison of NFSD with other methods using results obtained from the original papers.}
    \label{fig:app-comp-orig-3}
\end{figure}

\begin{figure}
    \centering
    \setlength{\tabcolsep}{1pt}
    \begin{tabular}{C{0.33\linewidth} C{0.33\linewidth} C{0.33\linewidth}}
        \hline \\[-8pt]
        \multicolumn{3}{c}{``Bagel filled with cream cheese and lox''} \\ 
        \includegraphics[width=\linewidth]{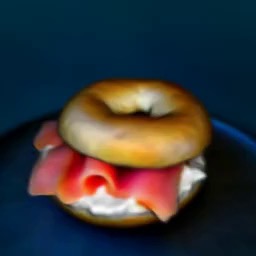} &
        \includegraphics[width=\linewidth]{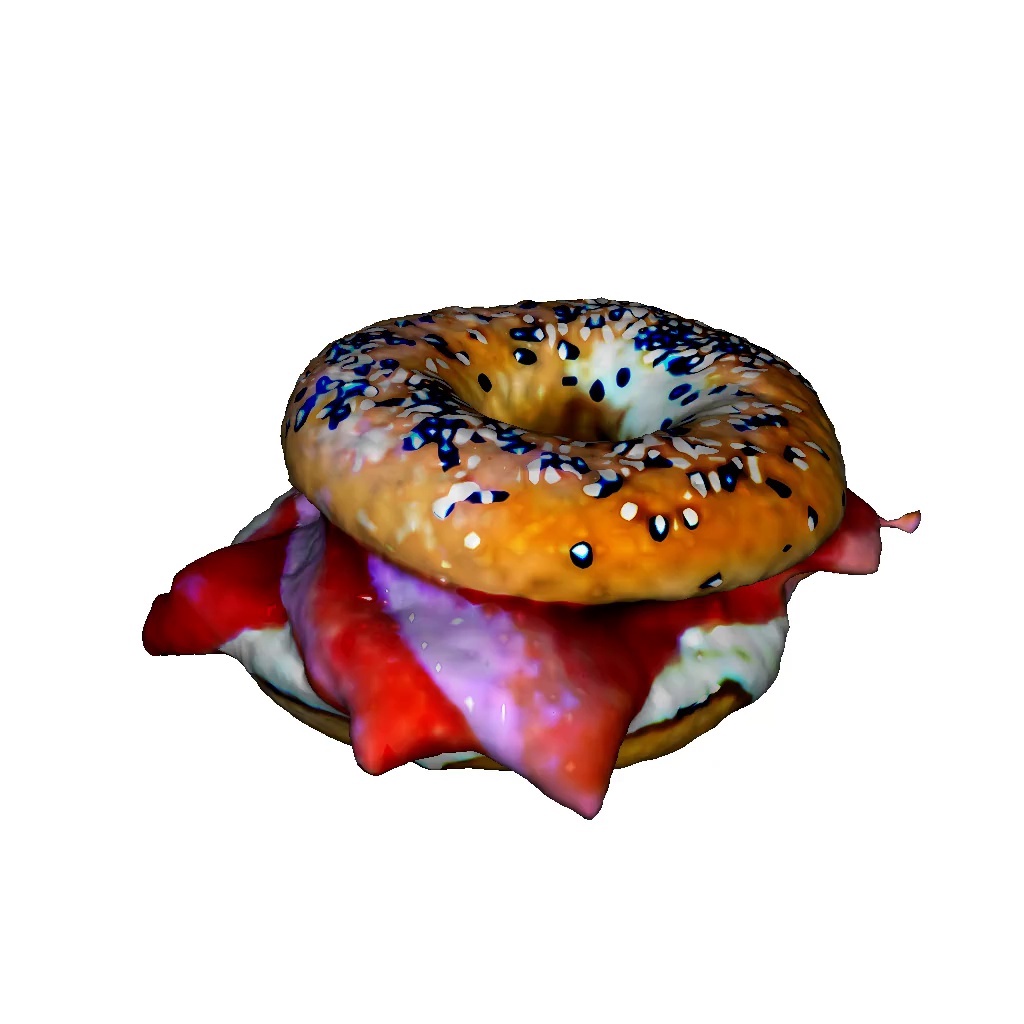} &
        \includegraphics[width=\linewidth]{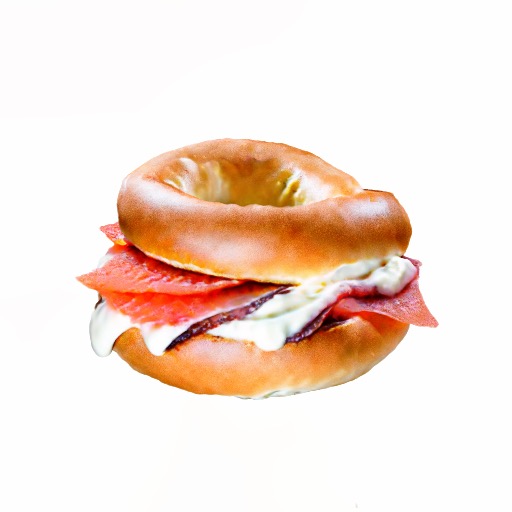} 
        \\
        DreamFusion & Magic3D  & NFSD (ours) \\
        \\[-9pt]
        \hline
        \hline \\[-8pt]
        \multicolumn{3}{c}{``A plush dragon toy''} \\ 
        \includegraphics[width=\linewidth]{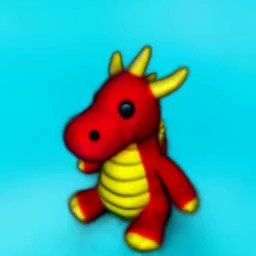} &
        \includegraphics[width=\linewidth]{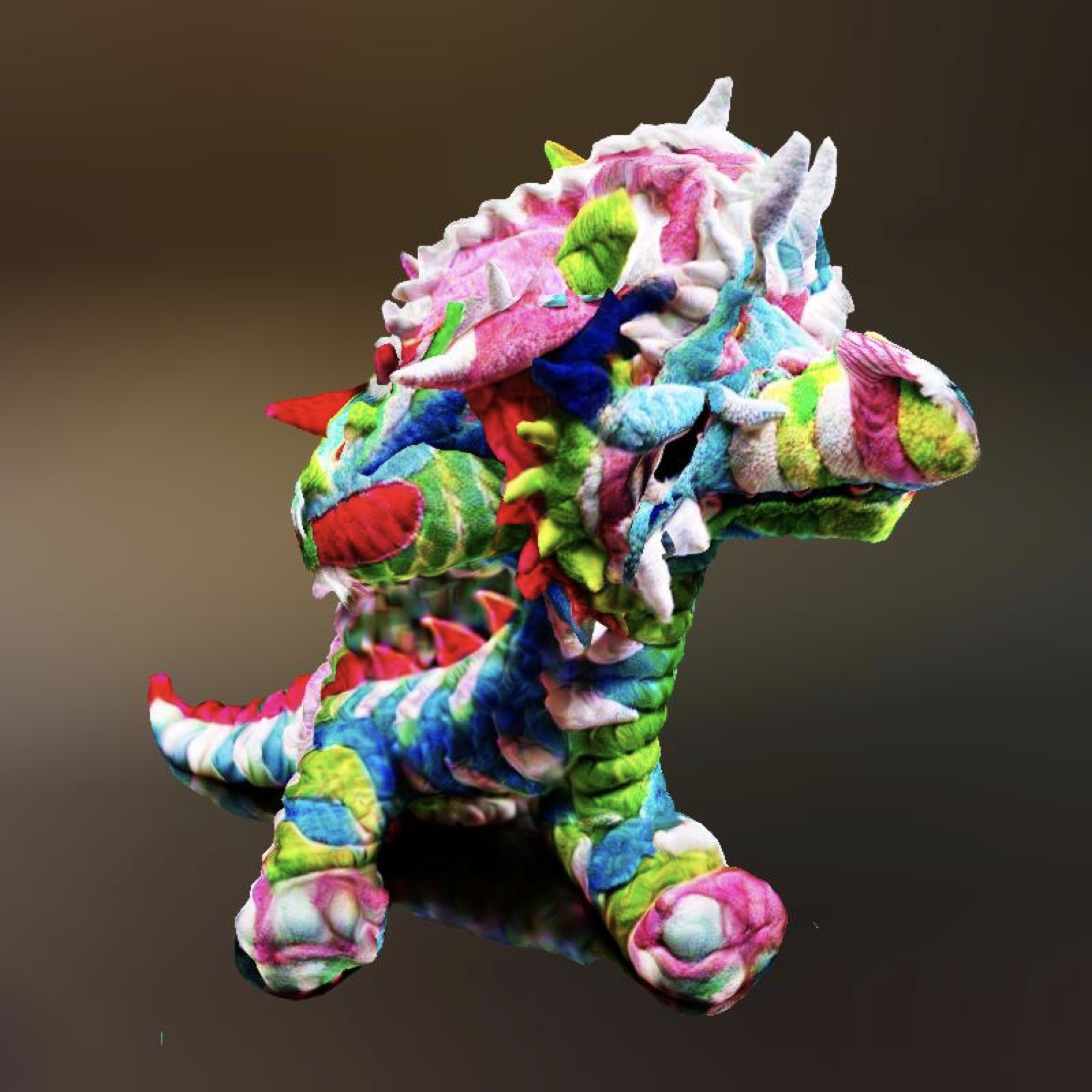} &
        \includegraphics[width=\linewidth]{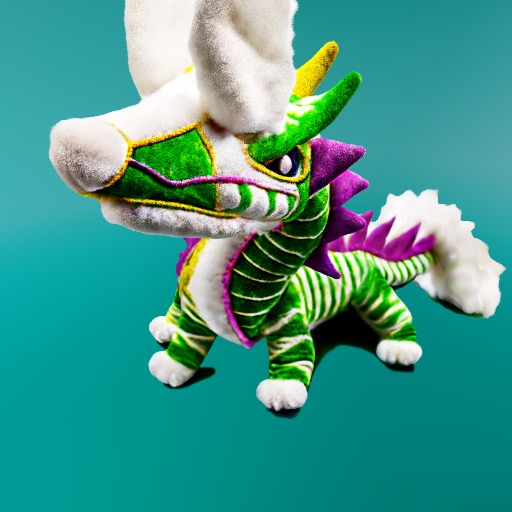} 
        \\
        DreamFusion & ProlificDreamer  & NFSD (ours) \\
        \\[-9pt]
        \hline
        \hline \\[-8pt]
        \multicolumn{3}{c}{``A ceramic lion''} \\ 
        \includegraphics[width=\linewidth]{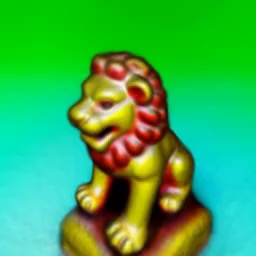} &
        \includegraphics[width=\linewidth]{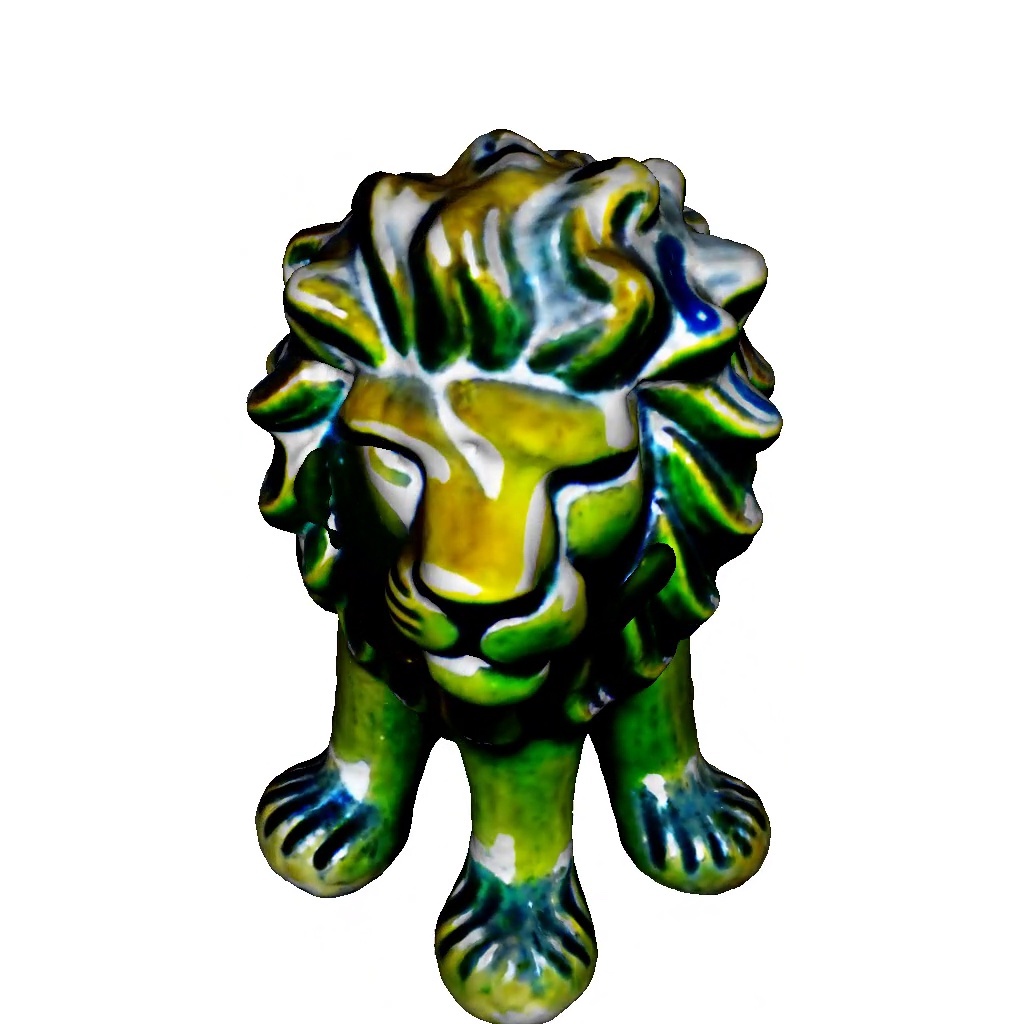} &
        \includegraphics[width=\linewidth]{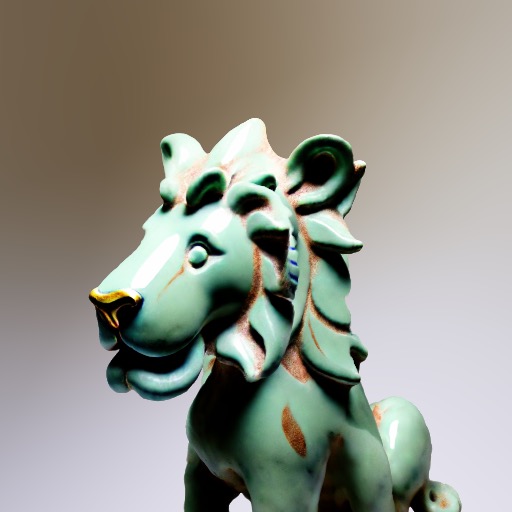} 
        \\
        DreamFusion & Magic3D  & NFSD (ours) \\
        \\[-9pt]
        \hline
        \hline \\[-8pt]
        \multicolumn{3}{c}{``Tower Bridge made out of gingerbread and candy''} \\ 
        \includegraphics[width=\linewidth]{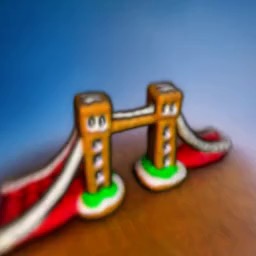} &
        \includegraphics[width=\linewidth]{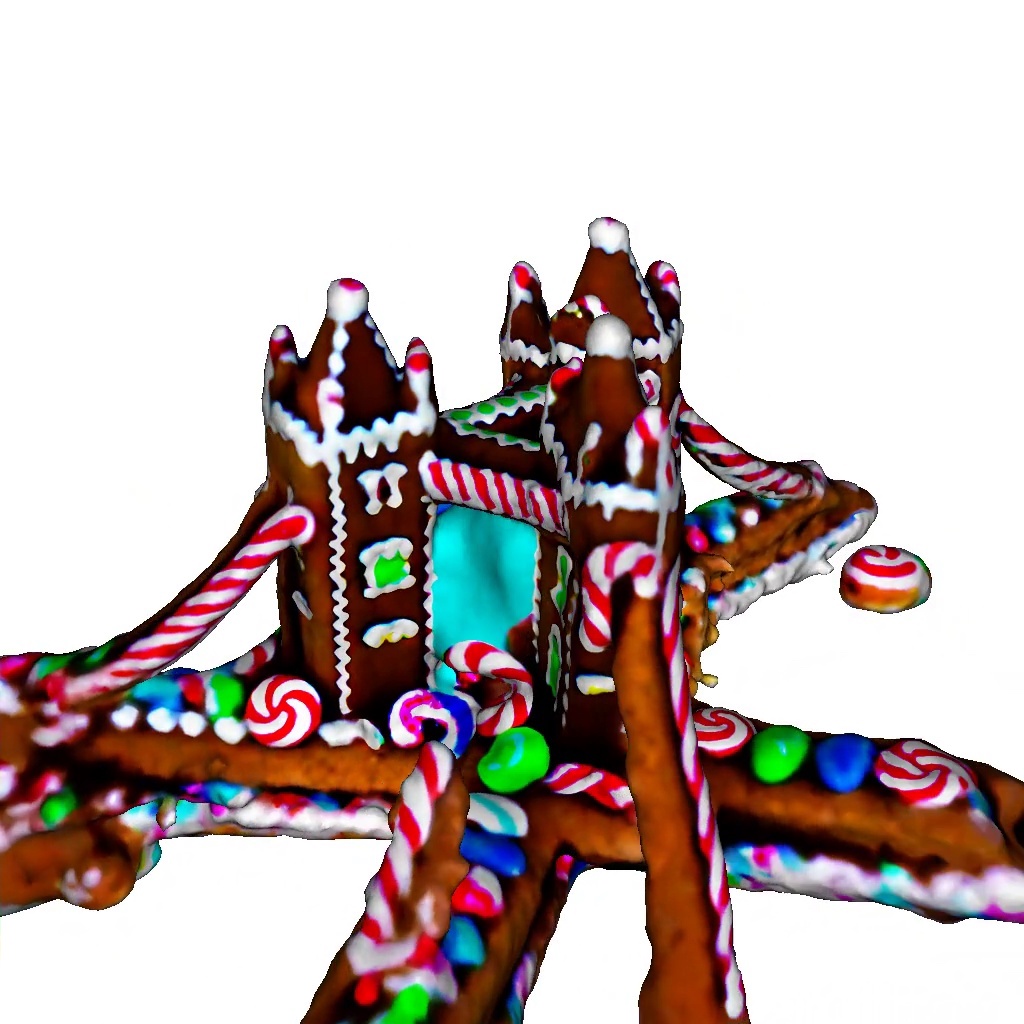} &
        \includegraphics[width=\linewidth]{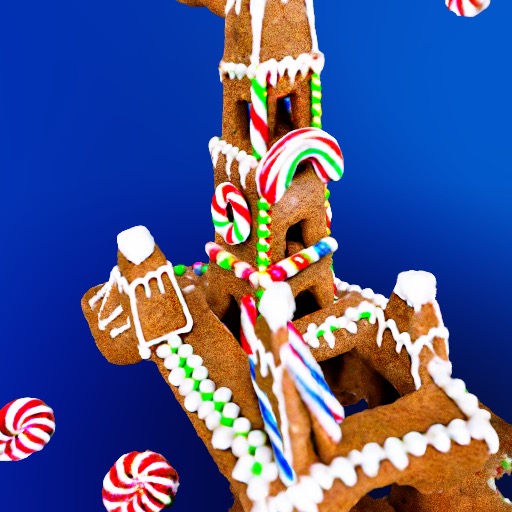} 
        \\
        DreamFusion & Magic3D  & NFSD (ours) \\
        \\[-9pt]
        \hline
    \end{tabular}
    \caption{Comparison of NFSD with other methods using results obtained from the original papers.}
    \label{fig:app-comp-orig-4}
\end{figure}

Additionally, in Figures~\ref{fig:app-comp-orig-1a},\ref{fig:app-comp-orig-1},\ref{fig:app-comp-orig-2a},\ref{fig:app-comp-orig-3} and \ref{fig:app-comp-orig-4} we show additional comparisons with results obtained by other methods, as reported in their respective original papers.


\end{document}